\PassOptionsToPackage{unicode}{hyperref}
\PassOptionsToPackage{hyphens}{url}
\documentclass[
  11pt,
]{article}
\usepackage{xcolor}
\usepackage{amsmath,amssymb}
\usepackage{iftex}
\ifPDFTeX
  \usepackage[T1]{fontenc}
  \usepackage[utf8]{inputenc}
  \usepackage{textcomp} 
\else 
  \usepackage{unicode-math} 
  \defaultfontfeatures{Scale=MatchLowercase}
  \defaultfontfeatures[\rmfamily]{Ligatures=TeX,Scale=1}
\fi
\usepackage{lmodern}
\ifPDFTeX\else
\fi
\IfFileExists{upquote.sty}{\usepackage{upquote}}{}
\IfFileExists{microtype.sty}{
  \usepackage[]{microtype}
  \UseMicrotypeSet[protrusion]{basicmath} 
}{}
\makeatletter
\@ifundefined{KOMAClassName}{
  \IfFileExists{parskip.sty}{%
    \usepackage{parskip}
  }{
    \setlength{\parindent}{0pt}
    \setlength{\parskip}{6pt plus 2pt minus 1pt}}
}{
  \KOMAoptions{parskip=half}}
\makeatother
\usepackage[margin=1in]{geometry}
\usepackage{amsmath,amssymb}
\usepackage{booktabs}
\usepackage{longtable}
\usepackage{multirow}
\usepackage{array}
\usepackage{calc}
\usepackage{caption}
\usepackage{graphicx}
\usepackage{pdflscape}
\usepackage{xcolor}
\usepackage{titling}
\usepackage{titlesec}
\titleformat{\section}[block]{\normalfont\large\bfseries\centering}{}{0em}{}
\titlespacing*{\section}{0pt}{2.5ex plus 1ex}{1.2ex}
\titleformat{\subsection}[block]{\normalfont\normalsize\bfseries\itshape}{}{0em}{}
\titlespacing*{\subsection}{0pt}{1.5ex}{0.4ex}

\newcounter{none}
\renewcommand{\baselinestretch}{1.5}
\usepackage{setspace}
\makeatletter
\renewcommand\@makefntext[1]{%
  \parindent 1em%
  \noindent\hb@xt@1.6em{\hss\@makefnmark}%
  \begingroup
    \scriptsize
    \setstretch{1.0}%
    #1\par
  \endgroup
}
\makeatother
\usepackage{hyperref}
\usepackage{needspace}
\definecolor{aercolor}{HTML}{1565C0}
\hypersetup{colorlinks=true,linkcolor=aercolor,urlcolor=aercolor,citecolor=aercolor,
            pdftitle={Who Uses AI? Platform Selection and the Measurement of Occupational AI Exposure},
            pdfauthor={Michelle Yin and Burhan Ogut},
            pdfcreator={LaTeX via pandoc}}
\pretitle{\begin{center}\Large\bfseries}
\posttitle{\par\end{center}\vskip 0.5em}
\preauthor{\begin{center}\large}
\postauthor{\par\end{center}}
\predate{}
\date{}
\postdate{}
\usepackage{bookmark}
\IfFileExists{xurl.sty}{\usepackage{xurl}}{} 
\renewenvironment{abstract}
  {\par\vspace{1em}\noindent\begin{center}\textbf{Abstract}\end{center}%
   \begin{list}{}{\setlength{\leftmargin}{0.5in}\setlength{\rightmargin}{0.5in}}%
   \item[]\small\setstretch{1.15}}
  {\end{list}\vspace{0.5em}}

\title{Who Uses AI? Platform Selection and the Measurement of Occupational AI Exposure}
\author{Michelle Yin and Burhan Ogut\thanks{\fontsize{7pt}{7.5pt}\selectfont\setlength{\parskip}{0pt}\setlength{\parsep}{0pt}\setlength{\itemsep}{0pt} Yin: School of Education and Social Policy, Northwestern University, 2120 Campus Drive, Evanston, IL 60208 (email: michelle.yin@northwestern.edu); Ogut: American Institutes for Research and School of Education and Social Policy, Northwestern University (email: burhan.ogut@northwestern.edu). We thank Alie Goldblatt, Megan Ashley Morrell, and Maggie Liao for excellent research assistance, and David Figlio, Peter Steiner, Hoa Vu, Regina Seo, Diego Guerrero, Ruhan Circi, and participants in the RISEI Lab Seminar Series at Northwestern University for helpful comments and suggestions. We are grateful to Alexander Bick, Adam Blandin, and David Deming for sharing the Real-Time Population Survey AI module micro release. The contents of this study were developed under Grant No.\ H421D220008 from the U.S. Department of Education, Rehabilitation Services Administration. The Department does not mandate or prescribe practices, models, or other activities described or discussed in this document. The contents of this study may contain examples of, adaptations of, and links to resources created and maintained by another public or private organization. The Department does not control or guarantee the accuracy, relevance, timeliness, or completeness of this outside information. The content of this study does not necessarily represent the policy of the Department. This publication is not intended to represent the views or policy of or be an endorsement of any views expressed or materials provided by any Federal agency (EDGAR \textsection{}75.620). The authors declare that they have no relevant or material financial interests that relate to the research described in this paper. Data and replication materials are available upon request from the corresponding author.}}
\date{}

\begin{document}
\renewcommand{\thefootnote}{\fnsymbol{footnote}}
\maketitle
\renewcommand{\thefootnote}{\arabic{footnote}}
\begin{abstract}
Conversation logs from AI platforms are increasingly used to measure occupational exposure to artificial intelligence, but the users observed in these logs are not a representative sample of the workforce. This paper shows that platform-derived exposure scores combine task-level AI applicability with the occupational composition of the platform's user base. Holding the empirical design fixed, changing only the platform input changes the post-ChatGPT employment coefficient by a factor of 1.9, and consumer and enterprise channels within the same vendor disagree in sign. We formalize the resulting non-classical measurement error, decompose it into between- and within-occupation selection, and construct workforce-reweighted partial-identification bounds. Reweighting to Bureau of Labor Statistics employment shares attenuates estimates by 42 to 93 percent. The bias captures augmentation among observed users more directly than substitution in the workforce.


\end{abstract}
\clearpage

A growing empirical literature measures occupational exposure to artificial intelligence using conversation logs from platforms such as Anthropic's Claude (\hyperlink{Handa2025}{Handa et al.\ 2025}; \hyperlink{Appel2025}{Appel et al.\ 2025}), OpenAI's ChatGPT (\hyperlink{Chatterji2025}{Chatterji et al.\ 2025}), and Microsoft Copilot (\hyperlink{Tomlinson2025}{Tomlinson et al.\ 2025}). These measures represent an important advance because they move exposure measurement from hypothetical capability toward observed use. The measurement problem is that observed platform use is not drawn from the workforce, but from the users who enter a particular platform. A platform-derived exposure score therefore combines the task content of an occupation with the occupational composition of the platform's user base. When platform participation differs systematically from workforce employment, estimates built on these scores inherit platform-specific selection.

This distinction matters because occupational AI exposure measures increasingly enter empirical work on employment, wages, and labor-demand reallocation as fixed regressors. A measure used in this way is implicitly treated as a stable characteristic of the occupation. Under that interpretation, changing the platform source should not materially change the coefficient once the empirical design is held fixed. In our estimates, changing only the platform input changes the employment coefficient substantially in magnitude and reverses its sign in several comparisons, even when the regression design is held fixed. The regressor therefore varies with both occupational exposure to AI and the occupational composition of the platform that produced it.

The measurement problem follows from how the AI exposure literature has evolved out of the task-based tradition in labor economics. \hyperlink{Autor2003}{Autor, Levy, and Murnane (2003)} treat an occupation as a bundle of tasks and ask which tasks a technology can substitute for or complement. That framework shaped the automation and job-polarization literature, where the choice of routine-task index affected downstream estimates and interpretation (\hyperlink{Acemoglu2011}{Acemoglu and Autor 2011}; \hyperlink{Goos2014}{Goos, Manning, and Salomons 2014}; \hyperlink{Acemoglu2020}{Acemoglu and Restrepo 2020}). Early AI exposure measures adapted this task-based logic using inputs external to observed platform use, including expert ratings (\hyperlink{Frey2017}{Frey and Osborne 2017}), patent-text matching (\hyperlink{Webb2020}{Webb 2020}), AI-capability benchmarks (\hyperlink{Brynjolfsson2018}{Brynjolfsson, Mitchell, and Rock 2018}; \hyperlink{Felten2021}{Felten, Raj, and Seamans 2021}), and the task-level applicability rubric of \hyperlink{Eloundou2024}{Eloundou et al.\ (2024)}, which has become one of the central capability-based measures in the literature.

Platform logs add a second source of measurement by observing where AI appears in use rather than only what AI could do. Recent work has begun to combine these margins. \hyperlink{Massenkoff2026}{Massenkoff and McCrory (2026)}, for example, weight the Eloundou rubric by Anthropic conversation shares and report that the resulting composite predicts Bureau of Labor Statistics employment projections while the rubric alone does not. This finding explains why platform data have become more popular. They bring exposure measurement closer to economic use, but they also change the source of variation in the measure. Once conversation shares enter the score, measured exposure reflects task-level AI applicability together with the occupations the platform reaches.

Existing labor-market data leave room for platform measures because no single source captures capability, adoption, and task intensity at scale. Worker surveys measure AI adoption in a population frame and observe both users and nonusers. \hyperlink{Bick2026}{Bick, Blandin, and Deming (2026)} and the U.S. Census Bureau's Household Trends and Outlook Pulse Survey (\hyperlink{USCensus2025}{U.S. Census Bureau 2025}) field such measures nationally, and the Organisation for Economic Co-operation and Development (OECD) has developed cross-national instruments for manufacturing and finance workers (\hyperlink{Lane2023}{Lane, Williams, and Broecke 2023}). These surveys identify the extensive margin of use, but their per-wave samples are often too small for detailed occupation-level rates and provide less task-level detail than platform logs. Administrative records and field experiments observe use more precisely in specific settings (\hyperlink{Noy2023}{Noy and Zhang 2023}; \hyperlink{BrynjolfssonLi2023}{Brynjolfsson, Li, and Raymond 2023}; \hyperlink{Cui2024}{Cui, Demirer, Jaffe, Musolff, Peng, and Salz 2024}; \hyperlink{DellAcqua2023}{Dell'Acqua, McFowland, Mollick, Lifshitz-Assaf, Kellogg, Rajendran, Krayer, Candelon, and Lakhani 2023}), but their workforces and task domains limit generalizability. Platform logs sit between these sources because they provide revealed task-level behavior at large scale, while the scale itself comes from a selected platform population rather than from a workforce sample.

The selection embedded in platform logs is large enough to change inference. A platform's conversation distribution is shaped by pricing, product launches, enterprise contracts, competitive position, and the occupations for which the platform is useful enough to generate repeated use. Computer and Mathematical occupations generate 32 percent of consumer Claude conversations and 52 percent of enterprise conversations (\hyperlink{Handa2025}{Handa et al.\ 2025}; \hyperlink{Appel2025}{Appel et al.\ 2025}), compared with 3.4 percent of U.S. employment. Food Preparation workers hold 8.8 percent of employment but generate only 0.7 percent of conversations. Across the 22 Standard Occupational Classification major groups, the ratio of conversation density to employment density spans a factor of 72, and no platform-derived measure correlates with BLS employment shares above a Spearman rank correlation of 0.33. A regression using such a measure loads on both underlying AI applicability and platform penetration. We show formally that the resulting coefficient generally does not identify the structural employment elasticity.

Our empirical design isolates the user-base channel by rebuilding the composite exposure score of \hyperlink{Massenkoff2026}{Massenkoff and McCrory (2026)}, which weights the \hyperlink{Eloundou2024}{Eloundou et al.\ (2024)} per-task rubric by each occupation's platform conversation share. We hold fixed the specification, sample, outcome, estimator, and task-capability rubric, and vary only the source of the conversation shares. The ten variants come from Anthropic Claude consumer and enterprise channels across several waves and from Microsoft Copilot. Holding the rubric fixed isolates the contribution of platform conversation shares, which is important because capability ratings are themselves sensitive to the model used to rate them (\hyperlink{Yin2026}{Yin, Vu, and Persico 2026}). We estimate the resulting specifications on a 2015 to 2024 American Community Survey panel of 13.1 million person-year observations merged at the six-digit occupation level, using a difference-in-differences design common in the AI-and-labor literature.

The estimates move in ways that are difficult to reconcile with a platform-independent occupational exposure measure. Across the ten variants, the post-2022 employment coefficient ranges by a factor of 1.9. Within Anthropic, replacing the consumer channel with the enterprise channel reverses the sign of the employment estimate at every observed release, even though the two channels share the same vendor, model family, classification procedure, and task rubric. This comparison leaves user-base composition as the remaining economically relevant source of variation. The exposure measure also changes over time within the same channel. A researcher using the earliest Claude consumer release would estimate that a one-standard-deviation increase in exposure lowers employment by 0.12 percentage points, while a researcher using the latest release would estimate a decline of 0.22 percentage points. The difference comes only from the wave of platform data used to construct exposure.

We formalize this instability as non-classical measurement error (\hyperlink{Bound2001}{Bound, Brown, and Mathiowetz 2001}; \hyperlink{Meng2018}{Meng 2018}). The platform proxy combines true occupational exposure with platform-specific selection, and that selection is correlated with the task structure the measure is intended to capture. The framework decomposes the bias into a between-occupation margin, which captures which occupations enter the platform user base, and a within-occupation margin, which captures which tasks workers perform once observed on the platform. Because the between-occupation component is observed in platform and workforce occupation shares, reweighting platform shares to BLS workforce shares removes that component under the maintained condition that within- and between-occupation selection move in the same direction. The baseline and workforce-reweighted estimates then define endpoints of a partial-identification interval for the structural employment elasticity (\hyperlink{Manski2003}{Manski 2003}). The width of the interval measures how much of the estimated association is attributable to observable platform selection rather than to the relationship between AI exposure and employment.

The empirical results follow the ordering implied by the framework. Measures drawn from user bases farther from the workforce move more after reweighting. For the Anthropic composite, reweighting attenuates the employment coefficient by 93 percent and leaves an estimate statistically indistinguishable from zero. For Microsoft Copilot, whose user base is closer to workforce employment shares, the coefficient attenuates by 42 percent and remains statistically significant. The same channel also appears in the published cross-occupation specification. Reweighting the \hyperlink{Massenkoff2026}{Massenkoff and McCrory (2026)} composite to workforce shares reduces the projected employment-growth association from 0.70 to 0.38 percentage points for a one-standard-deviation increase in exposure. This implies that 45 percent of the published magnitude disappears when the platform's occupational composition is replaced with workforce composition.

The paper contributes to the literature in four ways. It identifies a measurement problem specific to platform-derived AI exposure, where the error is selected on the same occupation-level task structure the measure is intended to capture. It then provides empirical diagnostics showing that platform source, product channel, and platform wave change downstream coefficients even when the regression design is fixed. The workforce-reweighting procedure and partial-identification bounds convert this instability into an estimable component of the coefficient. Finally, the distributional analysis shows that platform selection is correlated with wages, education, and occupational advantage, implying that exposure rankings based on platform logs can underweight occupations where low visibility reflects limited access, low adoption, or displacement risk rather than low exposure.

The findings clarify the parameter that platform data can support rather than diminishing the value of platform logs. Platform data are informative about task intensity among active users, especially at a scale and frequency unavailable in surveys or administrative records. They are less informative about nonusers, workers weakly represented on the platform, and tasks or workers displaced from the observed occupation-platform pair. Usage-based measures therefore recover augmentation among observed users more directly than substitution in the workforce. This distinction changes both inference and allocation. In a hypothetical retraining exercise, a platform-weighted ranking directs 39 percent of a ten-billion-dollar fund to occupations that a workforce-weighted ranking does not identify as exposed. More generally, applied work using platform-derived exposure should report workforce-reweighted counterparts, test sensitivity to platform source, and distinguish task capability, worker adoption, and use conditional on adoption. Without these distinctions, variation in measured exposure can reflect platform adoption across occupations rather than variation in underlying AI applicability.

The rest of the paper proceeds as follows. Section I develops the measurement framework. Section II documents platform-workforce divergence. Section III estimates its consequences for labor-market coefficients and partial-identification bounds. Section IV examines distributional incidence, and Section V reports robustness checks.

\section{I. Conceptual Framework}

Let $o$ index the six-digit occupational categories used throughout the paper, and let $E_o$ denote true AI exposure. We define $E_o$ as the share of an occupation's task bundle for which AI can substitute for or complement worker input at comparable quality. The relationship of interest is

\begin{equation}
Y_o = \beta E_o + X_o'\delta + \varepsilon_o, \tag{1}
\end{equation}

where $Y_o$ is a labor-market outcome, $X_o$ is a vector of controls, and $E[\varepsilon_o \mid E_o, X_o] = 0$. In practice, $E_o$ is unobserved. Researchers instead condition on a platform-derived proxy, $E_{o,p}$, constructed from conversation logs on platform $p$. The central measurement question is whether $E_{o,p}$ can be treated as a fixed occupational characteristic, analogous to a routine-task index or a robot-exposure measure. It cannot when the proxy depends on which occupations enter the platform's user base.

The departure from workforce exposure has two margins. Between occupations, let $f_p(o)$ be the share of platform $p$'s conversations associated with occupation $o$, and let $f(o)$ be the occupation's employment share in the BLS Occupational Employment and Wage Statistics. Define

\begin{equation}
\psi_{o,p} \equiv \frac{f_p(o)}{f(o)}. \tag{2}
\end{equation}

The parameter $\psi_{o,p}$ measures platform representation relative to the workforce. It exceeds one for occupations overrepresented on the platform and falls below one for occupations underrepresented on the platform. Within occupations, let $\theta_{o,k,p}$ measure how the platform share of task $k$ within occupation $o$ differs from the corresponding workforce task share. The two parameters capture distinct sources of selection. The platform-selection parameter $\psi_{o,p}$ governs which occupations appear in platform data. The task-selection parameter $\theta_{o,k,p}$ governs which tasks workers perform conditional on appearing there.

With this notation, the platform-derived proxy can be written as

\begin{equation}
E_{o,p} = \psi_{o,p} E_o + \eta_{o,p} + u_{o,p}, \tag{3}
\end{equation}

where $\eta_{o,p}$ collects within-occupation task-selection differences and $u_{o,p}$ is classical noise. Appendix A.2 derives this expression from task-level conversation shares. The between-occupation component, $\psi_{o,p}$, is observed from platform and workforce occupation shares. The within-occupation component, $\theta_{o,k,p}$, is not observed with current data because no worker-representative source links task-time use to the \textit{O*NET} task categories used by the platform measures.

Equation (3) implies non-classical measurement error. The error is not random noise around a fixed occupational exposure. It is correlated with true exposure when occupations overrepresented on the platform are also occupations with high task-level AI applicability. The data display exactly this pattern. Computer and Mathematical occupations are far more represented in platform conversations than in the workforce, while many service and manual occupations are underrepresented. A platform-derived exposure score therefore combines task exposure with platform penetration.

To see the implication for regression coefficients, suppose a researcher estimates equation (1) after replacing $E_o$ with $E_{o,p}$. After residualizing with respect to $X_o$, write the linear projection of the proxy on true exposure as

\[
E_{o,p} = \lambda_p E_o + v_{o,p}, \qquad \operatorname{Cov}(E_o, v_{o,p}) = 0.
\]

Let $\kappa_p = \operatorname{Var}(E_o)/\operatorname{Var}(v_{o,p})$. The probability limit of the coefficient on the platform proxy is

\begin{equation}
\operatorname{plim} \widehat{\beta}_p = \frac{\beta \lambda_p \kappa_p}{\lambda_p^2 \kappa_p + 1}. \tag{4}
\end{equation}

Appendix A.3 gives the derivation. The expression differs from the classical errors-in-variables formula because $\lambda_p$ need not equal one. When the platform user base mirrors the workforce, $\lambda_p = 1$, and equation (4) reduces to classical attenuation. When platform representation is skewed toward high-exposure occupations, $\lambda_p$ differs from one, and the coefficient becomes platform-specific. A regression with the same outcome, controls, sample, and estimator can therefore recover different coefficients solely because the exposure measure is drawn from a different platform or product channel.

This logic has two empirical implications. First, coefficients should vary across platforms and product channels if those platforms and channels reach different occupations. Second, coefficients should move over time within a platform when product launches, pricing changes, or enterprise contracts shift the user base. Additional conversations from the same selected platform improve precision for the platform-specific measure, but they do not change the population represented in the proxy. The probability limit remains platform-specific.

The framework also points to a correction. Because $\psi_{o,p}$ is observed, we can remove the between-occupation component by reweighting the proxy to the workforce:

\begin{equation}
\widetilde{E}_{o,p} \equiv \frac{E_{o,p}}{\psi_{o,p}}. \tag{5}
\end{equation}

Substituting equation (3) gives

\begin{equation}
\widetilde{E}_{o,p} = E_o + \frac{\eta_{o,p}}{\psi_{o,p}} + \frac{u_{o,p}}{\psi_{o,p}}. \tag{6}
\end{equation}

If within-occupation task selection is absent, so that $\eta_{o,p} = 0$, the reweighted proxy recovers true exposure up to classical noise. If within-occupation task selection remains, reweighting does not identify $\beta$ by itself, but it removes the observable between-occupation channel. Under the maintained ordering condition that within-occupation selection moves in the same direction as between-occupation selection, the baseline and workforce-reweighted estimates form endpoints of a partial-identification interval for the structural employment elasticity. The width of this interval measures the contribution of observable platform composition to the coefficient.

This framework locates platform-derived AI exposure outside the standard treatment of occupational exposure as predetermined. Routine-task indices and robot-exposure measures are fixed by construction. Platform-derived measures move with product design, pricing, and user adoption. They are therefore closer to non-probability samples, where the relevant distortion depends on the correlation between selection and the variable of interest rather than on sample size (\hyperlink{Meng2018}{Meng 2018}). In this setting, the corresponding object is the relationship between $\psi_{o,p}$ and $E_o$. If high-exposure occupations are also overrepresented on the platform, the platform proxy does not converge to the workforce exposure measure simply by adding more conversations.

\section{II. Platform Exposure Measures and the Workforce Benchmark}

This section describes the exposure measures, workforce benchmarks, and analytical samples used in the empirical analysis. It then establishes the central empirical premise of the paper: platform-derived exposure measures are not workforce-representative measures of occupational AI exposure. We document this in three ways. First, platform occupation shares differ sharply from workforce employment shares. Second, platform exposure rankings change over time within the same vendor. Third, workforce reweighting changes which occupations are classified as most exposed.

\subsection{A. AI Exposure Measures and Benchmark Data}

The primary platform source is the Anthropic Economic Index (AEI, \hyperlink{Handa2025}{Handa et al. 2025}, \hyperlink{Appel2025}{Appel et al. 2025}), which maps anonymized Claude conversation transcripts to \textit{O*NET} task descriptions and aggregates them to per-occupation conversation shares at the six-digit Standard Occupational Classification (SOC) level. The mapping uses the Clio classifier (\hyperlink{Tamkin2024}{Tamkin et al. 2024}), which embeds conversation text and the \textit{O*NET} task descriptions in a shared vector space and assigns each conversation to its nearest task. The AEI has five publicly released waves from December 2024 to February 2026. These waves span five Claude model versions, from Claude 3.5 Sonnet in Wave 1 to Opus 4.6 in Wave 5. The last three separately report Claude.ai consumer use and enterprise first-party Application Programming Interface (API) use.

The timing of the Anthropic waves is useful for this study because the period includes changes that plausibly shift the user base. Claude Code launched in March 2025, Claude Projects expanded in mid-2025, pricing changed at least twice, and the enterprise customer base expanded substantially. These changes mean that cross-wave variation is not only a comparison of model releases. It is also a comparison of the workers and occupations that enter the platform conversation pool. If measured occupational exposure changes around these events, the change reflects platform adoption as well as task-level AI applicability.

We also use several platform-derived measures outside Anthropic. The Microsoft Copilot AI applicability score is a six-digit occupation-level measure constructed from roughly 200,000 Copilot conversations sampled in September 2024 and scored on a zero-to-one applicability scale. We use the Copilot measure in raw and workforce-reweighted form. OpenAI ChatGPT plays a different role. We use the public ChatGPT occupational distribution as a second-platform reference distribution, but it does not enter the ten-measure regression set because the public release reports occupational shares at the SOC major-group level rather than at the six-digit SOC level.

The \hyperlink{Massenkoff2026}{Massenkoff and McCrory (2026)} composite aggregates AEI conversation shares to occupations using \textit{O*NET} task-time weights and the \hyperlink{Eloundou2024}{Eloundou et al. (2024)} GPT-4 applicability rubric. Three composite variants enter the analysis. The first is the published baseline composite. The second replaces platform conversation density with OEWS workforce density and the third replaces the original conversation weights with Wave 5 AEI weights. Together with three Claude.ai consumer waves, two enterprise API waves, and two Copilot variants, these three composites make up the ten platform-derived exposure measures. The composite enters the paper in two roles. In the measurement and DiD exercises, the composite is one of the ten platform-derived exposure measures used to test cross-platform dispersion. In the cross-occupation exercise, we return to the between-occupation regression for which the composite was originally proposed and ask whether the same user-base channel changes that published estimate.

We draw on three benchmark data sources. The BLS OEWS May 2024 release supplies the workforce density, $f(o)$, used to construct $\psi_{o,p}$ and implement the workforce reweighting. The BLS Employment Projections for 2024 to 2034 supply the dependent variable for the between-occupation regression. The Real-Time Population Survey (RPS) from \hyperlink{Bick2026}{Bick, Blandin, and Deming (2026)} provides an independent worker-reported measure of at-work AI use at the six-digit SOC level (Online Appendix Table B.12 documents the RPS micro release at the SOC major-group and six-digit-SOC level).

A related measurement concern is the mapping from conversations to tasks. \hyperlink{BBDS2026}{Bick, Blandin, Deming, and Schumacher (2026)} show that chat-based task shares diverge from worker-survey task shares because the chat-to-task step over-attributes conversations to generic \textit{O*NET} tasks. More than 15 percent of OpenAI chats are assigned to \"Edit written materials or documents,\" although only 1.6 percent of workers are employed in occupations containing that task. In their RPS data, the top 5 percent of tasks account for 19 percent of generative AI use, compared with 57 percent in chat-based data. The correlation between chat-based and survey-based measures is below 0.4.

Our mechanism is distinct. Their distortion arises within occupations, through the mapping from conversations to tasks. Ours arises between occupations, through which occupations enter the platform conversation pool. The two mechanisms can operate together. The workforce reweighting in Section I removes the between-occupation component, while any remaining gap reflects within-occupation task selection and classification differences.

\subsection{B. Analytical Sample}

The worker-level analysis uses a 2015 to 2024 ACS person-year panel constructed from Integrated Public Use Microdata Series (IPUMS). We restrict the sample to civilian wage-and-salary workers ages 16 to 64 with a valid six-digit SOC code and at least one of the ten exposure measures defined for the worker's occupation. The panel contains 13.1 million person-year observations across 553 occupations and ten survey years. Each record links the worker's occupation to a platform-derived exposure measure, workforce density, demographic controls, and labor-market outcomes.

Five of the ten measures, the three composite variants and the two Copilot variants, need a crosswalk between the ACS occupation code and the National Employment Matrix (NEM) code that the composite and the Copilot release use. The crosswalk reduces the analytical sample to 65.9 percent of the full (Online Appendix Table B.3 documents crosswalk coverage by SOC major group). The matched sample is more female (0.50 against 0.39), younger (39.6 against 41.5), and less educated (0.31 against 0.47 Bachelor's or higher) than the dropped records, which concentrate in higher-skill occupations the Anthropic and Copilot releases cover at six-digit SOC. Table 3 reports sample sizes alongside coefficients to differentiate coverage-driven from user-base-driven dispersion.

The cross-domain test in Section IV uses ten occupation-level outcomes, five from the ACS 2011 to 2022 (hours worked, three education shares, and disability prevalence) and five from the National Health Interview Survey (NHIS) Sample Adult files for 2017 and 2018 (Kessler-6 distress, smoking, drinking, short sleep, and self-rated health). Employment, labor-force participation, and log weekly wages stay out of this set because they are the main DiD outcomes. Online Appendix B adds those three back for a fifteen-outcome family with full multiple-testing corrections, and Online Appendix B.1 documents the variables and crosswalks. Table 1 reports summary statistics for the analytical sample, the NEM-matched subsample, and the ten measures. Online Appendix Table B.1 gives variable definitions, and Online Appendix Table B.2 Panel B reports occupation-level means by SOC major group.

\subsection{C. Platform-Workforce Divergence}

The platform measures differ sharply from the workforce benchmark. At the six-digit SOC level, no platform-derived measure has a Spearman correlation with BLS employment shares above $\rho = 0.33$, while Anthropic waves correlate much more strongly with one another, with pairwise correlations ranging from 0.84 to 0.98. The platform measures also differ from the worker-reported RPS benchmark, with no correlation above $\rho = 0.41$. Figure 1 summarizes these pairwise correlations. The pattern is consistent with the framework in Section I: platform-derived measures are closer to one another than to a workforce population frame.

The occupational composition of the gap is systematic. Table 2 reports $\psi_{o,p}$ by SOC major group across platforms (Online Appendix Table B.2 Panel A reports the full per-platform composition by SOC major group). Computer and Mathematical, Office and Administrative Support, and Business and Financial Operations occupations are overrepresented, while Transportation, Production, Construction, and Food Preparation occupations are underrepresented. This matters because the overrepresented groups are also the groups that AI capability rubrics tend to rate as highly applicable. Platform data therefore place greater weight on occupations where the rubric already predicts high exposure, which is precisely the covariance between platform selection and task applicability that generates non-classical measurement error.

The same user-base imbalance changes the exposure profile itself. Figure 2 compares the \hyperlink{Massenkoff2026}{Massenkoff and McCrory (2026)} composite with the \hyperlink{Eloundou2024}{Eloundou et al.\ (2024)} capability rubric across the 22 SOC major groups. The consumer composite tracks the rubric closely for Computer and Mathematical, Business and Financial, and Office and Administrative Support occupations, but falls below the rubric for Personal Care, Food Preparation, and Transportation, where consumer platform users are sparse. The enterprise API tilts further toward Computer and Mathematical occupations, which account for more than half of its conversations. Copilot remains closer to the rubric because its user base is closer to the workforce. Across panels, the exposure profile departs more from the rubric when the platform's users depart more from the workforce.

These comparisons establish the first empirical implication of the framework. Platform-derived measures are not alternative noisy estimates of the same workforce exposure profile. They embed occupation-specific platform representation. Downstream estimates based on these measures therefore inherit a population choice that is usually left implicit.

\subsection{D. Temporal Instability Within a Platform}

The divergence from the workforce is not only a fixed difference across vendors. It also changes within a vendor over time, which means that vendor fixed effects cannot absorb the user-base component when the composition of users shifts across waves.

The Anthropic Economic Index waves appear stable when summarized by rank correlations. Pairwise Spearman correlations range from 0.92 to 0.98 for adjacent waves and from 0.84 to 0.90 for non-adjacent waves. Even Wave 1 and Wave 5, which are fourteen months apart, have a correlation of $\rho$ = 0.85. Yet this aggregate stability masks substantial occupational movement. Between Wave 1 and Wave 5, 41.8 percent of occupations move to a different exposure quartile, and 4.4 percent move by two or more quartiles (Online Appendix Figure B.2 traces SOC major-group rank trajectories across waves, and Online Appendix Table B.4 reports the wave-pair quartile transitions).

The timing of these shifts points to changes in the user base rather than gradual improvements in model capability. Claude Code launched between Waves 2 and 3, and Claude Projects expanded between Waves 3 and 4. Over the same period, the Computer and Mathematical conversation share rose from 28 percent in Wave 1 to 36 percent in Wave 3 before settling at 32 percent in Wave 5, while Office and Administrative Support moved 6 percentage points in the opposite direction. These movements are more naturally interpreted as changes in who used the platform and what they used it for than as changes in the underlying capability rubric. They nevertheless enter the exposure scores that downstream researchers often treat as fixed occupational measures.

The implication is that a researcher using Wave 1 obtains a different occupational ranking than a researcher using Wave 5, even though both rely on the same vendor. We interpret this within-platform movement as a special case of equation (3), where $\psi_{o,p}$ varies across waves. The movement combines the $\psi$ channel, which captures which occupations enter the platform user base, and the $\theta$ channel, which captures which tasks users perform within occupations. Section III.F separates these channels empirically, and Appendix Table B.11 reports diagnostics on within-occupation task-share stability.

\subsection{E. Workforce Reweighting and Occupational Rankings}

If platform representation is the source of the divergence, removing it should change which occupations appear most exposed. The reweighted exposure measure, $\widetilde{E}_{o,p} \equiv E_{o,p}/\psi_{o,p}$, rescales each occupation from its platform share to its workforce share, as defined in Section I. For the composite measure, the correction preserves the rubric and task-time content but evaluates that content over the workforce rather than over platform users. For a pure platform-share measure, the same correction collapses the measure to workforce employment density.

The raw and reweighted composites correlate at 0.78 across six-digit SOC codes, suggesting moderate agreement in the full ranking. The economically relevant changes, however, occur at the top of the distribution. The raw ranking places statistical assistants, computer programmers, and actuaries among the most exposed occupations. After reweighting, the top occupations become customer service representatives, office clerks, and cashiers, which are large occupations in the workforce but relatively rare in platform conversations. Online Appendix Figure B.5 traces these movements occupation by occupation. Statistical Assistants fall 340 ranks after reweighting, while Customer Service Representatives rise 145 ranks and Office Clerks rise 132 ranks.

The rank reversals change the interpretation of the exposure measure. The raw ranking concentrates AI exposure in high-skill, high-wage work, while the reweighted ranking shifts exposure toward large, mid-skill service and administrative occupations. The two rankings are not noisy versions of the same object. The raw measure describes exposure among platform users, while the reweighted measure describes exposure in the workforce. Downstream studies that rely on a single platform-derived ranking implicitly choose between these populations.

The measurement evidence establishes that platform-derived exposure differs from workforce exposure in levels, over time, and after reweighting. Section III asks whether these differences are large enough to change downstream labor-market coefficients.

\section{III. Downstream Estimates and Bounds}

We estimate a standard difference-in-differences specification and vary only the exposure measure. We then isolate the user-base channel using within-vendor channel substitution and event-study timing. We apply workforce reweighting and decompose cross-wave variation into the between-occupation and within-occupation channels. We return to the published cross-occupation specification and construct partial-identification bounds.

\subsection{A. Difference-in-Differences Specification}

We use a DiD specification common in the AI-and-labor literature as a diagnostic for measurement instability. Equation (7) is defined at the occupation level. For person $i$ working in occupation $o$, state $s$, and year $t$, the DiD specification is

\begin{equation}
y_{i,o,s,t} = \alpha_{o} + \gamma_{s} + \delta_{t} + \beta \cdot (E_{o}^{(p)} \times \text{Post}_{t}) + X_{i,o,s,t}'\theta + \varepsilon_{i,o,s,t} \tag{7}
\end{equation}

where $y_{i,o,s,t}$ is the labor-market outcome (binary employment in the main specification), $E_{o}^{(p)}$ is the platform-derived exposure measure for occupation $o$ under platform $p$, ${Post}_{t}$ equals one for 2023 and 2024 with 2022 treated as a pre-period year, $\alpha_{o}$ is a six-digit SOC fixed effect, $\gamma_{s}$ is a state fixed effect, $\delta_{t}$ is a year fixed effect, and $X_{i,o,s,t}$ includes the demographic controls. Regressions are weighted by ACS person weights and standard errors are clustered at the state level.

Equation (4) gives the probability limit for the DiD coefficient and for the level regression in (1), because $E_{o}^{(p)} \times {Post}_{t} = \psi_{o}(E_{o} \times {Post}_{t}) + (\eta_{o} + u_{o}) \times {Post}_{t}$ and the scaling by $\psi_{o}$ is the same in both. Parallel trends are not needed for the dispersion to be informative. Cross-platform variation in the post-2022 coefficient and in pre-period behavior both trace to the user-base channel, and both show that the regressor is not a fixed occupational characteristic.

The event study replaces the post indicator with year-by-treatment interactions,

\begin{equation}
y_{i,o,s,t} = \alpha_{o} + \gamma_{s} + \delta_{t} + \sum_{k \neq 2022}\beta_{k}\,(E_{o,p} \times \mathbf{1}\{t = k\}) + X_{i,o,s,t}'\theta + \varepsilon_{i,o,s,t} \tag{8}
\end{equation}

with 2022 as the reference year. The pre-period coefficients test for differential trends across high- and low-exposure occupations through a Granger-style joint Wald $F$-test, and the post-period coefficients trace the timing of the post-2022 differential.

To make coefficients comparable, we z-score each $E_{o}^{(p)}$ within the analytic sample using ACS person weights before estimation. Each measure then runs through the same sample, controls, and estimator, and the only thing that changes across columns is the per-occupation exposure score.

\subsection{B. Cross-Platform Dispersion}

The downstream coefficients inherit the platform source used to construct exposure. Table 3 reports the post-2022 coefficients for employment, labor-force participation, and log weekly wages, with one column for each exposure measure (Online Appendix Table B.6 reports the full coefficient grid). The ten employment coefficients are uniformly negative, but their magnitudes differ by a factor of 1.9. Under the composite, a one-standard-deviation increase in exposure lowers employment by 0.14 percentage points, near the per-thousand-robots effect in \hyperlink{Acemoglu2020}{Acemoglu and Restrepo (2020)} and about one quarter of the youth-employment effect in \hyperlink{Brynjolfsson2025}{Brynjolfsson, Chandar, and Chen (2025)}. The estimate is economically plausible. The problem is that its magnitude depends on which platform supplies the conversation shares.

The same pattern appears within the Claude.ai consumer channel over time. The coefficient rises monotonically across the five waves, most sharply between Waves 3 and 5, when Claude Code adoption accelerated and the enterprise base expanded. The Wave 5 estimate of 0.22 percentage points is nearly twice the Wave 1 estimate of 0.12 percentage points, although the regressions differ only in the date of the conversation data. In the framework of Section I, the structural coefficient and true exposure are held fixed across these regressions. The coefficient moves because the slope of the platform proxy on true exposure changes with the user base.

This dispersion is not sampling noise. A Cochran Q test rejects a common coefficient across the ten measures on labor-force participation and wages, and nearly so on employment. Pairwise contrasts separate the platforms at the extremes. The measures also bear little relation to what workers themselves report. When we correlate each against the at-work AI use rate in the \hyperlink{Bick2026}{Bick, Blandin, and Deming (2026)} survey, the correlation stays low throughout. Even the platform closest to the workforce stands well apart from reported use, and farther from true exposure.

\subsection{C. Within-Vendor Channel Substitution and Event-Study Timing }

To isolate the mechanism, we compare two channels of the same platform. The Claude.ai consumer and enterprise API channels run on the same model in the same wave and differ only in which workers reach Claude through each. Table 3 Panel B reports the comparison across Waves 3 through 5 under the composite. At every observed wave, the consumer and enterprise coefficients carry opposite signs. The consumer channel produces negative and significant coefficients, while the enterprise channel produces positive and insignificant coefficients.

This within-vendor sign reversal is difficult to attribute to chat-to-task classification error. Both measures use the same Clio classifier applied to the same Claude model family and rubric. Any over-attribution in mapping conversations to tasks, including the type of error emphasized by \hyperlink{BBDS2026}{Bick, Blandin, Deming, and Schumacher (2026)}, should affect the consumer and enterprise measures similarly. The sign reversal therefore points to the input the two channels do not share: their user base. We trace the difference to occupational composition, since the enterprise channel concentrates more heavily in occupations rated as highly applicable by the \hyperlink{Eloundou2024}{Eloundou et al. (2024)} rubric, and the downstream coefficient inherits that composition.

The timing of the coefficients helps distinguish measurement instability from treatment-effect heterogeneity. A competing interpretation is that the dispersion reflects real differences in how AI affected the occupations emphasized by different platforms. Figure 3 weighs against that interpretation. The year-by-year interaction coefficients from equation (8) are close to zero between 2015 and 2020 across the seven exposure variants. Granger-style joint Wald tests of the pre-treatment interactions yield F-statistics ranging from 0.34 to 2.18, with the largest p-value at 0.21. None of the seven specifications rejects parallel trends at the 5 percent level. The exposure variants then separate sharply in 2023 and remain apart thereafter.

The timing is informative because genuine post-ChatGPT treatment heterogeneity would be expected to emerge as workers, firms, and platform users adjusted to the new technology. Instead, the separation appears at the onset of the post period and persists. That pattern is more consistent with measurement bias loading on time-invariant differences in platform user composition than with treatment effects that evolve gradually after 2022. We therefore interpret the event study as a diagnostic for the user-base channel, not as evidence that cross-measure dispersion is innocuous.

The framework also predicts which variants should exhibit clean pre-period behavior. When $\psi_{o,p}$ is stable, the pre-period coefficients should be close to zero. This is true by construction for the workforce-reweighted composite and empirically for adjacent Anthropic waves, whose cross-wave rank correlations exceed 0.95. By contrast, when $\psi_{o,p}$ changes during the pre-period as in variants spanning the Claude Code launch or similar product transitions, the pre-period coefficients can differ from zero. We interpret the asymmetric rejection in the full ten-measure test reported in Section V and Table B.8 as evidence of this mechanism.

\subsection{D. Subgroup Heterogeneity}

Online Appendix Figure B.6 reports DiD coefficients for fifteen demographic and occupation subgroups under five platform-derived exposure measures, totaling 75 cells (Online Appendix Table B.7 reports the same coefficients in tabular form). The composite produces a statistically significant negative coefficient in nine of fifteen subgroups, the Claude.ai consumer channel produces eight significant negatives, and the enterprise API channel produces three significant positives and four significant negatives. Sign disagreement across measures is most pronounced for three subgroups, with the coefficient on workers with disabilities positive under Claude.ai consumer and negative under the composite, the coefficient on workers under age 25 positive under the composite and negative under Microsoft Copilot, and the coefficient on female workers positive under the enterprise API and negative under all other measures.

The subgroup results matter because the applied literature often conditions on a single platform-derived measure without testing sensitivity to platform choice. The 75-cell grid shows that subgroup contrasts can be partly driven by the platform from which exposure is constructed. A researcher studying workers with disabilities, for example, would reach opposite conclusions under the composite and Claude.ai consumer measures. This is the same fragility documented in the aggregate estimates, projected onto subgroup comparisons.

The subgroup pattern raises a second concern distinct from coefficient dispersion. Platform conversation data observes only workers who use the platform, and workers who do not use the platform are invisible to the data, with the DiD coefficient for a given subgroup depending on two objects, the true effect of AI on that subgroup and the subgroup's representation in the platform user base. Subgroups less visible to the platform enter the regression through their workforce share rather than through their actual AI-use patterns, and their estimated coefficient inherits the user-base composition of the platform that supplied the exposure score. The disagreement observed in the disability, age-under-25, and female rows is therefore consistent with two mechanisms operating at once that the regression cannot separate, true heterogeneity in the AI-employment relationship across subgroups and differential platform visibility across the same subgroups.

The distinction matters for policy interpretation because the platform-visible workforce overrepresents workers with higher income, more education, urban residence, and non-Hispanic ethnicity, and underrepresents the corresponding disadvantaged subpopulations (Online Appendix Figure B.3 plots occupations along the joint dimensions of economic vulnerability and platform visibility). A subgroup that is both economically disadvantaged and underrepresented on the platform sits in the quadrant of Figure B.3 where the DiD coefficient is least informative about the true effect, because the platform's view of that subgroup is constructed from a small and likely non-representative sample of its members. Policy rankings that condition on a single platform-derived exposure will therefore systematically underweight the populations whose AI-related labor-market risks the rankings are intended to identify.

\subsection{E. Workforce Reweighting in the DiD Specification}

Section I implies that reweighting platform conversation density to match workforce density addresses the between-occupation component of the bias. Table 3 reports this correction for two of the ten exposure variants. For the composite measure, the coefficient falls from -0.139, significant at the 1 percent level, to -0.010 and statistically insignificant. This is a 93 percent attenuation. For Microsoft Copilot, the coefficient falls from -0.191 to -0.110. The attenuation is smaller, at 42 percent, and the estimate remains statistically significant.

The difference between the two corrections is informative. Copilot is the platform-derived measure closest to workforce employment, with a correlation of $\rho$ = 0.33, and its coefficient moves less after reweighting. The composite lies farther from the workforce and attenuates more. This ordering is exactly what the framework predicts. The farther the platform's users are from the workforce, the more of the coefficient should be attributable to the between-occupation user-base component.

We use the baseline and reweighted composite estimates to bound the employment elasticity in the DiD setting. A one-standard-deviation increase in exposure is associated with an employment decline between 0.01 and 0.14 percentage points under the maintained ordering condition. The width of the interval measures the contribution of observable platform composition to the baseline coefficient. In this case, the baseline estimate is almost entirely driven by the component removed by workforce reweighting.

\subsection{F. Decomposing Cross-Wave Variation into the \texorpdfstring{$\psi$}{psi} and \texorpdfstring{$\theta$}{theta} Channels}

The within-platform movement can arise from two sources. Product changes may alter which occupations enter the platform user base, captured by $\psi$, or they may alter which tasks users perform within occupations, captured by $\theta$. Reweighting separates these channels because it removes the between-occupation component while leaving any within-occupation task-selection residual.

We implement this decomposition by taking each Anthropic Economic Index wave in turn, substituting that wave's conversation share into the \hyperlink{Massenkoff2026}{Massenkoff and McCrory (2026)} composite, and re-estimating the DiD models for employment, labor-force participation, and log wages. We then repeat the exercise after reweighting the composite to match workforce density. Any cross-wave variation that disappears after reweighting is attributable to shifts in occupational representation. Any variation that remains reflects within-occupation task selection or residual classification differences.

Table 4 reports the decomposition. Reweighting closes 51 percent of the cross-wave span in the employment coefficient, 15 percent in labor-force participation, and 32 percent in log wages. The remaining variation is the part not explained by shifts in occupational representation. This residual is consistent with within-occupation task selection and classification differences that reweighting cannot remove. Appendix Table B.11 provides diagnostics on within-occupation task-share stability.

The between-occupation channel is also visible in worker-reported use. The framework predicts that occupations overrepresented in platform data should be occupations where workers report greater at-work AI use. The RPS microdata support this prediction. Across Claude.ai consumer, Claude enterprise API, and Microsoft Copilot, the correlation between platform representation and survey-reported at-work AI use is positive and statistically significant, ranging from 0.25 to 0.39 (Online Appendix Figure B.7 plots the three measures across SOC major groups). Platform data therefore overweight occupations where AI use is already more common. This test does not address the within-occupation version of the maintained ordering condition, but it confirms that between-occupation composition is empirically relevant.

\subsection{G. Cross-Occupation Estimates}

The same user-base channel appears in the cross-occupation specification for which the composite measure was originally proposed. \hyperlink{Massenkoff2026}{Massenkoff and McCrory (2026)} regress each occupation's BLS-projected employment growth from 2024 to 2034 on standardized composite exposure, weighted by 2024 OEWS employment. The coefficient is the change in projected decadal employment growth associated with a one-standard-deviation increase in exposure. This setting is useful because the dependent variable, weights, and occupational universe can be held fixed while the exposure measure is reweighted or replaced.

In the published estimate, a one-standard-deviation increase in composite exposure is associated with projected decadal employment growth lower by 0.70 percentage points. We apply the workforce-reweighting operation directly to the composite, leaving the dependent variable, weights, and occupational universe fixed. Figure 4 Panel A shows that the employment-growth differential falls from 0.70 to 0.38 percentage points after reweighting. The difference of 0.32 percentage points means that 45 percent of the published magnitude disappears when platform conversation density is replaced with workforce density.

The redistribution across major groups follows the same logic. Figure 4 Panel B shows that the largest downward shifts under reweighting occur in Computer and Mathematical, Office and Administrative Support, and Sales occupations, the groups most overrepresented in the platform user base. Reweighting therefore reduces exposure most where platform representation departs most from workforce representation. The cross-occupation attenuation of 45 percent falls within the DiD reweighting range of 42 to 93 percent, indicating that the same channel appears in both empirical settings.

We next substitute alternative platform-derived measures into the same between-occupation regression. Table 5 Panel B reports estimates from the ten regression measures and two benchmark distributions (Online Appendix Table B.10 reports the parallel substitution across five published applied-AI target methods). The dependent variable, employment weights, and occupational universe remain fixed. Only the occupation-level exposure score changes. The substitution moves the estimate across both signs. A one-standard-deviation increase in exposure is associated with projected decadal employment growth lower by 1.50 percentage points under Microsoft Copilot and higher by 1.74 percentage points under OpenAI ChatGPT, with the Anthropic measures spanning both signs across waves. This range does not imply that the structural employment effect of AI differs across platforms. It shows that the same cross-occupation regression inherits the occupational composition of the platform used to construct exposure.

We also re-estimate the between-occupation regression using the RPS measure of at-work AI use at the six-digit SOC level. A one-standard-deviation increase in survey-reported at-work AI use is associated with projected decadal employment growth higher by 1.67 percentage points, with a standard error of 0.46 percentage points. This estimate lies near the positive end of the platform-derived range, close to the OpenAI ChatGPT estimate and opposite in sign to the published composite estimate. The mixed signs therefore do not arise only from platform logs. They also appear when the right-hand-side variable is a survey-based measure of at-work AI use. Table 5 Panel B and Appendix Table B.9 report the RPS estimate alongside the platform-derived measures and the SOC-major-expanded SWAA row.

\subsection{H. Partial-Identification Bounds}

We use the baseline composite coefficient and its workforce-reweighted counterpart to construct a partial-identification interval for the structural employment elasticity. The interval relies on the maintained ordering condition stated in Section I and proved in Appendix A.5. Under this condition, the baseline composite contains both task exposure and platform selection, while the workforce-reweighted composite removes the between-occupation selection component. The interval implies that a one-standard-deviation increase in composite exposure is associated with projected decadal employment growth lower by between 0.38 and 0.70 percentage points. Its width is 0.32 percentage points. This width is the estimated contribution of between-occupation platform selection to the published coefficient. Equivalently, nearly half of the published association disappears when the platform's occupational composition is replaced with the workforce composition.

This changes the interpretation of the published estimate. Reporting only the baseline coefficient treats the 0.70 percentage point decline as a point estimate of the structural employment elasticity. The reweighting exercise shows that the same specification supports a smaller decline once the platform's occupational density is replaced with the workforce density. The data therefore identify a range of elasticities under the maintained ordering condition rather than a single platform-specific coefficient.

The cross-platform substitutes in Table 5 Panels B and C serve a different purpose. They show how far the between-occupation estimate moves when the platform source changes. The estimates range from projected decadal employment growth lower by 1.50 percentage points under Microsoft Copilot to growth higher by 1.74 percentage points under OpenAI ChatGPT. The cross-platform span is 3.24 percentage points, which is several times larger than the published composite estimate itself.

We do not interpret this span as a partial-identification interval because the substitutes are not the same object as the composite. They enter as raw conversation densities rather than rubric-weighted task aggregates, and their reweighted versions no longer measure AI applicability. We report the span as descriptive evidence that the between-occupation specification is highly sensitive to the platform source used to construct exposure.

The DiD and cross-occupation exercises deliver the same substantive conclusion. In both settings, the workforce-reweighted estimate reduces the component of the coefficient driven by between-occupation platform selection. The baseline estimate, by contrast, depends on which platform supplies the occupational exposure measure. The relevant empirical object is therefore not a single coefficient from one platform-derived regressor. It is the range of estimates that remains once platform selection is made explicit.

This framing follows the taxonomy of \hyperlink{Tamer2010}{Tamer (2010)}. The bounds are not outcome bounds generated by limited sample size. They are selection-mechanism bounds generated by the unobserved platform-selection process. The interval collapses only with an external source of variation that breaks the dependence between true exposure, platform representation, and within-occupation task selection. In the absence of such an instrument, the interval width has a direct economic interpretation. It is the part of the projected employment-growth association that disappears when the platform's occupational composition is replaced with the workforce composition.

\section{IV. Distributional Incidence}

The results in Section III show that platform selection changes downstream employment estimates. This section asks where that selection falls in the workforce. If platform representation is correlated with occupation-level socioeconomic status, then platform-derived exposure can transmit the user-base channel into outcomes beyond employment and into policy rankings that depend on measured exposure. We test this implication in three steps. We first compare platform and workforce rankings against additional occupation-level outcomes. We then describe the demographic composition of the platform-visible workforce. We finally examine how population weighting changes a retraining allocation based on the same exposure concept.

\subsection{A. Cross-Domain Outcomes}

If platform visibility is correlated with occupational socioeconomic status, the user-base channel should appear outside the employment regressions. We test this implication using ten occupation-level outcomes from the ACS and the National Health Interview Survey. The ACS outcomes cover 2011 to 2022 and include hours worked, three education shares, and disability prevalence. The NHIS Sample Adult outcomes cover 2017 and 2018 and include the Kessler-6 distress score, current-smoker prevalence, drinking behavior, short-sleep prevalence, and self-rated health.

The diagnostic compares two occupation rankings. The first ranks occupations by Anthropic Wave 5 Claude.ai consumer conversation share. The second ranks occupations by OEWS workforce employment share. For each outcome, we compute how strongly the outcome is associated with each ranking and examine the difference. If platform representation were unrelated to occupational socioeconomic status, the platform ranking and the workforce ranking would have similar relationships with wages, education, health, and distress. A systematic difference means that the platform ranking carries information about the composition of the occupations it overrepresents.

When the three main labor-market outcomes are added to this set, five of the fifteen outcomes show nominally significant platform-versus-workforce gaps at the 5 percent level. These outcomes are log real hourly wages, Bachelor's-or-higher share, short-sleep prevalence, current-smoker prevalence, and K6 severe-distress prevalence. The economic pattern is the important result. The Anthropic consumer ranking is more aligned than the workforce ranking with high-wage and high-education occupations. It is less aligned with occupations that have higher smoking rates, more short sleep, and higher severe distress. The exact gap statistics are largest for log wages, short sleep, current smoking, Bachelor's-or-higher share, and severe distress, with all five signs in the direction predicted by platform selection.

This pattern means that the platform ranking does not only order occupations by measured AI exposure. It also orders them along a socioeconomic gradient. Occupations that are more visible on the platform are more advantaged on average, while occupations that are less visible on the platform are more likely to have lower wages and worse health-related outcomes. This is consistent with the platform overrepresenting Computer and Mathematical occupations and underrepresenting service and manual-labor occupations.

We interpret these results as evidence on the pattern of selection rather than as five separate outcome-specific findings. Because we test fifteen outcomes, the individual p-values should not be read as independent confirmation for each outcome. Online Appendix Figure B.1 reports the family-wise and false-discovery-rate adjustments. Under the most conservative interpretation, no single outcome should carry the argument on its own. The relevant evidence is that the outcomes most tied to occupational socioeconomic status move in the predicted direction. Wages and education are more strongly aligned with the platform ranking, while smoking, short sleep, and severe distress are more strongly aligned with occupations underrepresented on the platform.

\subsection{B. Composition of the Platform-Visible Workforce}

We next ask which workers are represented when the platform conversation distribution is mapped onto the workforce. We construct the platform-visible workforce using the Anthropic Wave 5 Claude.ai consumer conversation density at the six-digit SOC level and the 2024 OEWS employment distribution within occupations. This construction combines the demographic composition of workers within each occupation with the platform's between-occupation density. We then compare the resulting population with the BLS workforce using ACS 2024 demographics.

The platform-visible workforce differs systematically from the U.S. workforce. It overrepresents workers in higher-wage occupations and workers with Bachelor's-or-higher degrees. It underrepresents workers in occupations with high disability prevalence and high non-Hispanic Black workforce representation (Online Appendix Figure B.4 reports the full demographic profile of the 95 most-populous SOCs). The direction of these differences matches the cross-domain evidence. The platform places greater weight on occupations that are more advantaged in the workforce and less weight on occupations that are more economically vulnerable.

This composition clarifies why the cross-domain gaps arise. The platform-visible workforce is tilted toward high-wage and high-education occupations, and away from lower-wage, disability-prevalent, and manual-labor occupations. The outcomes with nominally significant platform-workforce gaps are precisely the outcomes tied to this occupational divide. They include wages, education, health behavior, mental distress, and sleep. The implication for interpretation is direct. A platform-derived exposure measure is least informative for workers in occupations that are both weakly represented on the platform and more economically vulnerable in the workforce.

Online Appendix Figure B.3 plots occupations by economic vulnerability and platform visibility. The high-vulnerability, low-visibility quadrant is the key region for interpretation. These occupations are economically exposed in the workforce but weakly represented in platform data. For them, the platform-derived coefficient is least informative about the true effect because the platform observes few workers and observes them selectively. This quadrant disproportionately includes personal-care, food-preparation, and building-maintenance occupations.

\subsection{C. Allocation Implications}

The distributional consequences become concrete when exposure rankings are used for allocation. Online Appendix C reports a hypothetical ten-billion-dollar retraining-fund simulation (Figure C.1). The exercise is not a welfare calculation. It holds the budget fixed and asks how the allocation changes when the same exposure concept is weighted by platform visibility rather than workforce representation.

The platform-weighted rule directs 3.87 billion dollars to occupations with above-median wages and Bachelor's-or-higher shares above 60 percent. Under the workforce-weighted rule, the same dollars shift toward middle-wage occupations whose Bachelor's-or-higher shares are closer to the workforce mean. The difference between the two rules is therefore not only a technical change in weights. It changes which workers are treated as exposed. Platform weighting shifts resources toward occupations already visible in AI data, while workforce weighting shifts resources toward larger workforce occupations that are less represented in platform conversations.

The exercise illustrates the policy relevance of the measurement problem. Exposure rankings are often interpreted as rankings of labor-market risk. When those rankings are built from selected platform use, low platform visibility can be mistaken for low exposure. Yet low visibility may reflect limited access, low adoption, or the absence of workers whose tasks have already been displaced. For policy purposes, the relevant population is not only current platform users. It is the workforce facing adjustment costs as AI changes task demand.

\subsection{D. Implications for Cross-Domain Applied Work}

The cross-domain and composition results imply that platform-derived AI exposure can transmit user-base selection into any occupation-level outcome correlated with socioeconomic status. This concern extends beyond employment. Wages, education, disability prevalence, mental health, sleep, and smoking all vary systematically across occupations. Several of these outcomes also align differently with platform-based and workforce-based rankings.

The implication for applied work is methodological. A regression that treats platform-derived exposure as a fixed occupational characteristic may recover part of the platform user-base channel rather than only the structural relationship between AI exposure and the outcome. The concern is strongest when the outcome is correlated with $\psi_{o,p}$, since the exposure measure then embeds both the platform's occupational composition and the outcome's socioeconomic gradient.

The workforce-reweighting and partial-identification procedure extends directly to this setting. For any downstream outcome, researchers can report the baseline coefficient and the workforce-reweighted coefficient as endpoints of an identified set under the maintained ordering condition. The width of the interval gives the contribution of the user-base channel for that outcome and specification. This procedure does not solve the within-occupation task-selection problem, but it separates the component of bias that is observable from workforce and platform occupation shares.

\section{V. Robustness}

The preceding sections show that platform-derived exposure measures differ from workforce exposure and that this difference carries into downstream coefficients. We now ask whether the coefficient dispersion can be explained by standard specification choices rather than by the user-base channel. Across the exercises, we hold fixed the analytic sample and the ten exposure measures, then vary one feature of the empirical design. The statistic we track is the spread in the post-2022 coefficient across exposure measures. In the baseline employment specification, the largest absolute coefficient is 1.9 times the smallest.

A natural concern is that the post-2022 dispersion reflects differential pre-period trends rather than platform selection. We estimate the event-study specification in equation (8) for each exposure measure and test the seven pre-period interactions from 2015 through 2021, with 2022 omitted. Six of the ten measures reject the zero-pre-trend null at the 5 percent level. This pattern is consistent with the framework. The exposure regressor partly reflects the platform user base, and platform representation is correlated with occupation-level trends that predate the broad diffusion of generative AI. The relevant question is whether the cross-platform dispersion depends on the pandemic years or on the choice of 2022 as the reference year. It does not. When we drop 2020 and 2021 and normalize the event study to 2019, the dispersion ratio is 1.8, nearly identical to the baseline ratio of 1.9.

The dispersion also appears beyond the employment margin. Table 3 reports the same exercise for employment, labor-force participation, and log weekly wages. The cross-platform range is 1.9 for employment, 2.5 for labor-force participation, and 3.7 for log weekly wages. These margins need not move in the same direction, since AI exposure can affect employment, participation, and wages through different adjustment mechanisms. The relevant result is that platform choice generates substantial dispersion for each outcome. The instability is therefore attached to the exposure measure rather than to a particular labor-market outcome.

One might also worry that the full ACS sample combines workers whose platform visibility and labor-market responses differ in ways that mechanically widen the coefficient range. We address this by re-estimating the specification for prime-age workers, non-Hispanic white workers, and women. The dispersion ratios are 1.8, 2.1, and 2.0, close to the full-sample ratio of 1.9. These subsamples change the demographic composition of the estimating sample while leaving the exposure measures unchanged. The dispersion remains, and it is at least as large in groups closer to the platform-visible workforce. That pattern is consistent with the user-base channel rather than with an idiosyncrasy of the full sample.

A further possibility is that the coefficient range reflects the way the regression absorbs occupation, geography, and demographics. We estimate five versions of the model. The baseline includes year, state, and six-digit occupation fixed effects with demographic controls. We then drop demographic controls, drop state fixed effects, replace six-digit occupation fixed effects with two-digit SOC major-group fixed effects, and add state-by-year interactions. Across these specifications, the dispersion ratio ranges from 1.7 to 2.1. The levels of the estimates change, as expected when the identifying comparisons change. The cross-platform spread does not. The result is not an artifact of one control set or one fixed-effect structure.

The same logic applies to the capability rubric embedded in the composite. If the instability were driven mainly by the \hyperlink{Eloundou2024}{Eloundou et al. (2024)} rubric, replacing that rubric should materially reduce the cross-platform range. We test this by substituting the \hyperlink{Felten2021}{Felten, Raj, and Seamans (2021)} AI occupational-impact score, which is constructed differently and has a rank correlation of 0.74 with the Eloundou rubric across occupations. The cross-platform dispersion ratio remains 1.8. Changing the capability measure changes the level of exposure, but it does not remove the platform-composition component embedded in the conversation shares.

Because our preferred specification clusters standard errors by state, we also check whether inference with a moderate number of clusters affects the conclusions. The baseline specification has 51 clusters, including the District of Columbia. We re-estimate the preferred specification using wild-cluster bootstrap confidence intervals with 999 replications. The confidence intervals for the composite and the Anthropic Wave 5 Claude.ai consumer coefficient are substantively unchanged. The within-Anthropic sign disagreement is therefore not a product of conventional cluster-robust inference.

The remaining concern is that the within-Anthropic time pattern may reflect model changes rather than changes in the user base. Adjacent-wave comparisons help separate these forces. The Wave 4 to Wave 5 transition is close to a model-refresh comparison. It spans three months, has no major product launch, and leaves the occupational ranking almost unchanged, with a rank correlation of 0.98. The employment coefficient changes by only 0.008. The Wave 2 to Wave 3 transition is different. It coincides with the Claude Code launch and the Sonnet 4 update, a change likely to shift both capabilities and the occupations using the platform. The occupational ranking becomes less stable, with the rank correlation falling to 0.92, and the coefficient changes by 0.031. The economic magnitude is therefore about four times larger when the transition includes a product change likely to alter the user base. This comparison does not rule out a model contribution, but it shows that most of the within-platform movement occurs when the platform's occupational composition is most likely to change.

Taken together, the robustness exercises narrow the set of plausible explanations (Online Appendix Table B.5 summarizes significance counts across the seven dimensions). The cross-platform dispersion remains when we change the reference year, the outcome, the sample, the controls, the fixed effects, the capability rubric, and the inference method. It also remains after bounding the role of adjacent model changes. The common element across the exercises is the platform-derived exposure measure. The evidence therefore supports the interpretation that coefficient dispersion is a feature of platform-based measurement rather than a consequence of a single regression choice.

\section{VI. Conclusion}

Platform conversation logs have moved occupational AI exposure measures closer to observed use, but the move changes the interpretation of the measure. A measure built from platform conversations does more than summarize which tasks AI can perform. It also weights those tasks by the occupations that appear on a particular platform. When platform representation is correlated with occupational task structure, the resulting error is non-classical and the coefficient inherits the platform source used to construct exposure, even when the regression design is held fixed.

The framework separates what current data can correct from what remains unidentified. The between-occupation gap between platform conversation shares and workforce employment shares is observed, and workforce reweighting removes that component under the maintained ordering condition. The remaining within-occupation task-selection residual cannot be removed without worker-task data measured on a common taxonomy. The baseline and workforce-reweighted coefficients therefore define a partial-identification interval for the structural employment elasticity. The interval is not a causal estimate. Its role is to show how much of the coefficient is supported once the observable platform-composition component is removed.

The same logic clarifies why observed use should not be interpreted as displacement. Platform conversations are generated by workers and tasks that remain visible in the occupation-platform pair. They do not recover nonusers, infrequent users, or workers and tasks that leave that pair. Reweighting corrects occupational composition, but it does not recover the share of worker task time performed by AI, the productivity of that use, or the labor-demand response when task costs fall. Employment and wage effects therefore require adoption and use measures in a workforce frame, together with credible variation in exposure or adoption.

The policy implication is not that exposure rankings should be discarded. It is that rankings built from selected platform use should not be treated as rankings of workforce risk. Retraining funds, career advising, curriculum investment, and workforce planning all require judgments about where adjustment costs are likely to arise. If platform visibility is correlated with wages, education, and occupational advantage, a platform-weighted ranking can direct attention toward occupations where AI use is easy to observe and away from occupations where risk is less visible in platform data. For these uses, the relevant object is exposure among the workforce at risk, not exposure among current platform users.

The value of platform logs is that they observe task use at a scale and frequency unavailable in surveys or administrative records. The limitation is that scale does not change the population represented in the data. Additional conversations can make a platform-specific exposure measure more precise without making it workforce representative. The next empirical step is therefore to link platform logs to worker-representative adoption measures, task-time data, and research designs that distinguish exposure, adoption, use, and displacement. Without that link, estimates of AI exposure may continue to measure the platform's occupational composition as much as the labor-market exposure the literature seeks to explain.

\clearpage
\section*{Tables}
\addcontentsline{toc}{section}{Tables}
\refstepcounter{none}\label{tab:1}
\begin{landscape}
\begingroup
\scriptsize
\setlength{\tabcolsep}{2pt}
\renewcommand{\arraystretch}{0.78}
\renewcommand{\baselinestretch}{0.85}\selectfont
\textbf{Table 1. Analytical Sample and Exposure Measures}

\emph{\textbf{Panel A. American Community Survey analytic sample versus
full sample (2015 to 2024, ages 16 to 64)}}

{\def\LTcaptype{none} 
\begin{longtable}[]{@{}
  >{\raggedright\arraybackslash}p{0.3143\linewidth}
  >{\centering\arraybackslash}p{0.1143\linewidth}
  >{\centering\arraybackslash}p{0.1143\linewidth}
  >{\centering\arraybackslash}p{0.1143\linewidth}
  >{\centering\arraybackslash}p{0.1143\linewidth}
  >{\centering\arraybackslash}p{0.1143\linewidth}
  >{\centering\arraybackslash}p{0.1143\linewidth}@{}}
\toprule\noalign{}
\begin{minipage}[b]{\linewidth}\raggedright
\textbf{Variable}
\end{minipage} & \begin{minipage}[b]{\linewidth}\centering
\textbf{Analytic mean}
\end{minipage} & \begin{minipage}[b]{\linewidth}\centering
\textbf{Analytic SD}
\end{minipage} & \begin{minipage}[b]{\linewidth}\centering
\textbf{Full ACS mean}
\end{minipage} & \begin{minipage}[b]{\linewidth}\centering
\textbf{Full ACS SD}
\end{minipage} & \begin{minipage}[b]{\linewidth}\centering
\textbf{Diff}
\end{minipage} & \begin{minipage}[b]{\linewidth}\centering
\textbf{SMD}
\end{minipage} \\
\midrule\noalign{}
\endhead
\bottomrule\noalign{}
\endlastfoot
Employed & 0.844 & 0.363 & 0.844 & 0.363 & -0.000 & -0.000 \\
Age & 39.169 & 13.467 & 39.649 & 13.461 & -0.480 & -0.036 \\
Female & 0.507 & 0.500 & 0.5 & 0.5 & +0.007 & +0.014 \\
Bachelor\textquotesingle s or higher & 0.335 & 0.472 & 0.308 & 0.462 &
+0.027 & +0.057 \\
Married & 0.474 & 0.499 & 0.481 & 0.5 & -0.007 & -0.014 \\
Adults with disability & 0.075 & 0.264 & 0.076 & 0.265 & -0.001 &
-0.003 \\
Non-Hispanic white & 0.591 & 0.492 & 0.596 & 0.491 & -0.005 & -0.009 \\
Non-Hispanic Black & 0.125 & 0.331 & 0.115 & 0.319 & +0.010 & +0.031 \\
Hispanic & 0.182 & 0.386 & 0.187 & 0.39 & -0.005 & -0.013 \\
Number of observations & 13,104,072 & & 14,546,824 & & -1,442,752 &
-9.9\% \\
\end{longtable}
}

\needspace{16\baselineskip}
\emph{\textbf{
Panel B. Ten exposure-measure variants
(occupation-level)}}

{\def\LTcaptype{none} 
\begin{longtable}[]{@{}
  >{\raggedright\arraybackslash}p{0.4400\linewidth}
  >{\centering\arraybackslash}p{0.1400\linewidth}
  >{\centering\arraybackslash}p{0.1400\linewidth}
  >{\centering\arraybackslash}p{0.1400\linewidth}
  >{\centering\arraybackslash}p{0.1400\linewidth}@{}}
\toprule\noalign{}
\begin{minipage}[b]{\linewidth}\raggedright
\textbf{Variable}
\end{minipage} & \begin{minipage}[b]{\linewidth}\centering
\textbf{Mean}
\end{minipage} & \begin{minipage}[b]{\linewidth}\centering
\textbf{SD}
\end{minipage} & \begin{minipage}[b]{\linewidth}\centering
\textbf{Min}
\end{minipage} & \begin{minipage}[b]{\linewidth}\centering
\textbf{Max}
\end{minipage} \\
\midrule\noalign{}
\endfirsthead
\endhead
\bottomrule\noalign{}
\endlastfoot
Baseline Composite (Anthropic) & 0.076 & 0.133 & 0.000 & 0.745 \\
Composite Reweighted & 0.006 & 0.033 & 0.000 & 0.461 \\
Claude.ai Wave 1 & 0.100 & 0.328 & 0.000 & 5.912 \\
Claude.ai Wave 3 & 0.099 & 0.354 & 0.000 & 5.635 \\
Claude.ai Wave 5 & 0.100 & 0.312 & 0.000 & 5.499 \\
1P API Wave 3 & 0.103 & 0.474 & 0.000 & 10.734 \\
1P API Wave 5 & 0.101 & 0.551 & 0.000 & 14.218 \\
Copilot raw & 0.162 & 0.096 & 0.000 & 0.492 \\
Copilot reweighted & 0.163 & 0.040 & 0.000 & 0.479 \\
Composite with Wave 5 AEI weights & 0.067 & 0.127 & 0.000 & 1.070 \\
\end{longtable}
}

\emph{\textbf{Panel C. Key benchmark data sources}}

{\def\LTcaptype{none} 
\begin{longtable}[]{@{}
  >{\raggedright\arraybackslash}p{0.4500\linewidth}
  >{\centering\arraybackslash}p{0.1833\linewidth}
  >{\centering\arraybackslash}p{0.1833\linewidth}
  >{\centering\arraybackslash}p{0.1833\linewidth}@{}}
\toprule\noalign{}
\begin{minipage}[b]{\linewidth}\raggedright
\textbf{Source}
\end{minipage} & \begin{minipage}[b]{\linewidth}\centering
\textbf{Type}
\end{minipage} & \begin{minipage}[b]{\linewidth}\centering
\textbf{Period}
\end{minipage} & \begin{minipage}[b]{\linewidth}\centering
\textbf{Role in analysis}
\end{minipage} \\
\midrule\noalign{}
\endhead
\bottomrule\noalign{}
\endlastfoot
IPUMS-USA ACS 1-year & Household survey & 2015 to 2024 & Person-year outcome panel ($N = 13{,}104{,}072$) \\
Anthropic Economic Index (Claude.ai + 1P API) & Platform usage & Dec 2024 to Feb 2026 & Primary platform exposure measure ($p$) \\
BLS OEWS May 2024 & Establishment survey & 2024 & Workforce density $f(o)$ for $\psi_{o,p}$ \\
BLS Employment Projections 2024 to 2034 & Administrative & 2024 to 2034 & Dependent variable for cross-occupation regression \\
Real-Time Population Survey (\hyperlink{Bick2026}{Bick, Blandin, and Deming 2026}) & Population survey & Aug to Nov 2024 & Independent worker-reported AI use rate \\
Microsoft Copilot (\hyperlink{Tomlinson2025}{Tomlinson et al. 2025}) & Platform usage & Sept 2024 & Cross-platform robustness comparison \\
\end{longtable}
}

\needspace{6\baselineskip}
\textbf{Notes:} \emph{Panel A reports person-weighted means and standard deviations on the canonical analytical sample (American Community Survey wage-and-salary workers ages 16 to 64, 2015 to 2024) compared with the full American Community Survey universe of working-age adults. SMD denotes the standardized mean difference between samples. Panel B reports occupation-level descriptive statistics for the ten artificial-intelligence exposure variants used in the regression analysis (Baseline Composite; Composite Reweighted; Claude.ai consumer Waves 1, 3, and 5; first-party Application Programming Interface enterprise Waves 3 and 5; Microsoft Copilot raw and reweighted; Composite with Wave 5 AEI weights). Panel C lists the key benchmark data sources entering the analysis. Online Appendix Table B.1 reports the full variable definitions, the Anthropic Economic Index wave characteristics, the complete data-source inventory, the occupation-crosswalk coverage statistics, and the National Health Interview Survey cross-domain outcome summary statistics. Abbreviations: SOC = Standard Occupational Classification; AEI = Anthropic Economic Index; API = first-party Application Programming Interface enterprise; Composite = AI exposure score constructed following \hyperlink{Massenkoff2026}{Massenkoff and McCrory (2026)}; BLS = Bureau of Labor Statistics; OEWS = Occupational Employment and Wage Statistics; ACS = American Community Survey.}

\endgroup
\end{landscape}

\clearpage
\refstepcounter{none}\label{tab:2}
\begin{landscape}
\begingroup
\scriptsize
\setlength{\tabcolsep}{2pt}
\renewcommand{\arraystretch}{0.78}
\renewcommand{\baselinestretch}{0.85}\selectfont
\textbf{Table 2. Occupational Composition of the Platform-Visible
Workforce and the U.S. Workforce}

{\def\LTcaptype{none} 
\begin{longtable}[]{@{}
  >{\raggedright\arraybackslash}p{0.3143\linewidth}
  >{\centering\arraybackslash}p{0.1143\linewidth}
  >{\centering\arraybackslash}p{0.1143\linewidth}
  >{\centering\arraybackslash}p{0.1143\linewidth}
  >{\centering\arraybackslash}p{0.1143\linewidth}
  >{\centering\arraybackslash}p{0.1143\linewidth}
  >{\centering\arraybackslash}p{0.1143\linewidth}@{}}
\toprule\noalign{}
\begin{minipage}[b]{\linewidth}\raggedright
\textbf{Major group}
\end{minipage} & \begin{minipage}[b]{\linewidth}\centering
\textbf{BLS workforce \%}
\end{minipage} & \begin{minipage}[b]{\linewidth}\centering
\textbf{Claude.ai W5 \%}
\end{minipage} & \begin{minipage}[b]{\linewidth}\centering
\textbf{1P API W5 \%}
\end{minipage} & \begin{minipage}[b]{\linewidth}\centering
\textbf{ChatGPT pct\_work†}
\end{minipage} & \begin{minipage}[b]{\linewidth}\centering
\textbf{Microsoft Copilot score‡}
\end{minipage} & \begin{minipage}[b]{\linewidth}\centering
\textbf{BBD GenAI use \%§}
\end{minipage} \\
\midrule\noalign{}
\endhead
\bottomrule\noalign{}
\endlastfoot
Management & 7.1 & 4.3 & 3.4 & 50.0 & 0.135 & 45.2 \\
Business and Financial & 6.7 & 4.8 & 3.5 & 50.0 & 0.235 & 40.0 \\
Computer and Mathematical & 3.4 & 32.3 & 51.7 & 57.0 & 0.295 & 47.9 \\
Architecture and Engineering & 1.7 & 2.0 & 1.7 & 48.0 & 0.215 & 26.4 \\
Life, Physical, and Social Science & 0.9 & 6.3 & 4.4 & 48.0 & 0.189 &
26.4 \\
Community and Social Service & 1.7 & 3.2 & 0.5 & 44.0 & 0.251 & 23.0 \\
Legal & 0.8 & 0.6 & 0.2 & 44.0 & 0.135 & 23.0 \\
Educational Instruction and Library & 5.8 & 13.5 & 3.1 & 44.0 & 0.208 &
33.5 \\
Arts, Design, Entertainment, Sports, Media & 1.4 & 10.4 & 5.2 & 44.0 &
0.243 & 31.0 \\
Healthcare Practitioners and Technical & 6.2 & 3.0 & 1.0 & 44.0 & 0.120
& 30.2 \\
Healthcare Support & 4.8 & 0.3 & 0.2 & 40.0 & 0.047 & 11.9 \\
Protective Service & 2.4 & 0.4 & 0.5 & 40.0 & 0.123 & 11.9 \\
Food Preparation and Serving & 8.8 & 0.7 & 0.1 & 40.0 & 0.169 & 11.9 \\
Building and Grounds Cleaning & 2.9 & 0.2 & 0.1 & 40.0 & 0.076 & 11.9 \\
Personal Care and Service & 2.0 & 1.2 & 0.4 & 40.0 & 0.183 & 11.9 \\
Sales and Related & 8.7 & 4.9 & 4.1 & 40.0 & 0.292 & 20.9 \\
Office and Administrative Support & 11.8 & 9.4 & 17.6 & 40.0 & 0.262 &
18.5 \\
Farming, Fishing, and Forestry & 0.3 & 0.5 & 0.1 & 40.0 & 0.058 & n/a \\
Construction and Extraction & 4.1 & 0.2 & 0.3 & 40.0 & 0.074 & 20.4 \\
Installation, Maintenance, and Repair & 3.9 & 0.7 & 0.4 & 40.0 & 0.104 &
20.4 \\
Production & 5.7 & 0.7 & 1.2 & 40.0 & 0.107 & 20.4 \\
Transportation and Material Moving & 8.9 & 0.2 & 0.2 & 40.0 & 0.097 &
20.4 \\
\end{longtable}
}

\needspace{6\baselineskip}
\textbf{Notes:} \emph{How to read this table: Columns are grouped by
measurement type. BLS workforce \% is the population benchmark , the May
2024 Bureau of Labor Statistics Occupational Employment and Wage
Statistics employment share at the 2-digit Standard Occupational
Classification major-group level. Claude.ai W5 \% and 1P API W5 \% are
Wave 5 Anthropic Economic Index conversation shares (consumer Claude.ai
and first-party Application Programming Interface enterprise,
respectively); these three columns are directly comparable (each sums to
100 percent across the 22 major groups). The remaining three columns use
different unit conventions and should NOT be cross-compared row-by-row
to the share columns: † ChatGPT pct\_work is the \hyperlink{Chatterji2025}{\hyperlink{Chatterji2025}{Chatterji et al. (2025)}}
5-occupation-group distribution mapped step-wise within parent group (so
all rows in the same Chatterji parent group receive an identical value);
‡ Microsoft Copilot score is the \hyperlink{Tomlinson2025}{\hyperlink{Tomlinson2025}{Tomlinson et al. (2025)}} artificial-intelligence
applicability score on a 0-to-1 index (employment-weighted to Standard
Occupational Classification major group), not a share; § BBD GenAI use
\% is the \hyperlink{Bick2026}{Bick, Blandin, and Deming (2026)} Real-Time Population Survey
December 2024 share of workers using generative artificial
intelligence at work in the past month, a population survey use rate,
not a conversation share. marks BBD values from aggregated occupation
groups (17\_19 Architecture/Engineering/Science = 26.4\%; 21\_23
Legal/Social Services = 23.0\%; 31-39 Personal Services (Healthcare
Support, Protective, Food Prep, Building/Grounds, Personal Care) =
11.9\%; 47-53 Blue Collar (Construction, Installation, Production,
Transportation) = 20.4\%), \hyperlink{Bick2026}{Bick, Blandin, and Deming (2026)} collapsed these
Standard Occupational Classification major groups in their published
table 2 due to small sample sizes; the same aggregated rate applies to
every sub-major within the group. Major group 45 (Farming, Fishing,
Forestry) is too small for any reliable BBD estimate even after
aggregation; n/a denotes not surveyed. The systematic
over-representation of Computer and Mathematical (15) occupations in
platform-visible data (32.3\% Claude.ai W5; 51.7\% 1P API W5) relative
to the underlying U.S. workforce (3.4\% BLS) and the corresponding
under-representation of Office and Administrative Support (43), Sales
and Related (41), Food Preparation and Serving (35), and Transportation
and Material Moving (53) are the central composition findings discussed
in Section IV.}

\endgroup
\end{landscape}

\clearpage
\refstepcounter{none}\label{tab:3}
\begin{landscape}
\begingroup
\scriptsize
\setlength{\tabcolsep}{2pt}
\renewcommand{\arraystretch}{0.78}
\renewcommand{\baselinestretch}{0.85}\selectfont
\textbf{Table 3. Difference-in-Differences Estimates of AI Exposure on
Three Labor-Market Outcomes}

{\def\LTcaptype{none} 
\begin{longtable}[]{@{}
  >{\raggedright\arraybackslash}p{0.2200\linewidth}
  >{\centering\arraybackslash}p{0.0750\linewidth}
  >{\centering\arraybackslash}p{0.0750\linewidth}
  >{\centering\arraybackslash}p{0.0750\linewidth}
  >{\centering\arraybackslash}p{0.0300\linewidth}
  >{\raggedright\arraybackslash}p{0.2200\linewidth}
  >{\centering\arraybackslash}p{0.0750\linewidth}
  >{\centering\arraybackslash}p{0.0750\linewidth}
  >{\centering\arraybackslash}p{0.0750\linewidth}@{}}
\toprule\noalign{}
\multicolumn{4}{@{}>{\centering\arraybackslash}p{0.4444\linewidth + 6\tabcolsep}}{%
\begin{minipage}[b]{\linewidth}\centering
\emph{\textbf{Panel A. Cross-source DiD on three labor-market outcomes}}
\end{minipage}} & \begin{minipage}[b]{\linewidth}\centering
\end{minipage} &
\multicolumn{4}{>{\centering\arraybackslash}p{0.4444\linewidth + 6\tabcolsep}@{}}{%
\begin{minipage}[b]{\linewidth}\centering
\emph{\textbf{Panel B. Within-vendor channel substitution}}
\end{minipage}} \\
\midrule\noalign{}
\endhead
\bottomrule\noalign{}
\endlastfoot
\textbf{Variant} & \textbf{Emp.} & \textbf{LFP} & \textbf{Log wage} & &
\textbf{Variant} & \textbf{Emp.} & \textbf{LFP} & \textbf{Log wage} \\
Baseline Composite (Anthropic) & -0.139*** & +0.056* & -0.355*** & &
Consumer Claude.ai (W3) & -0.087* & +0.082* & -0.716*** \\
& (0.035) & (0.032) & (0.107) & & & (0.052) & (0.041) & (0.155) \\
\addlinespace[2pt]
Composite Reweighted & -0.010 & +0.028 & +0.204 & & Consumer Claude.ai
(W4) & -0.098** & +0.094** & -0.358*** \\
& (0.045) & (0.039) & (0.131) & & & (0.047) & (0.039) & (0.126) \\
\addlinespace[2pt]
Claude.ai consumer (W1) & -0.116*** & +0.005 & -0.445*** & & Consumer
Claude.ai (W5) & -0.076* & +0.094** & -0.550*** \\
& (0.035) & (0.030) & (0.099) & & & (0.043) & (0.041) & (0.129) \\
\addlinespace[2pt]
Claude.ai consumer (W3) & -0.163*** & -0.045 & -0.383*** & & 1P API
enterprise (W3) & +0.015 & +0.080 & -0.089 \\
& (0.033) & (0.030) & (0.106) & & & (0.058) & (0.048) & (0.143) \\
\addlinespace[2pt]
Claude.ai consumer (W5) & -0.222*** & -0.126*** & -0.322*** & & 1P API
enterprise (W4) & +0.041 & +0.080** & +0.051 \\
& (0.042) & (0.037) & (0.109) & & & (0.047) & (0.035) & (0.113) \\
\addlinespace[2pt]
1P API enterprise (W3) & -0.144*** & -0.029 & -0.141 & & 1P API
enterprise (W5) & +0.055 & +0.115*** & -0.204 \\
& (0.035) & (0.031) & (0.089) & & & (0.044) & (0.037) & (0.128) \\
\addlinespace[2pt]
1P API enterprise (W5) & -0.152*** & -0.032 & -0.165* & & Pooled W3
(Claude.ai + 1P API) & -0.064 & +0.074* & -0.392*** \\
& (0.033) & (0.030) & (0.085) & & & (0.046) & (0.039) & (0.142) \\
\addlinespace[2pt]
Microsoft Copilot (raw) & -0.191*** & -0.028 & +0.009 & & Pooled W4
(Claude.ai + 1P API) & -0.098** & +0.059* & -0.422*** \\
& (0.042) & (0.036) & (0.100) & & & (0.038) & (0.032) & (0.135) \\
\addlinespace[2pt]
Microsoft Copilot (reweighted) & -0.110*** & +0.026 & -0.681*** & &
Pooled W5 (Claude.ai + 1P API) & -0.039 & +0.089*** & -0.492*** \\
& (0.039) & (0.034) & (0.073) & & & (0.038) & (0.033) & (0.135) \\
\addlinespace[2pt]
Composite with Wave 5 AEI weights & -0.155*** & +0.049 & -0.752*** & & &
& & \\
& (0.031) & (0.031) & (0.119) & & & & & \\
\addlinespace[2pt]
\textbf{N (range across variants)} & \textbf{8,630,961 to 12,279,674} &
\textbf{8,630,961 to 12,279,674} & \textbf{7,862,841 to 11,181,713} & &
\textbf{N (range across variants)} & \textbf{4,498,101 to 6,172,181} &
\textbf{4,498,101 to 6,172,181} & \textbf{4,139,500 to 5,662,442} \\
\end{longtable}
}

\needspace{6\baselineskip}
\textbf{Notes:} \emph{Each cell reports a beta-times-100 coefficient
with the cluster-robust standard error in parentheses below.
Individual-level weighted least squares on the canonical American
Community Survey analytical sample (wage-and-salary workers, ages 16 to
64, 2015 to 2024); two-way fixed effects on year, state, and 6-digit
Standard Occupational Classification 2018; controls for sex, marital
status, three educational-attainment indicators, age, age squared,
number of own children. Exposure z-scored within sample using
employment-weighted means; person-weighted by perwt; standard errors
clustered on state (51 states including the District of Columbia).
Post-treatment indicator = 1 if year is 2023 or later. Panel A:
cross-source dispersion across ten platform-input variants on three
labor-market outcomes (binary employment, labor-force participation, log
weekly wage conditional on positive earnings). Panel B: within-vendor
channel substitution under the Baseline Composite (Anthropic) recipe at
Anthropic Economic Index Waves 3, 4, and 5 across three channels
(Claude.ai consumer, first-party Application Programming Interface
enterprise, pooled). Each panel\textquotesingle s last row reports the
range of person-year observations across variants per outcome.
Robustness to dropping 2020 and 2021 (COVID-19 disruption years) is
reported in Online Appendix Table B.6. *** p\textless0.01, **
p\textless0.05, * p\textless0.10.}

\endgroup
\end{landscape}

\clearpage
\refstepcounter{none}\label{tab:4}
\begin{landscape}
\begingroup
\scriptsize
\setlength{\tabcolsep}{2pt}
\renewcommand{\arraystretch}{0.78}
\renewcommand{\baselinestretch}{0.85}\selectfont
\textbf{Table 4. Wave-by-Wave Reweighted Composite
Difference-in-Differences as a $\psi$ versus $\theta$ Channel Diagnostic}

{\def\LTcaptype{none} 
\begin{longtable}[]{@{}
  >{\raggedright\arraybackslash}p{1.00in}
  >{\centering\arraybackslash}p{0.85in}
  >{\centering\arraybackslash}p{0.85in}
  >{\centering\arraybackslash}p{0.85in}
  >{\centering\arraybackslash}p{0.85in}
  >{\centering\arraybackslash}p{0.85in}
  >{\centering\arraybackslash}p{0.85in}@{}}
\toprule\noalign{}
\multirow{2}{=}{\begin{minipage}[b]{\linewidth}\raggedright
\textbf{Wave (release)}
\end{minipage}} &
\multicolumn{2}{>{\centering\arraybackslash}p{1.70in + 2\tabcolsep}}{%
\begin{minipage}[b]{\linewidth}\centering
\textbf{Employment}
\end{minipage}} &
\multicolumn{2}{>{\centering\arraybackslash}p{1.70in + 2\tabcolsep}}{%
\begin{minipage}[b]{\linewidth}\centering
\textbf{LFP}
\end{minipage}} &
\multicolumn{2}{>{\centering\arraybackslash}p{1.70in + 2\tabcolsep}@{}}{%
\begin{minipage}[b]{\linewidth}\centering
\textbf{Log wage}
\end{minipage}} \\
& \begin{minipage}[b]{\linewidth}\centering
\textbf{Baseline}
\end{minipage} & \begin{minipage}[b]{\linewidth}\centering
\textbf{Reweighted}
\end{minipage} & \begin{minipage}[b]{\linewidth}\centering
\textbf{Baseline}
\end{minipage} & \begin{minipage}[b]{\linewidth}\centering
\textbf{Reweighted}
\end{minipage} & \begin{minipage}[b]{\linewidth}\centering
\textbf{Baseline}
\end{minipage} & \begin{minipage}[b]{\linewidth}\centering
\textbf{Reweighted}
\end{minipage} \\
\midrule\noalign{}
\endhead
\bottomrule\noalign{}
\endlastfoot
W1 (Dec 2024) & -0.113*** & -0.014 & +0.049* & +0.001 & -0.478*** &
+0.135*** \\
& (0.027) & (0.012) & (0.028) & (0.011) & (0.091) & (0.040) \\
W2 (Feb 2025) & -0.120*** & -0.017 & +0.047* & +0.002 & -0.605*** &
+0.104*** \\
& (0.028) & (0.013) & (0.029) & (0.012) & (0.098) & (0.038) \\
W3 (Aug 2025) & -0.110*** & -0.009 & +0.047* & +0.003 & -0.492*** &
+0.076** \\
& (0.026) & (0.009) & (0.027) & (0.008) & (0.090) & (0.030) \\
W4 (Nov 2025) & -0.091*** & -0.003 & +0.047** & -0.000 & -0.462*** &
+0.100*** \\
& (0.026) & (0.013) & (0.024) & (0.011) & (0.085) & (0.036) \\
W5 (Feb 2026) & -0.083*** & -0.021* & +0.054** & -0.002 & -0.478*** &
+0.038 \\
& (0.024) & (0.012) & (0.023) & (0.011) & (0.082) & (0.035) \\
\textbf{Cross-wave span} & \textbf{0.037} & \textbf{0.018} &
\textbf{0.007} & \textbf{0.006} & \textbf{0.143} & \textbf{0.097} \\
\textbf{Span ratio} &
\multicolumn{2}{>{\centering\arraybackslash}p{1.70in + 2\tabcolsep}}{%
\textbf{0.49}} &
\multicolumn{2}{>{\centering\arraybackslash}p{1.70in + 2\tabcolsep}}{%
\textbf{0.85}} &
\multicolumn{2}{>{\centering\arraybackslash}p{1.70in + 2\tabcolsep}@{}}{%
\textbf{0.68}} \\
\end{longtable}
}

\needspace{6\baselineskip}
\textbf{Notes:} \emph{For each Anthropic Economic Index wave we substitute that
wave\textquotesingle s conversation share into the \hyperlink{Massenkoff2026}{Massenkoff and McCrory (2026)} composite formula (rubric $\times$ task-time $\times$ within-occupation
task share) and re-estimate the DiD on each
labor-market outcome. The baseline block uses the resulting composite
directly; the workforce-reweighted block multiplies by $f_\mathrm{BLS} / f_p^{(w)}$,
where $f_p^{(w)}$ is the wave-$w$ cross-occupation conversation density. Coefficients report $\beta \times 100$ per standard deviation
with cluster-robust standard errors clustered at the state level in
parentheses on the row beneath, from individual-level weighted least
squares on the 2015 to 2024 American Community Survey panel (13.1
million person-year observations) with six-digit Standard Occupational
Classification, state, and year fixed effects and the demographic
controls of Section II.D. Cross-wave span is the maximum minus minimum
coefficient across the five waves for each outcome $\times$ specification
combination; the span ratio is the reweighted span divided by the
baseline span. A span ratio near zero indicates that the
cross-occupation user-base channel ($\psi$) accounts for most of the
cross-wave coefficient variation; a span ratio near one would indicate
that within-occupation behavior change ($\theta$) is the central source.
Supporting diagnostic evidence on $\psi$ and $\theta$ is reported in Table B.11.
Significance levels: * p < 0.10, ** p < 0.05, *** p < 0.01.}

\endgroup
\end{landscape}

\clearpage
\refstepcounter{none}\label{tab:5}
\begin{landscape}
\begingroup
\scriptsize
\setlength{\tabcolsep}{2pt}
\renewcommand{\arraystretch}{0.78}
\renewcommand{\baselinestretch}{0.85}\selectfont
\textbf{Table 5. Replication of \hyperlink{Massenkoff2026}{Massenkoff and McCrory (2026)} Composite
Construction and Workforce Reweighting}

{\def\LTcaptype{none} 
\begin{longtable}[]{@{}
  >{\raggedright\arraybackslash}p{0.4400\linewidth}
  >{\centering\arraybackslash}p{0.1400\linewidth}
  >{\centering\arraybackslash}p{0.1400\linewidth}
  >{\centering\arraybackslash}p{0.1400\linewidth}
  >{\centering\arraybackslash}p{0.1400\linewidth}@{}}
\toprule\noalign{}
\begin{minipage}[b]{\linewidth}\raggedright
\textbf{Exposure measure}
\end{minipage} & \begin{minipage}[b]{\linewidth}\centering
\textbf{N}
\end{minipage} & \begin{minipage}[b]{\linewidth}\centering
\textbf{Original}
\end{minipage} & \begin{minipage}[b]{\linewidth}\centering
\textbf{BLS-reweighted}
\end{minipage} & \begin{minipage}[b]{\linewidth}\centering
\textbf{Attenuation in \textbar Coef\textbar{}}
\end{minipage} \\
\midrule\noalign{}
\endhead
\bottomrule\noalign{}
\endlastfoot
Panel A. Massenkoff-McCrory baseline composite reweighting test & & &
& \\
M-M baseline composite (Massenkoff-McCrory 2026) & 756 / 766 & -70.09***
& -38.28** & 45.4\% \\
& & (17.90) & (17.35) & \\
Panel B. Alternative-rubric and alternative-platform exposures
(unreweighted; reweighted versions in Online Appendix Table B.9) & & &
& \\
Anthropic Claude.ai consumer Wave 1 (Dec 2024) & 582 & +43.30 & & \\
& & (33.64) & & \\
Anthropic Claude.ai consumer Wave 2 (Feb 2025) & 579 & -8.16 & & \\
& & (34.23) & & \\
Anthropic Claude.ai consumer Wave 3 (Aug 2025) & 538 & -12.10 & & \\
& & (35.08) & & \\
Anthropic Claude.ai consumer Wave 4 (Nov 2025) & 559 & -28.12 & & \\
& & (33.46) & & \\
Anthropic Claude.ai consumer Wave 5 (Feb 2026) & 561 & -68.01** & & \\
& & (28.80) & & \\
Anthropic 1P API enterprise Wave 3 (Aug 2025) & 480 & +27.61 & & \\
& & (37.16) & & \\
Anthropic 1P API enterprise Wave 4 (Nov 2025) & 485 & +23.88 & & \\
& & (36.38) & & \\
Anthropic 1P API enterprise Wave 5 (Feb 2026) & 480 & +34.00 & & \\
& & (36.03) & & \\
Microsoft Bing Copilot 2024 (\hyperlink{Tomlinson2025}{\hyperlink{Tomlinson2025}{Tomlinson et al. 2025}}) & 785 & -150.04*** &
& \\
& & (24.00) & & \\
OpenAI ChatGPT 2024 (\hyperlink{Chatterji2025}{\hyperlink{Chatterji2025}{Chatterji et al. 2025}}, SOC-major-expanded) & 1112 &
+174.00*** & & \\
& & (13.61) & & \\
\hyperlink{Bick2026}{Bick, Blandin, and Deming (2026)}, SWAA SOC-major-expanded & 1091 &
+102.26*** & & \\
& & (13.28) & & \\
\hyperlink{Bick2026}{Bick, Blandin, and Deming (2026)}, RPS six-digit SOC micro & 270 &
+166.81*** & & \\
& & (45.80) & & \\
& & & & \\
Manski partial-identification bounds on \textbar Coef per SD\textbar{}
(x100) & & & & \\
Lower bound: M-M baseline composite, BLS-reweighted & & 38.28 & & \\
Upper bound: OpenAI ChatGPT 2024 (\hyperlink{Chatterji2025}{Chatterji et al. 2025}, SOC-major-expanded) &
& 174.00 & & \\
Identified-set width & & 135.72 & & \\
\end{longtable}
}

\needspace{6\baselineskip}
\textbf{Notes:} \emph{Panel A reports the \hyperlink{Massenkoff2026}{Massenkoff and McCrory (2026)}
baseline composite and its workforce-reweighted version. Panel B reports
alternative-platform and alternative-rubric exposures, unreweighted, in
the same between-occupation specification. Two \hyperlink{Bick2026}{Bick, Blandin, and Deming (2026)} rows appear in Panel B for methodological comparison. The
BBD-SWAA row uses the SOC-major-aggregated public release expanded to
NEM (n = 1,091). The BBD-RPS six-digit micro row uses the early-2026
micro release at the native six-digit SOC level (n = 270). The RPS is a
national online survey fielded on Qualtrics with awareness-adjusted
demographic weights, not a probability sample of the U.S. workforce.
Coefficients are scaled per standard deviation of the exposure measure
times 100. Employment-weighted standard errors in parentheses. ***
p\textless0.01, ** p\textless0.05, * p\textless0.10.}

\endgroup
\end{landscape}

\clearpage
\section*{Figures}
\addcontentsline{toc}{section}{Figures}
\begin{figure}[!p]
\centering
{\normalsize\bfseries Figure 1. Pairwise Spearman rank correlations across eleven occupational AI-exposure measures and the BLS workforce.}\par\medskip
\includegraphics[width=0.97\linewidth,height=0.82\textheight,keepaspectratio]{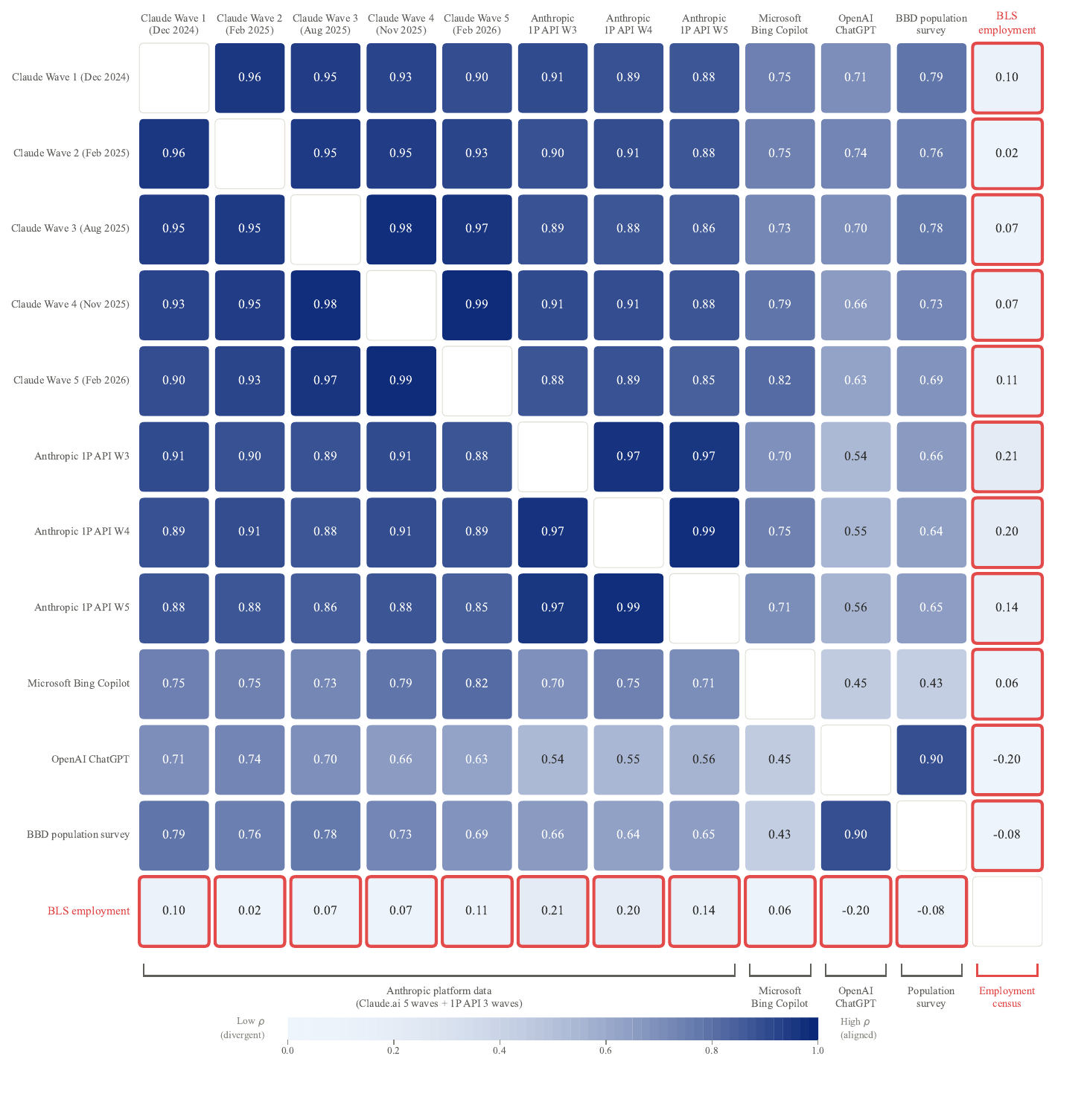}
\refstepcounter{none}\label{fig:1}
\par\medskip
\begin{minipage}{0.92\linewidth}
\color[HTML]{333333}\footnotesize
{\bfseries Notes:} \textit{Each cell reports the Spearman rank correlation between two AI-exposure measures across 22 SOC major occupational groups. The five Anthropic Claude.ai waves plus three Anthropic API waves cluster at correlations above 0.85; correlations with Microsoft Copilot, OpenAI ChatGPT, and the BBD population survey fall in the 0.43 to 0.79 range. The red-bordered BLS employment row and column highlight that no platform-derived measure correlates with the workforce above $\rho = 0.33$.}
\end{minipage}
\end{figure}
\clearpage
\begin{landscape}
\begin{figure}[!p]
\centering
{\normalsize\bfseries Figure 2. Theoretical AI capability versus each platform's published main exposure measure.}\par\medskip
\includegraphics[width=0.97\linewidth,height=0.78\textheight,keepaspectratio]{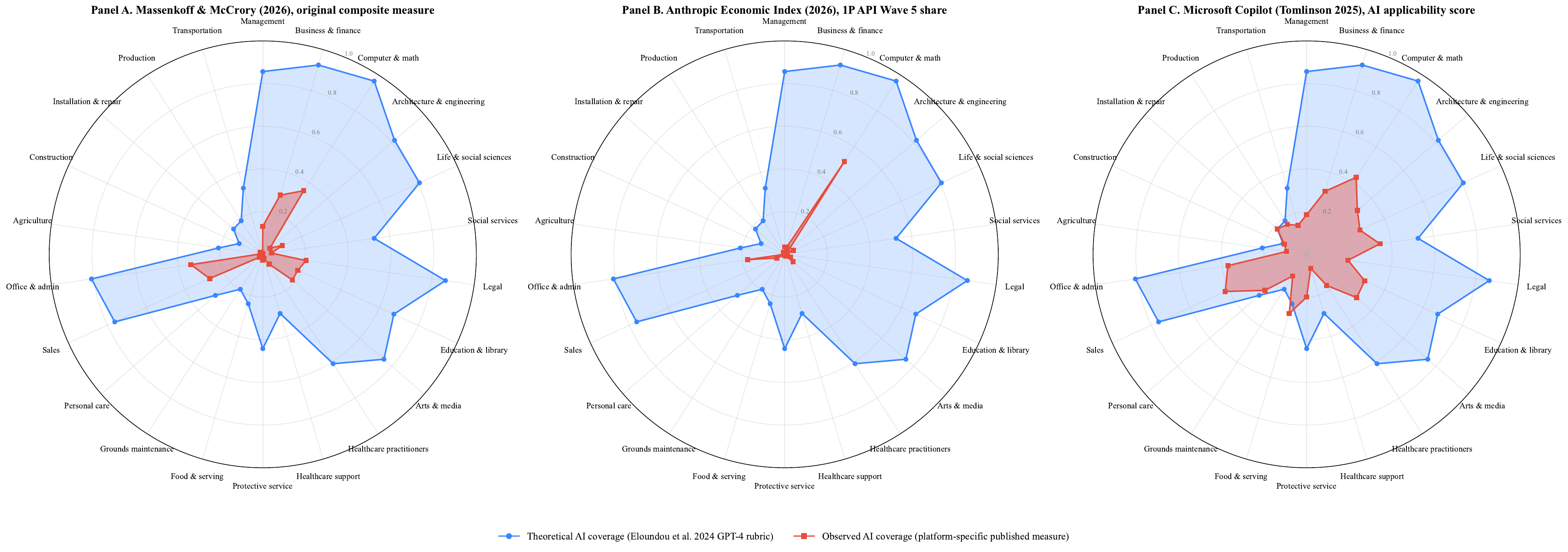}
\refstepcounter{none}\label{fig:2}
\par\medskip
\begin{minipage}{0.92\linewidth}
\color[HTML]{333333}\footnotesize
{\bfseries Notes:} \textit{Three-panel polar visualization. Each panel plots one platform's published main exposure measure (radial position) against the theoretical capability benchmark from \hyperlink{Eloundou2024}{\hyperlink{Eloundou2024}{Eloundou et al. (2024)}} (angular position), with each point one of the 22 SOC major groups. Panel A reports the Baseline Composite (Anthropic); Panel B the Anthropic 1P API Wave 5 share; Panel C the Tomlinson Microsoft Copilot AI applicability score.}
\end{minipage}
\end{figure}
\end{landscape}
\clearpage
\begin{figure}[!p]
\centering
{\normalsize\bfseries Figure 3. Event-study coefficients on the employment indicator by exposure variant, 2015 to 2024.}\par\medskip
\includegraphics[width=0.97\linewidth,height=0.82\textheight,keepaspectratio]{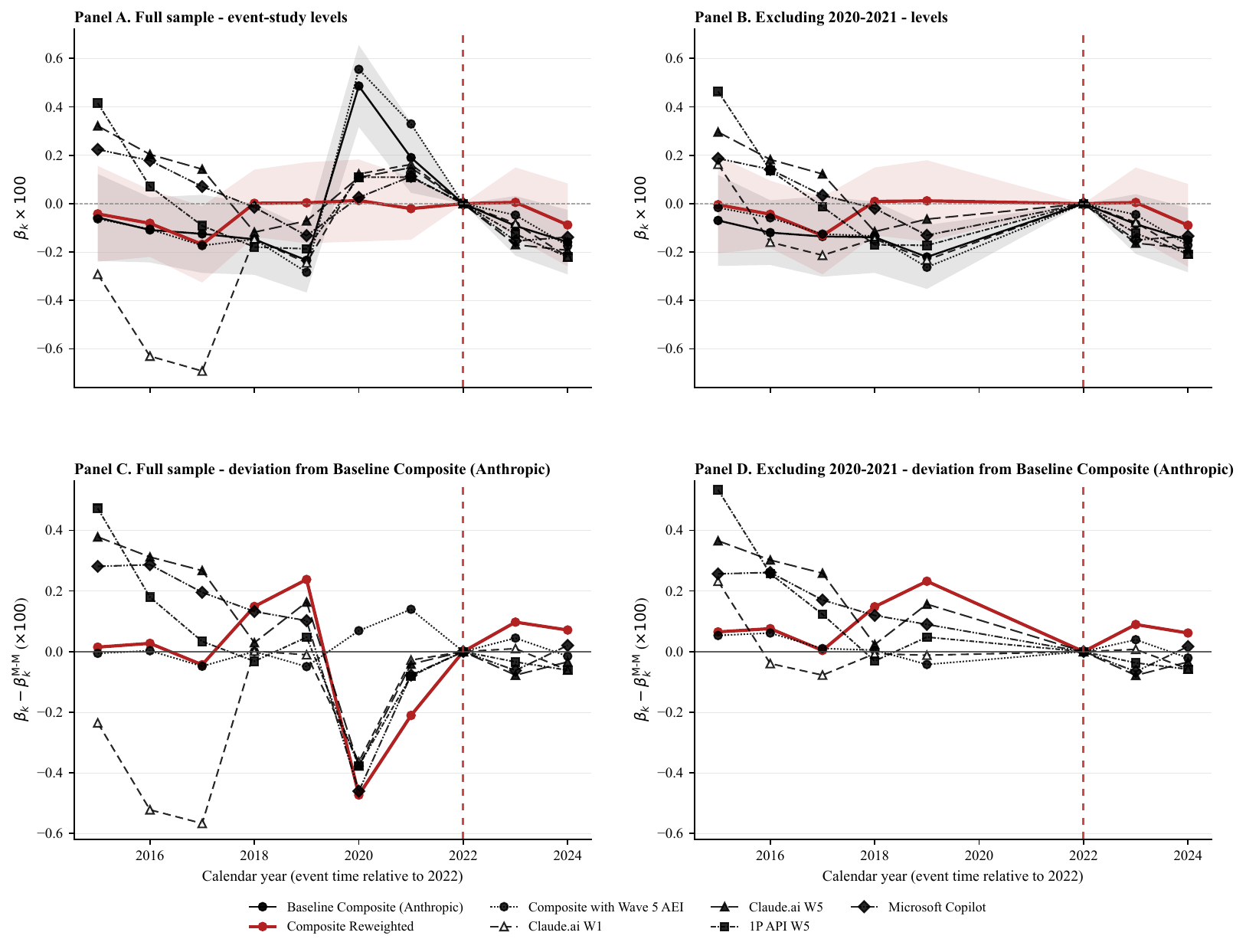}
\refstepcounter{none}\label{fig:3}
\par\medskip
\begin{minipage}{0.92\linewidth}
\color[HTML]{333333}\footnotesize
{\bfseries Notes:} \textit{Year-by-year interaction coefficients on the employment indicator from the event-study extension of the main DiD specification (equation 6) for seven exposure variants, 2015 to 2024 with 2022 as the reference year. Pre-period coefficients are statistically indistinguishable from zero for all seven variants; post-treatment coefficients diverge immediately in 2023.}
\end{minipage}
\end{figure}
\clearpage
\begin{landscape}
\begin{figure}[!p]
\centering
{\normalsize\bfseries Figure 4. Slope attenuation under workforce reweighting in the cross-occupation BLS Projections regression.}\par\medskip
\includegraphics[width=0.97\linewidth,height=0.78\textheight,keepaspectratio]{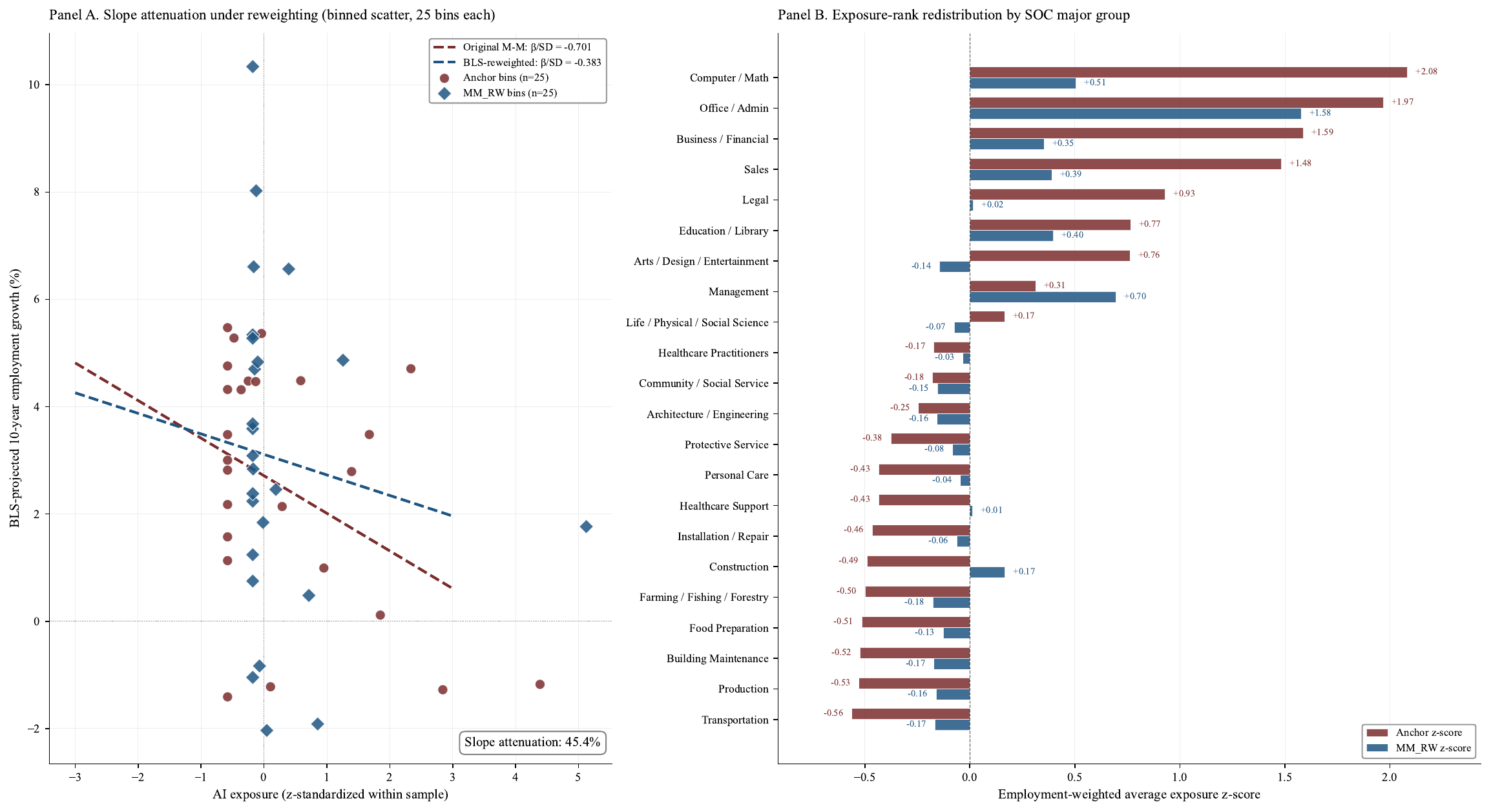}
\refstepcounter{none}\label{fig:4}
\par\medskip
\begin{minipage}{0.92\linewidth}
\color[HTML]{333333}\footnotesize
{\bfseries Notes:} \textit{Panel A is a binned scatter (25 equal-size bins) of BLS-projected 2024 to 2034 employment growth against the standardized composite exposure, with the published unreweighted slope of $-0.70$ per standard deviation and the workforce-reweighted slope of $-0.38$ overlaid. Panel B reports the cross-occupation redistribution at the SOC major-group level, where the largest negative shifts under reweighting fall on Computer and Mathematical, Office and Administrative Support, and Sales.}
\end{minipage}
\end{figure}
\end{landscape}
\clearpage

\clearpage
\section*{References}
\addcontentsline{toc}{section}{References}
\begingroup
\setlength{\parindent}{0pt}
\setlength{\parskip}{0.5em}
\noindent\hangindent=1.5em\hangafter=1 \hypertarget{Acemoglu2011}{Acemoglu, Daron, and David Autor.} 2011. "Skills, Tasks and Technologies: Implications for Employment and Earnings." In \textit{Handbook of Labor Economics}, Vol. 4B, edited by Orley Ashenfelter and David Card, 1043--1171. Amsterdam: Elsevier.\par
\noindent\hangindent=1.5em\hangafter=1 \hypertarget{Acemoglu2020}{Acemoglu, Daron, and Pascual Restrepo.} 2020. "Robots and Jobs: Evidence from US Labor Markets." \textit{Journal of Political Economy} 128 (6): 2188--2244.\par
\noindent\hangindent=1.5em\hangafter=1 \hypertarget{Appel2025}{Appel, Ruth, Peter McCrory, Alex Tamkin, Miles McCain, Tyler Neylon, and Michael Stern.} 2025. "Anthropic Economic Index Report: Uneven Geographic and Enterprise AI Adoption." arXiv:2511.15080.\par
\noindent\hangindent=1.5em\hangafter=1 \hypertarget{Autor2003}{Autor, David H., Frank Levy, and Richard J. Murnane.} 2003. "The Skill Content of Recent Technological Change: An Empirical Exploration." \textit{Quarterly Journal of Economics} 118 (4): 1279--1333.\par
\noindent\hangindent=1.5em\hangafter=1 \hypertarget{Bick2026}{Bick, Alexander, Adam Blandin, and David J. Deming.} 2026. "The Rapid Adoption of Generative AI." NBER Working Paper 32966.\par
\noindent\hangindent=1.5em\hangafter=1 \hypertarget{BBDS2026}{Bick, Alexander, Adam Blandin, David J. Deming, and Tyler Schumacher.} 2026. "What Work Does Generative AI Do?" Federal Reserve Bank of St. Louis Working Paper, April 27, 2026.\par
\noindent\hangindent=1.5em\hangafter=1 \hypertarget{Bound2001}{Bound, John, Charles Brown, and Nancy Mathiowetz.} 2001. "Measurement Error in Survey Data." In \textit{Handbook of Econometrics}, Vol. 5, edited by James J. Heckman and Edward Leamer, 3705--3843. Amsterdam: Elsevier.\par
\noindent\hangindent=1.5em\hangafter=1 \hypertarget{Brynjolfsson2018}{Brynjolfsson, Erik, Tom Mitchell, and Daniel Rock.} 2018. "What Can Machines Learn, and What Does It Mean for Occupations and the Economy?" \textit{AEA Papers and Proceedings} 108: 43--47.\par
\noindent\hangindent=1.5em\hangafter=1 \hypertarget{BrynjolfssonLi2023}{Brynjolfsson, Erik, Danielle Li, and Lindsey R. Raymond.} 2023. "Generative AI at Work." NBER Working Paper 31161.\par
\noindent\hangindent=1.5em\hangafter=1 \hypertarget{Brynjolfsson2025}{Brynjolfsson, Erik, Bharat Chandar, and Ruyu Chen.} 2025. "Canaries in the Coal Mine? Six Facts about the Recent Employment Effects of Artificial Intelligence." Stanford Digital Economy Lab Working Paper.\par
\noindent\hangindent=1.5em\hangafter=1 \hypertarget{Chatterji2025}{Chatterji, Aaron, Thomas Cunningham, David J. Deming, Zoe Hitzig, Christopher Ong, Carl Yan Shan, and Kevin Wadman.} 2025. "How People Use ChatGPT." NBER Working Paper 34255.\par
\noindent\hangindent=1.5em\hangafter=1 \hypertarget{Chen2025}{Chen, Danqing, Carina Kane, Austin Kozlowski, Nadav Kunievsky, and James A. Evans.} 2025. "The (Short-Term) Effects of Large Language Models on Unemployment and Earnings." arXiv:2509.15510.\par
\noindent\hangindent=1.5em\hangafter=1 \hypertarget{Cui2024}{Cui, Zheyuan (Kevin), Mert Demirer, Sonia Jaffe, Leon Musolff, Sida Peng, and Tobias Salz.} 2024. "The Effects of Generative AI on High-Skilled Work: Evidence from Three Field Experiments with Software Developers." SSRN Working Paper 4945566.\par
\noindent\hangindent=1.5em\hangafter=1 \hypertarget{DellAcqua2023}{Dell'Acqua, Fabrizio, Edward McFowland III, Ethan R. Mollick, Hila Lifshitz-Assaf, Katherine Kellogg, Saran Rajendran, Lisa Krayer, François Candelon, and Karim R. Lakhani.} 2023. "Navigating the Jagged Technological Frontier: Field Experimental Evidence of the Effects of Artificial Intelligence on Knowledge Worker Productivity and Quality." Harvard Business School Working Paper 24-013.\par
\noindent\hangindent=1.5em\hangafter=1 \hypertarget{Eloundou2024}{Eloundou, Tyna, Sam Manning, Pamela Mishkin, and Daniel Rock.} 2024. "GPTs are GPTs: Labor Market Impact Potential of LLMs." \textit{Science} 384 (6702): 1306--1308.\par
\noindent\hangindent=1.5em\hangafter=1 \hypertarget{Felten2021}{Felten, Edward W., Manav Raj, and Robert Seamans.} 2021. "Occupational, Industry, and Geographic Exposure to Artificial Intelligence: A Novel Dataset and Its Potential Uses." \textit{Strategic Management Journal} 42 (12): 2195--2217.\par
\noindent\hangindent=1.5em\hangafter=1 \hypertarget{Frey2017}{Frey, Carl Benedikt, and Michael A. Osborne.} 2017. "The Future of Employment: How Susceptible are Jobs to Computerisation?" \textit{Technological Forecasting and Social Change} 114: 254--280.\par
\noindent\hangindent=1.5em\hangafter=1 \hypertarget{Goos2014}{Goos, Maarten, Alan Manning, and Anna Salomons.} 2014. "Explaining Job Polarization: Routine-Biased Technological Change and Offshoring." \textit{American Economic Review} 104 (8): 2509--2526.\par
\noindent\hangindent=1.5em\hangafter=1 \hypertarget{Handa2025}{Handa, Kunal, Alex Tamkin, Miles McCain, Saffron Huang, Esin Durmus, Sarah Heck, Jared Mueller, et al.} 2025. "Which Economic Tasks are Performed with AI? Evidence from Millions of Claude Conversations." Anthropic Technical Report.\par
\noindent\hangindent=1.5em\hangafter=1 \hypertarget{Lane2023}{Lane, Marguerita, Morgan Williams, and Stijn Broecke.} 2023. "The Impact of AI on the Workplace: Main Findings from the OECD AI Surveys of Employers and Workers." OECD Social, Employment and Migration Working Papers No. 288. Paris: OECD Publishing.\par
\noindent\hangindent=1.5em\hangafter=1 \hypertarget{Manski2003}{Manski, Charles F.} 2003. \textit{Partial Identification of Probability Distributions}. New York: Springer.\par
\noindent\hangindent=1.5em\hangafter=1 \hypertarget{Massenkoff2026}{Massenkoff, Maxim, and Peter McCrory.} 2026. "Labor Market Impacts of AI: A New Measure and Early Evidence." Anthropic Research, March 5.\par
\noindent\hangindent=1.5em\hangafter=1 \hypertarget{Meng2018}{Meng, Xiao-Li.} 2018. "Statistical Paradises and Paradoxes in Big Data (I): Law of Large Populations, Big Data Paradox, and the 2016 US Presidential Election." \textit{Annals of Applied Statistics} 12 (2): 685--726.\par
\noindent\hangindent=1.5em\hangafter=1 \hypertarget{Noy2023}{Noy, Shakked, and Whitney Zhang.} 2023. "Experimental Evidence on the Productivity Effects of Generative Artificial Intelligence." \textit{Science} 381 (6654): 187--192.\par
\noindent\hangindent=1.5em\hangafter=1 \hypertarget{Tamer2010}{Tamer, Elie.} 2010. "Partial Identification in Econometrics." \textit{Annual Review of Economics} 2: 167--195.\par
\noindent\hangindent=1.5em\hangafter=1 \hypertarget{Tamkin2024}{Tamkin, Alex, Miles McCain, Kunal Handa, Esin Durmus, Liane Lovitt, Ankur Rathi, Saffron Huang, et al.} 2024. "Clio: Privacy-Preserving Insights into Real-World AI Use." Anthropic Technical Report.\par
\noindent\hangindent=1.5em\hangafter=1 \hypertarget{Tomlinson2025}{Tomlinson, Kiran, Sonia Jaffe, Will Wang, Scott Counts, and Siddharth Suri.} 2025. "Working with AI: Measuring the Applicability of Generative AI to Occupations." Microsoft Research Technical Report (arXiv:2507.07935).\par
\noindent\hangindent=1.5em\hangafter=1 \hypertarget{USCensus2025}{U.S. Census Bureau.} 2025. "Household Trends and Outlook Pulse Survey (HTOPS): Public-Use Data Files." Washington, DC: U.S. Department of Commerce.\par
\noindent\hangindent=1.5em\hangafter=1 \hypertarget{Webb2020}{Webb, Michael.} 2020. "The Impact of Artificial Intelligence on the Labor Market." SSRN Working Paper 3482150.\par
\noindent\hangindent=1.5em\hangafter=1 \hypertarget{Yin2026}{Yin, Michelle, Hoa Vu, and Claudia Persico.} 2026. "How (un)Stable Are LLM Occupational Exposure Scores? Evidence from Multi-Model Replication." NBER Working Paper 35110.\par
\endgroup

\clearpage
\addcontentsline{toc}{section}{Appendix A}
\section{Appendix A. Derivation of the Probability Limit and Proofs}\label{appendix-a.-derivation-of-the-probability-limit-and-proofs}

This appendix collects the derivations supporting Section I. Appendix A.1 states the setup. Appendix A.2 derives the additive decomposition of the platform-derived proxy. Appendix A.3 derives the probability limit for the OLS coefficient. Appendix A.4 verifies the special cases discussed in the main text. Appendix A.5 derives the reweighted proxy and states the maintained ordering condition used for partial identification. Appendix A.6 shows how cross-platform aggregation can reduce between-occupation selection when platform-selection parameters are imperfectly correlated. Appendix A.7 gives the monotonicity prediction used in the reweighting tests. Appendix A.8 gives supplementary remarks on adoption confounding, effective sample size, and the scope of the framework.

\subsection{A.1. Setup}\label{a.1.-setup}

All variables are residualized with respect to the controls $X_o$ and demeaned. The structural equation is

\begin{equation}
Y_o = \beta E_o + \varepsilon_o, \tag{A.1}
\end{equation}

with $\operatorname{Cov}(E_o, \varepsilon_o) = 0$. The researcher observes a platform-derived proxy, $E_{o,p}$, not true exposure $E_o$. The proxy is constructed from platform conversations and therefore reflects both the task composition of occupation $o$ and the representation of occupation $o$ in the platform user base.

The derivation assumes that the structural residual is orthogonal to the observed proxy after conditioning on the controls used in the estimating equation:

\begin{equation}
\operatorname{Cov}(\varepsilon_o, E_{o,p}) = 0. \tag{A.2}
\end{equation}

This assumption isolates the measurement problem in the regressor. It is the analogue of exogeneity for the diagnostic regressions used in the paper. The classical noise term in the proxy satisfies

\begin{equation}
\operatorname{Cov}(E_o, u_{o,p}) = 0. \tag{A.3}
\end{equation}

The within-occupation task-selection residual, $\eta_{o,p}$, may be correlated with $E_o$. That correlation is one reason the measurement error is non-classical.

\subsection{A.2. Platform Proxy Decomposition}\label{a.2.-platform-proxy-decomposition}

Let $q_{o,k}$ denote the workforce task-time share of task $k$ in occupation $o$, with $\sum_k q_{o,k} = 1$. Let $\tau_k$ denote the task-level AI capability score. True occupational exposure is

\begin{equation}
E_o = \sum_k q_{o,k} \tau_k. \tag{A.4}
\end{equation}

Let $q_{o,k,p}$ denote the within-occupation task share observed in platform $p$'s conversations for occupation $o$. Define the within-occupation task-selection parameter

\begin{equation}
\theta_{o,k,p} \equiv \frac{q_{o,k,p}}{q_{o,k}}. \tag{A.5}
\end{equation}

Let $f_p(o)$ be the platform conversation share of occupation $o$, and let $f(o)$ be the workforce employment share of occupation $o$. The between-occupation platform-selection parameter is

\begin{equation}
\psi_{o,p} \equiv \frac{f_p(o)}{f(o)}. \tag{A.6}
\end{equation}

A platform-derived exposure proxy that weights task capability by platform conversation density can be written as

\begin{equation}
E_{o,p} = \psi_{o,p} \sum_k q_{o,k,p} \tau_k + u_{o,p}. \tag{A.7}
\end{equation}

Substituting $q_{o,k,p} = \theta_{o,k,p} q_{o,k}$ gives

\begin{equation}
E_{o,p} = \psi_{o,p} \sum_k \theta_{o,k,p} q_{o,k} \tau_k + u_{o,p}. \tag{A.8}
\end{equation}

Add and subtract $\psi_{o,p} \sum_k q_{o,k} \tau_k$. Since $E_o = \sum_k q_{o,k} \tau_k$,

\begin{equation}
E_{o,p} = \psi_{o,p} E_o + \psi_{o,p} \sum_k (\theta_{o,k,p} - 1) q_{o,k} \tau_k + u_{o,p}. \tag{A.9}
\end{equation}

Define

\begin{equation}
\eta_{o,p} \equiv \psi_{o,p} \sum_k (\theta_{o,k,p} - 1) q_{o,k} \tau_k. \tag{A.10}
\end{equation}

Then

\begin{equation}
E_{o,p} = \psi_{o,p} E_o + \eta_{o,p} + u_{o,p}. \tag{A.11}
\end{equation}

This is equation (3) in the main text. The first term is the between-occupation component. The second term is the within-occupation task-selection residual. The third term is classical noise.

The parameter $\psi_{o,p}$ is identified from platform occupation shares and OEWS workforce shares. The parameter $\theta_{o,k,p}$ requires worker-representative task-time data measured on the same task taxonomy as the platform. Current survey data do not provide such a crosswalk at the \textit{O*NET} task level, so $\eta_{o,p}$ remains unobserved.

\subsection{A.3. OLS Probability Limit}\label{a.3.-ols-probability-limit}

The researcher estimates

\begin{equation}
Y_o = \beta_p E_{o,p} + \xi_o \tag{A.12}
\end{equation}

using the platform proxy. Let the linear projection of the proxy on true exposure be

\begin{equation}
E_{o,p} = \lambda_p E_o + v_{o,p}, \qquad \operatorname{Cov}(E_o, v_{o,p}) = 0, \tag{A.13}
\end{equation}

where

\begin{equation}
\lambda_p = \frac{\operatorname{Cov}(E_{o,p}, E_o)}{\operatorname{Var}(E_o)}. \tag{A.14}
\end{equation}

Let

\begin{equation}
\sigma^2_{v,p} \equiv \operatorname{Var}(v_{o,p}), \qquad \kappa_p \equiv \frac{\operatorname{Var}(E_o)}{\sigma^2_{v,p}}. \tag{A.15}
\end{equation}

The probability limit of the OLS coefficient is

\begin{equation}
\operatorname{plim} \widehat{\beta}_p = \frac{\operatorname{Cov}(Y_o, E_{o,p})}{\operatorname{Var}(E_{o,p})}. \tag{A.16}
\end{equation}

Using $Y_o = \beta E_o + \varepsilon_o$ and $\operatorname{Cov}(\varepsilon_o, E_{o,p}) = 0$,

\begin{equation}
\operatorname{Cov}(Y_o, E_{o,p}) = \beta \operatorname{Cov}(E_o, E_{o,p}) = \beta \lambda_p \operatorname{Var}(E_o). \tag{A.17}
\end{equation}

From equation (A.13),

\begin{equation}
\operatorname{Var}(E_{o,p}) = \lambda_p^2 \operatorname{Var}(E_o) + \sigma^2_{v,p}. \tag{A.18}
\end{equation}

Combining equations (A.16) through (A.18),

\begin{equation}
\operatorname{plim} \widehat{\beta}_p = \frac{\beta \lambda_p \operatorname{Var}(E_o)}{\lambda_p^2 \operatorname{Var}(E_o) + \sigma^2_{v,p}}. \tag{A.19}
\end{equation}

Dividing numerator and denominator by $\sigma^2_{v,p}$ yields

\begin{equation}
\operatorname{plim} \widehat{\beta}_p = \frac{\beta \lambda_p \kappa_p}{\lambda_p^2 \kappa_p + 1}. \tag{A.20}
\end{equation}

This is equation (4) in the main text.

\subsection{A.4. Special Cases}\label{a.4.-special-cases}

\textbf{Classical measurement error.} If the platform proxy is a classical noisy measure of true exposure, then $\lambda_p = 1$. Equation (A.20) becomes

\begin{equation}
\operatorname{plim} \widehat{\beta}_p = \frac{\beta \kappa_p}{\kappa_p + 1}, \tag{A.21}
\end{equation}

which is the standard attenuation formula.

\textbf{Overrepresentation of high-exposure occupations.} Suppose $\lambda_p > 1$. This occurs when the platform overrepresents occupations with high true exposure. With no residual projection error, $\sigma^2_{v,p} = 0$, equation (A.19) becomes

\begin{equation}
\operatorname{plim} \widehat{\beta}_p = \frac{\beta}{\lambda_p}. \tag{A.22}
\end{equation}

For $\beta > 0$, this is smaller than $\beta$. For $\beta < 0$, it is closer to zero in magnitude. With residual projection error, the coefficient is further scaled by the signal-to-noise ratio in equation (A.20). Thus, overrepresentation of high-exposure occupations changes the magnitude of the coefficient through $\lambda_p$ and generally produces attenuation relative to the structural coefficient when $\lambda_p > 1$ and the signal-to-noise ratio is sufficiently large.

\textbf{Underrepresentation of high-exposure occupations.} Suppose $0 < \lambda_p < 1$. With no residual projection error,

\begin{equation}
\operatorname{plim} \widehat{\beta}_p = \frac{\beta}{\lambda_p}, \tag{A.23}
\end{equation}

which exceeds $\beta$ in absolute value when $\beta \neq 0$. Underrepresentation of high-exposure occupations can therefore amplify the coefficient rather than attenuate it. This is not possible under classical measurement error, where attenuation is always toward zero.

\textbf{Platform-specific probability limits.} If two platforms $p$ and $p'$ have different $\lambda$ parameters or different residual variances, then

\[
\operatorname{plim} \widehat{\beta}_p \neq \operatorname{plim} \widehat{\beta}_{p'}
\]

even when both regressions use the same outcome, controls, sample, and estimator. This is the formal source of the cross-platform coefficient dispersion tested in the paper.

\subsection{A.5. Reweighting and the Maintained Ordering Condition}\label{a.5.-reweighting-and-the-maintained-ordering-condition}

Define the workforce-reweighted proxy

\begin{equation}
\widetilde{E}_{o,p} \equiv \frac{E_{o,p}}{\psi_{o,p}}. \tag{A.24}
\end{equation}

Using equation (A.11),

\begin{equation}
\widetilde{E}_{o,p} = E_o + \frac{\eta_{o,p}}{\psi_{o,p}} + \frac{u_{o,p}}{\psi_{o,p}}. \tag{A.25}
\end{equation}

Thus reweighting removes the between-occupation component exactly. If $\eta_{o,p} = 0$ and $u_{o,p} = 0$, then $\widetilde{E}_{o,p} = E_o$. With classical noise but no within-occupation task selection, the reweighted proxy is a classical noisy measure of true exposure. With within-occupation task selection, the reweighted proxy remains contaminated by $\eta_{o,p}/\psi_{o,p}$.

Let $\widetilde{\lambda}_p$ denote the projection coefficient of $\widetilde{E}_{o,p}$ on $E_o$:

\begin{equation}
\widetilde{\lambda}_p = \frac{\operatorname{Cov}(\widetilde{E}_{o,p}, E_o)}{\operatorname{Var}(E_o)}. \tag{A.26}
\end{equation}

Let $\widetilde{v}_{o,p} = \widetilde{E}_{o,p} - \widetilde{\lambda}_p E_o$ and $\widetilde{\kappa}_p = \operatorname{Var}(E_o)/\operatorname{Var}(\widetilde{v}_{o,p})$. The probability limit of OLS using the reweighted proxy is

\begin{equation}
\operatorname{plim} \widetilde{\beta}_p = \frac{\beta \widetilde{\lambda}_p \widetilde{\kappa}_p}{\widetilde{\lambda}_p^2 \widetilde{\kappa}_p + 1}. \tag{A.27}
\end{equation}

The reweighted coefficient identifies $\beta$ only under additional conditions. A sufficient condition is

\begin{equation}
\eta_{o,p} = 0, \qquad u_{o,p} = 0. \tag{A.28}
\end{equation}

More generally, reweighting removes the observable between-occupation component but leaves the within-occupation residual. The empirical bounds in the paper rely on the following maintained ordering condition.

\textbf{Maintained ordering condition.} Within-occupation task selection biases the coefficient in the same direction as between-occupation platform selection and does not reverse the ordering between the baseline and reweighted coefficients.

Under this maintained condition, the structural coefficient lies between the baseline platform-derived coefficient and the workforce-reweighted coefficient. The identified set is therefore

\begin{equation}
\beta \in \left[\min\{\widehat{\beta}_p, \widetilde{\beta}_p\}, \max\{\widehat{\beta}_p, \widetilde{\beta}_p\}\right]. \tag{A.29}
\end{equation}

When the outcome coefficient is reported in absolute-value terms, the corresponding interval is defined over magnitudes. The width of the interval measures the contribution of observable between-occupation platform selection to the reported coefficient.

The maintained ordering condition cannot be verified with current platform releases alone. It would require worker-representative task-time data and platform task-time data measured on the same task taxonomy. The condition is economically plausible when early adopters select high-payoff tasks both across occupations and within occupations, but the bounds reported in the paper remain conditional on this maintained assumption.

\subsection{A.6. Cross-Platform Aggregation}\label{a.6.-cross-platform-aggregation}

Let $p = 1, \ldots, P$ index platforms. For platform $p$,

\begin{equation}
E_{o,p} = \psi_{o,p} E_o + \eta_{o,p} + u_{o,p}. \tag{A.30}
\end{equation}

Consider an aggregate proxy

\begin{equation}
\bar{E}_o = \sum_{p=1}^P w_p E_{o,p}, \qquad w_p \geq 0, \qquad \sum_{p=1}^P w_p = 1. \tag{A.31}
\end{equation}

Substituting equation (A.30),

\begin{equation}
\bar{E}_o = \left(\sum_{p=1}^P w_p \psi_{o,p}\right) E_o + \sum_{p=1}^P w_p \eta_{o,p} + \sum_{p=1}^P w_p u_{o,p}. \tag{A.32}
\end{equation}

Define

\begin{equation}
\psi^w_o \equiv \sum_{p=1}^P w_p \psi_{o,p}, \qquad \eta^w_o \equiv \sum_{p=1}^P w_p \eta_{o,p}, \qquad u^w_o \equiv \sum_{p=1}^P w_p u_{o,p}. \tag{A.33}
\end{equation}

Then

\begin{equation}
\bar{E}_o = \psi^w_o E_o + \eta^w_o + u^w_o. \tag{A.34}
\end{equation}

The aggregate proxy has the same structural form as a single-platform proxy. It identifies true exposure only if $\psi^w_o = 1$ for all occupations and $\eta^w_o = 0$ for all occupations. Exact cancellation of platform selection is generally infeasible when the number of occupations exceeds the number of platforms. However, aggregation can reduce between-occupation selection variance when platform-specific selection parameters are imperfectly correlated.

To see this, let $\Delta_{o,p} = \psi_{o,p} - 1$ and $\Delta^w_o = \sum_p w_p \Delta_{o,p}$. Then

\begin{equation}
\operatorname{Var}(\Delta^w_o) = \sum_p w_p^2 \operatorname{Var}(\Delta_{o,p}) + 2 \sum_{p<q} w_p w_q \operatorname{Cov}(\Delta_{o,p}, \Delta_{o,q}). \tag{A.35}
\end{equation}

If selection parameters are imperfectly correlated across platforms, aggregation can reduce $\operatorname{Var}(\Delta^w_o)$ relative to a single platform. This result motivates cross-platform aggregation as a partial correction, not as a substitute for workforce-representative measurement.

\subsection{A.7. Skew-Attenuation Monotonicity}\label{a.7.-skew-attenuation-monotonicity}

The reweighting exercise implies a testable prediction. Platforms whose occupation distributions are closer to the workforce should exhibit smaller attenuation when reweighted.

Let between-occupation skew be measured by $\operatorname{Var}(\psi_{o,p})$ or by $\operatorname{Std}(\log \psi_{o,p})$. Suppose that, across platforms, the within-occupation task-selection residual $\eta_{o,p}$ is held fixed or is small relative to the between-occupation component. As $\operatorname{Var}(\psi_{o,p}) \to 0$ and $E[\psi_{o,p}] \to 1$, the proxy approaches

\begin{equation}
E_{o,p} \to E_o + \eta_{o,p} + u_{o,p}. \tag{A.36}
\end{equation}

If $\eta_{o,p}$ is also invariant across platforms, then the platform-specific movement in the probability limit is driven primarily by $\psi_{o,p}$. The reweighting correction approaches the identity as $\psi_{o,p} \to 1$. Hence the attenuation from reweighting is monotone in the distance between the platform user base and the workforce, provided the between-occupation component dominates the platform-specific variation.

This prediction is tested empirically in Section III.E by comparing reweighting attenuation across the composite and Microsoft Copilot measures. The platform closer to workforce employment shares attenuates less after reweighting.

\subsection{A.8. Supplementary Remarks}\label{a.8.-supplementary-remarks}

\textbf{Adoption confounding.} The main derivation treats the structural equation as a relationship between labor-market outcomes and true exposure. If adoption enters the structural equation separately, the relationship becomes

\begin{equation}
Y_o = \beta E_o + \rho A_o + \varepsilon_o, \tag{A.37}
\end{equation}

where $A_o$ is AI adoption. If the platform proxy is correlated with adoption, then the probability limit includes an additional term:

\begin{equation}
\operatorname{plim} \widehat{\beta}_p = \frac{\beta \operatorname{Cov}(E_o, E_{o,p}) + \rho \operatorname{Cov}(A_o, E_{o,p})}{\operatorname{Var}(E_{o,p})}. \tag{A.38}
\end{equation}

The main text focuses on the measurement problem in $E_{o,p}$, holding the downstream specification fixed. Equation (A.38) shows that adoption can create an additional channel when adoption is itself an omitted determinant of outcomes. This does not remove the platform-selection problem. It adds another reason to distinguish capability, adoption, and conditional use.

\textbf{Effective sample size.} The measurement-error problem is not governed by the number of individuals in the downstream ACS regression. The exposure measure varies at the occupation level. Thus, the relevant variation is the number of distinct occupational exposure cells, not the number of person-year observations. Adding more individuals improves precision for outcome means conditional on occupation, but it does not change the probability limit generated by a selected occupation-level proxy.

\textbf{Scope of the framework.} The framework applies to several sources of variation used in the paper. It applies across platforms, such as Anthropic, OpenAI, Microsoft, and survey-based measures. It applies across product channels within the same platform, such as Claude.ai consumer use and the enterprise API. It applies across waves within the same platform when product launches or pricing changes shift the user base. It also applies to comparisons between survey and platform data, since each source represents a different population frame and a different measurement unit.

\textbf{Interpretation of platform data.} The framework does not imply that platform logs are uninformative. Platform logs are informative about task use among observed users. The problem arises when observed platform use is interpreted as workforce exposure without adjustment. The reweighting procedure removes the part of this selection that is observed in occupation shares. The remaining component requires worker-task data that observe both platform users and nonusers on a common task taxonomy.

\clearpage
\section*{Online Appendix B. Supplementary Tables and Figures}
\addcontentsline{toc}{section}{Online Appendix B}
This online appendix contains supplementary tables (B.1 through B.12) and supplementary figures (B.1 through B.7) referenced in the main text.
\vspace{1em}
\clearpage

\begin{landscape}
\begingroup
\scriptsize
\setlength{\tabcolsep}{2pt}
\renewcommand{\arraystretch}{0.78}
\renewcommand{\baselinestretch}{0.85}\selectfont
\textbf{Table B.1. Variable Definitions, Data Sources, and Summary Statistics}

\emph{\textbf{Panel A. American Community Survey analytical sample
variables: definitions and person-weighted summary statistics
(wage-and-salary workers, ages 16 to 64, 2015 to 2024)}}

{\def\LTcaptype{none} 
\begin{longtable}[]{@{}
  >{\raggedright\arraybackslash}p{0.1167\linewidth}
  >{\centering\arraybackslash}p{0.4167\linewidth}
  >{\centering\arraybackslash}p{0.1167\linewidth}
  >{\centering\arraybackslash}p{0.1167\linewidth}
  >{\centering\arraybackslash}p{0.1167\linewidth}
  >{\centering\arraybackslash}p{0.1167\linewidth}@{}}
\toprule\noalign{}
\begin{minipage}[b]{\linewidth}\raggedright
\textbf{Variable}
\end{minipage} & \begin{minipage}[b]{\linewidth}\centering
\textbf{Definition}
\end{minipage} & \begin{minipage}[b]{\linewidth}\centering
\textbf{Mean}
\end{minipage} & \begin{minipage}[b]{\linewidth}\centering
\textbf{SD}
\end{minipage} & \begin{minipage}[b]{\linewidth}\centering
\textbf{Min}
\end{minipage} & \begin{minipage}[b]{\linewidth}\centering
\textbf{Max}
\end{minipage} \\
\midrule\noalign{}
\endhead
\bottomrule\noalign{}
\endlastfoot
year & Survey year (2015 to 2024) & n/a & n/a & 2015 & 2024 \\
statefip & State FIPS code (51 levels) & 51 levels & n/a & 1 & 56 \\
occsoc & Six-digit SOC occupation code (SOC-2018) & 553 levels & n/a &
111021 & 999920 \\
perwt & Person weight (sample-to-population scaling) & 189.86 & 166.22 &
1.00 & 3,773 \\
post & Post-2022 indicator (= 1 if year ≥ 2023) & 0.204 & 0.403 & 0.000
& 1.00 \\
employed & Binary employment indicator (= 1 if employed; ACS empstat) &
0.844 & 0.363 & 0.000 & 1.00 \\
age & Age in years (continuous; restricted to 16 to 64) & 39.17 & 13.47
& 16.00 & 64.00 \\
age2 & Age squared & 1,716 & 1,094 & 256.00 & 4,096 \\
female & Female indicator (ACS sex) & 0.507 & 0.500 & 0.000 & 1.00 \\
married & Currently married indicator (ACS marst) & 0.474 & 0.499 &
0.000 & 1.00 \\
hsgrad & High-school graduate, no college (ACS educ) & 0.334 & 0.472 &
0.000 & 1.00 \\
scoll & Some college, no bachelor\textquotesingle s (ACS educ) & 0.253 &
0.434 & 0.000 & 1.00 \\
bachelor\_higher & Bachelor\textquotesingle s degree or higher (ACS
educ) & 0.335 & 0.472 & 0.000 & 1.00 \\
nchild & Number of own children in household (ACS nchild) & 0.756 & 1.11
& 0.000 & 9.00 \\
nhwhite & Non-Hispanic white (ACS race × hispan) & 0.591 & 0.492 & 0.000
& 1.00 \\
nhblack & Non-Hispanic Black (ACS race × hispan) & 0.125 & 0.331 & 0.000
& 1.00 \\
hisp & Hispanic, any race (ACS hispan) & 0.182 & 0.386 & 0.000 & 1.00 \\
disabled & Adult with disability, union of six ACS items (diffrem,
diffphys, diffmob, diffcare, diffeye, diffhear) & 0.075 & 0.264 & 0.000
& 1.00 \\
incwage & Wage and salary income, current-year USD (ACS) & 46,690 &
58,368 & 0.000 & 907,000 \\
\end{longtable}
}

\emph{\textbf{Panel B. National Health Interview Survey Sample Adult
cross-domain outcome variables: definitions and source variable codes
(2017 to 2018; N = 49,066; aggregated at Standard Occupational
Classification major-group level)}}

{\def\LTcaptype{none} 
\begin{longtable}[]{@{}
  >{\raggedright\arraybackslash}p{0.2250\linewidth}
  >{\centering\arraybackslash}p{0.5500\linewidth}
  >{\centering\arraybackslash}p{0.2250\linewidth}@{}}
\toprule\noalign{}
\begin{minipage}[b]{\linewidth}\raggedright
\textbf{Variable}
\end{minipage} & \begin{minipage}[b]{\linewidth}\centering
\textbf{Definition}
\end{minipage} & \begin{minipage}[b]{\linewidth}\centering
\textbf{Source}
\end{minipage} \\
\midrule\noalign{}
\endhead
\bottomrule\noalign{}
\endlastfoot
k6\_score & Kessler 6-item psychological-distress scale (0 to 24) & NHIS
KESSLER6 \\
k6\_distress & =1 if K6 score \textgreater= 13 (clinically meaningful
distress) & NHIS KESSLER6 \textgreater= 13 \\
smoke\_ever & =1 if ever smoked \textgreater= 100 cigarettes lifetime &
NHIS SMOKEV = 2 \\
smoke\_current & =1 if current daily or some-day smoker & NHIS
SMOKESTATUS2 in \{1,2\} \\
drinker\_pastyr & =1 if any alcohol consumption in past 12 months & NHIS
ALCSTAT1 \textgreater= 2 \\
drinks\_per\_day & Mean drinks per day \textbar{} past-year drinker &
NHIS ALCDAYSYR * ALCAMT / 365 \\
binge\_any & =1 if any binge-drinking episode in past year & NHIS
ALC5UPYR \textgreater= 1 \\
sleep\_short & =1 if usual sleep \textless{} 6 hours per night & NHIS
HRSLEEP \textless{} 6 \\
self\_rated\_poor & =1 if self-rated health is fair" or "poor"" & NHIS
HEALTH in \{4,5\} \\
\end{longtable}
}

\emph{\textbf{Panel C. Anthropic Economic Index wave characteristics and
Computer \& Mathematical share}}

{\def\LTcaptype{none} 
\begin{longtable}[]{@{}
  >{\raggedright\arraybackslash}p{0.3667\linewidth}
  >{\centering\arraybackslash}p{0.1267\linewidth}
  >{\centering\arraybackslash}p{0.1267\linewidth}
  >{\centering\arraybackslash}p{0.1267\linewidth}
  >{\centering\arraybackslash}p{0.1267\linewidth}
  >{\centering\arraybackslash}p{0.1267\linewidth}@{}}
\toprule\noalign{}
\begin{minipage}[b]{\linewidth}\raggedright
\textbf{Wave}
\end{minipage} & \begin{minipage}[b]{\linewidth}\centering
\textbf{Period}
\end{minipage} & \begin{minipage}[b]{\linewidth}\centering
\textbf{Model}
\end{minipage} & \begin{minipage}[b]{\linewidth}\centering
\textbf{Product launch}
\end{minipage} & \begin{minipage}[b]{\linewidth}\centering
\textbf{SOC cells}
\end{minipage} & \begin{minipage}[b]{\linewidth}\centering
\textbf{C\&M share}
\end{minipage} \\
\midrule\noalign{}
\endhead
\bottomrule\noalign{}
\endlastfoot
Wave 1 & Dec 2024 & Sonnet 3.5 & n/a & 981 & 37\% \\
Wave 2 & Feb 2025 & Sonnet 3.7 & n/a & 999 & 40\% \\
Wave 3 & Aug 2025 & Sonnet 4 & Claude Code launch & 928 & 39\% \\
Wave 4 & Nov 2025 & Sonnet 4.5 & Projects expansion & 974 & 36\% \\
Wave 5 & Feb 2026 & Opus 4.6 & n/a & 964 & 32\% \\
\end{longtable}
}

\emph{\textbf{Panel D. Data sources}}

{\def\LTcaptype{none} 
\begin{longtable}[]{@{}
  >{\raggedright\arraybackslash}p{0.4500\linewidth}
  >{\centering\arraybackslash}p{0.1833\linewidth}
  >{\centering\arraybackslash}p{0.1833\linewidth}
  >{\centering\arraybackslash}p{0.1833\linewidth}@{}}
\toprule\noalign{}
\begin{minipage}[b]{\linewidth}\raggedright
\textbf{Source}
\end{minipage} & \begin{minipage}[b]{\linewidth}\centering
\textbf{Type}
\end{minipage} & \begin{minipage}[b]{\linewidth}\centering
\textbf{Period}
\end{minipage} & \begin{minipage}[b]{\linewidth}\centering
\textbf{Sample}
\end{minipage} \\
\midrule\noalign{}
\endhead
\bottomrule\noalign{}
\endlastfoot
AEI Claude.ai (Anthropic, consumer) & Platform usage & Dec 2024-Feb 2026
& 5 waves \\
AEI 1P API (Anthropic, enterprise) & Platform usage & Aug 2025-Feb 2026
& 3 waves \\
OpenAI ChatGPT (\hyperlink{Chatterji2025}{Chatterji et al. 2025}) & Platform usage & May 2024-Jun 2025 &
5-group occupation distribution \\
Microsoft Copilot (\hyperlink{Tomlinson2025}{Tomlinson et al. 2025}) & Platform usage (consumer) & Sept
2024 & AI applicability at SOC-6 \\
BBD-RPS (Bick, Blandin, and Deming 2026) & Population survey & Dec 2024 &
Per-occupation GenAI use rate \\
BLS OES & Establishment survey & May 2024 & \textasciitilde1.1M
establishments \\
BLS Employment Projections & Administrative & 2024-2034 & 10-year
occupation projections \\
IPUMS-USA ACS 1-year (analytic outcome panel) & Household survey &
2015-2024 & 13,104,072 person-years (16-64) \\
\hyperlink{Eloundou2024}{\hyperlink{Eloundou2024}{Eloundou et al. 2024}} GPT-4 binary $\beta$ rubric & Task-level capability &
2023 & Per-task AI applicability \\
NHIS Sample Adult (cross-domain §VI) & Household survey & 2017-2018 &
49,066 adults \\
\end{longtable}
}

\emph{\textbf{Panel E. Anthropic Economic Index coverage and Standard
Occupational Classification match rate}}

{\def\LTcaptype{none} 
\begin{longtable}[]{@{}
  >{\raggedright\arraybackslash}p{0.5000\linewidth}
  >{\centering\arraybackslash}p{0.5000\linewidth}@{}}
\toprule\noalign{}
\begin{minipage}[b]{\linewidth}\raggedright
\textbf{Statistic}
\end{minipage} & \begin{minipage}[b]{\linewidth}\centering
\textbf{Value}
\end{minipage} \\
\midrule\noalign{}
\endhead
\bottomrule\noalign{}
\endlastfoot
Total BLS SOC2018 detailed cells (OES May 2024) & 831 \\
AEI Claude.ai nonzero NEM cells (range W1-W5) & 928-999 \\
Microsoft Copilot SOC cells (\hyperlink{Tomlinson2025}{Tomlinson et al. 2025}) & 1336 \\
Distinct SOC-6 codes in ACS analytic sample & 553 \\
ACS occsoc codes matched to BLS SOC2018 & 390 of 553 (70.5\% by code) \\
ACS person-year-weighted match rate to BLS SOC2018 & 68.06\% (9,900,621
of 13,104,072) \\
\end{longtable}
}

\emph{\textbf{Panel F. National Health Interview Survey Sample Adult
2017 to 2018 cross-domain outcomes}}

{\def\LTcaptype{none} 
\begin{longtable}[]{@{}
  >{\raggedright\arraybackslash}p{0.4400\linewidth}
  >{\centering\arraybackslash}p{0.1400\linewidth}
  >{\centering\arraybackslash}p{0.1400\linewidth}
  >{\centering\arraybackslash}p{0.1400\linewidth}
  >{\centering\arraybackslash}p{0.1400\linewidth}@{}}
\toprule\noalign{}
\begin{minipage}[b]{\linewidth}\raggedright
\textbf{Variable}
\end{minipage} & \begin{minipage}[b]{\linewidth}\centering
\textbf{Mean}
\end{minipage} & \begin{minipage}[b]{\linewidth}\centering
\textbf{SD}
\end{minipage} & \begin{minipage}[b]{\linewidth}\centering
\textbf{Min}
\end{minipage} & \begin{minipage}[b]{\linewidth}\centering
\textbf{Max}
\end{minipage} \\
\midrule\noalign{}
\endhead
\bottomrule\noalign{}
\endlastfoot
K6 distress score (0 to 24) & 2.777 & 0.436 & 1.992 & 3.762 \\
K6 severe distress (\textgreater= 13) & 0.036 & 0.015 & 0.009 & 0.064 \\
Ever smoker & 0.371 & 0.077 & 0.239 & 0.539 \\
Current smoker & 0.141 & 0.060 & 0.044 & 0.251 \\
Drinker (past year) & 0.676 & 0.077 & 0.549 & 0.846 \\
Drinks per day \textbar{} past-year drinker & 2.327 & 0.392 & 1.804 &
3.246 \\
Any binge-drinking episode & 0.387 & 0.056 & 0.289 & 0.498 \\
Short sleep (\textless{} 6 hrs) & 0.096 & 0.025 & 0.043 & 0.153 \\
\end{longtable}
}

\needspace{6\baselineskip}
\textbf{Notes:} \emph{Panels A and B document the definitions and weighted summary statistics for every variable used in the analysis on the canonical American Community Survey analytical sample (wage-and-salary workers ages 16 to 64, 2015 to 2024) and on the cross-domain National Health Interview Survey Sample Adult sample. Panel C summarizes the five Anthropic Economic Index waves (period, model version, notable product launch, number of nonzero Standard Occupational Classification 6-digit cells, and the share of conversations attributed to Computer and Mathematical occupations). Panel D reports the complete data-source inventory. Panel E reports the occupation-crosswalk coverage between American Community Survey occupation codes and 2018 Standard Occupational Classification 6-digit codes. Panel F reports cross-domain outcome summary statistics from the National Health Interview Survey Sample Adult file (2017 to 2018) used in Section IV. Statistics are person-weighted
using the American Community Survey person weight (perwt). Mean and
standard deviation are reported for binary and continuous variables; for
categorical and identifier variables, the level count is reported in
place of mean and standard deviation. The disability indicator is
constructed as the union of the six American Community Survey disability
items (cognitive, ambulatory, hearing, vision, self-care, independent
living difficulties: diffrem, diffphys, diffhear, diffeye, diffcare,
diffmob). The post indicator equals one for survey years 2023 and 2024
(zero otherwise). Min and max report the observed minimum and maximum
values within the analytical sample. n/a denotes not applicable
(typically for categorical or identifier variables where mean/SD are not
meaningful). Abbreviations: SOC = Standard Occupational Classification;
AEI = Anthropic Economic Index; CAI = Claude.ai consumer; API =
first-party Application Programming Interface enterprise; Composite = AI
exposure score constructed following \hyperlink{Massenkoff2026}{Massenkoff and McCrory (2026)}; BBD
= Bick-Blandin-Deming Survey of Working Arrangements and Attitudes; BLS
= Bureau of Labor Statistics; OES = Occupational Employment and Wage
Statistics; CKK = Chen, Kane, Kozlowski et al. (2025); ACS = American
Community Survey; CPS-MORG = Current Population Survey-Outgoing Rotation
Group.}

\endgroup
\end{landscape}
\clearpage

\begin{landscape}
\begingroup
\tiny
\setlength{\tabcolsep}{2pt}
\renewcommand{\arraystretch}{0.7}
\renewcommand{\baselinestretch}{0.85}\selectfont
\textbf{Table B.2. Cross-Platform Composition and Descriptive Correlates by SOC Major Group}

\emph{\textbf{Panel A. Cross-platform composition: conversation shares, use rates, and workforce share}}

{\def\LTcaptype{none} 
\begin{longtable}[]{@{}
  >{\raggedright\arraybackslash}p{0.26\linewidth}
  >{\centering\arraybackslash}p{0.10\linewidth}
  >{\centering\arraybackslash}p{0.10\linewidth}
  >{\centering\arraybackslash}p{0.10\linewidth}
  >{\centering\arraybackslash}p{0.13\linewidth}
  >{\centering\arraybackslash}p{0.10\linewidth}
  >{\centering\arraybackslash}p{0.10\linewidth}@{}}
\toprule\noalign{}
\begin{minipage}[b]{\linewidth}\raggedright
\textbf{SOC major group}
\end{minipage} & \begin{minipage}[b]{\linewidth}\centering
\textbf{Claude.ai W5 (\%)}
\end{minipage} & \begin{minipage}[b]{\linewidth}\centering
\textbf{1P API W5 (\%)}
\end{minipage} & \begin{minipage}[b]{\linewidth}\centering
\textbf{ChatGPT pct\_work}
\end{minipage} & \begin{minipage}[b]{\linewidth}\centering
\textbf{Microsoft Copilot score}
\end{minipage} & \begin{minipage}[b]{\linewidth}\centering
\textbf{BBD GenAI use (\%)}
\end{minipage} & \begin{minipage}[b]{\linewidth}\centering
\textbf{BLS workforce (\%)}
\end{minipage} \\
\midrule\noalign{}
\endhead
\bottomrule\noalign{}
\endlastfoot
Computer and Mathematical & 32.3 & 51.7 & 57.0 & 0.29 & 47.9 & 3.4 \\
Educational Instruction and Library & 13.5 & 3.1 & 44.0 & 0.21 & 33.5 &
5.8 \\
Arts, Design, Entertainment, Sports, Media & 10.4 & 5.2 & 44.0 & 0.24 &
31.0 & 1.4 \\
Office and Administrative Support & 9.4 & 17.6 & 40.0 & 0.26 & 18.5 &
11.8 \\
Life, Physical, and Social Science & 6.3 & 4.4 & 48.0 & 0.19 & n/a &
0.9 \\
Sales and Related & 4.9 & 4.1 & 40.0 & 0.29 & 20.9 & 8.7 \\
Business and Financial & 4.8 & 3.5 & 50.0 & 0.24 & 40.0 & 6.7 \\
Management & 4.3 & 3.4 & 50.0 & 0.14 & 45.2 & 7.1 \\
Community and Social Service & 3.2 & 0.5 & 44.0 & 0.25 & n/a & 1.7 \\
Healthcare Practitioners and Technical & 3.0 & 1.0 & 44.0 & 0.12 & 30.2
& 6.2 \\
Architecture and Engineering & 2.0 & 1.7 & 48.0 & 0.22 & n/a & 1.7 \\
Personal Care and Service & 1.2 & 0.4 & 40.0 & 0.18 & n/a & 2.0 \\
Food Preparation and Serving & 0.7 & 0.1 & 40.0 & 0.17 & n/a & 8.8 \\
Production & 0.7 & 1.2 & 40.0 & 0.11 & n/a & 5.7 \\
Installation, Maintenance, and Repair & 0.7 & 0.4 & 40.0 & 0.10 & n/a &
3.9 \\
Legal & 0.6 & 0.2 & 44.0 & 0.13 & n/a & 0.8 \\
Farming, Fishing, and Forestry & 0.5 & 0.1 & 40.0 & 0.06 & n/a & 0.3 \\
Protective Service & 0.4 & 0.5 & 40.0 & 0.12 & n/a & 2.4 \\
Healthcare Support & 0.3 & 0.2 & 40.0 & 0.05 & n/a & 4.8 \\
Transportation and Material Moving & 0.2 & 0.2 & 40.0 & 0.10 & n/a &
8.9 \\
Construction and Extraction & 0.2 & 0.3 & 40.0 & 0.07 & n/a & 4.1 \\
Building and Grounds Cleaning & 0.2 & 0.1 & 40.0 & 0.08 & n/a & 2.9 \\
\end{longtable}
}

\emph{\textbf{Panel B. Descriptive correlates by SOC major group: education, wages, disability, and platform exposure}}

{\def\LTcaptype{none} 
\begin{longtable}[]{@{}
  >{\raggedright\arraybackslash}p{0.30\linewidth}
  >{\centering\arraybackslash}p{0.12\linewidth}
  >{\centering\arraybackslash}p{0.13\linewidth}
  >{\centering\arraybackslash}p{0.14\linewidth}
  >{\centering\arraybackslash}p{0.15\linewidth}
  >{\centering\arraybackslash}p{0.08\linewidth}@{}}
\toprule\noalign{}
\begin{minipage}[b]{\linewidth}\raggedright
\textbf{SOC major group}
\end{minipage} & \begin{minipage}[b]{\linewidth}\centering
\textbf{BA+ share}
\end{minipage} & \begin{minipage}[b]{\linewidth}\centering
\textbf{Wage (2022 \$)}
\end{minipage} & \begin{minipage}[b]{\linewidth}\centering
\textbf{Disability share}
\end{minipage} & \begin{minipage}[b]{\linewidth}\centering
\textbf{C\&M conv. share}
\end{minipage} & \begin{minipage}[b]{\linewidth}\centering
\textbf{psi}
\end{minipage} \\
\midrule\noalign{}
\endhead
\bottomrule\noalign{}
\endlastfoot
Management & 0.57 & 92.3 & 0.05 & 0.075 & 0.55 \\
Business and Financial & 0.68 & 84.3 & 0.05 & 0.087 & 0.99 \\
Computer and Mathematical & 0.67 & 97.4 & 0.05 & 0.32 & 9.41 \\
Architecture and Engineering & 0.72 & 94.5 & 0.04 & 0.052 & 1.86 \\
Life, Physical, Social Science & 0.87 & 82.2 & 0.04 & 0.026 & 2.36 \\
Community and Social Service & 0.73 & 47.5 & 0.07 & 0.018 & 0.86 \\
Legal & 0.5 & 54.0 & 0.06 & 0.027 & 2.45 \\
Education, Training, Library & 0.75 & 45.6 & 0.06 & 0.046 & 0.78 \\
Arts, Design, Entertainment & 0.58 & 52.8 & 0.06 & 0.043 & 2.0 \\
Healthcare Practitioners & 0.58 & 81.2 & 0.05 & 0.024 & 0.32 \\
Healthcare Support & 0.13 & 26.6 & 0.1 & 0.005 & 0.11 \\
Protective Service & 0.27 & 58.1 & 0.08 & 0.006 & 0.27 \\
Food Preparation & 0.08 & 17.2 & 0.09 & 0.005 & 0.08 \\
Building, Grounds Cleaning & 0.06 & 23.9 & 0.1 & 0.002 & 0.06 \\
Personal Care & 0.16 & 20.0 & 0.08 & 0.005 & 0.18 \\
Sales & 0.25 & 46.0 & 0.08 & 0.024 & 0.3 \\
Office and Administrative & 0.22 & 36.4 & 0.08 & 0.046 & 0.39 \\
Farming, Fishing, Forestry & 0.11 & 33.3 & 0.1 & 0.001 & 0.13 \\
Construction and Extraction & 0.06 & 45.5 & 0.08 & 0.005 & 0.1 \\
Installation, Maintenance, Repair & 0.08 & 51.9 & 0.08 & 0.011 & 0.32 \\
Production & 0.09 & 42.8 & 0.09 & 0.02 & 0.32 \\
Transportation, Material Moving & 0.09 & 35.3 & 0.1 & 0.011 & 0.13 \\
\end{longtable}
}

\needspace{6\baselineskip}
\textbf{Notes:} \emph{Panel A is a wide-format companion to Table 3 (main text), restricted to the major groups with the highest platform conversation share or use rate. Panel B reports descriptive correlates of SOC major groups, including the share of workers with a Bachelor's degree or higher (BA+ share), annual wage and salary income in 2022 dollars (thousands), the share of adults with disability under the union-of-six American Community Survey disability items, the Anthropic Economic Index Wave 5 share of conversations attributed to the Computer and Mathematical major group (C\&M conv. share), and the Baseline Composite (Anthropic) exposure score (psi) from \hyperlink{Eloundou2024}{Eloundou et al. (2024)}. All Panel B statistics are person-weighted using the American Community Survey person weight (perwt). Each row reports a 2-digit Standard
Occupational Classification major group with the per-platform
conversation share, use rate, or applicability score. Claude.ai W5 (\%)
and 1P API W5 (\%) are Wave 5 Anthropic Economic Index conversation
shares for the consumer Claude.ai and the first-party Application
Programming Interface enterprise platforms, respectively. ChatGPT
pct\_work is the percentage-of-work-tasks measure from \hyperlink{Chatterji2025}{\hyperlink{Chatterji2025}{Chatterji et al. (2025)}}. Microsoft Copilot score is the artificial-intelligence
applicability score (\hyperlink{Tomlinson2025}{\hyperlink{Tomlinson2025}{Tomlinson et al. 2025}}) employment-weighted to
Standard Occupational Classification major group; values range from 0 to
1. BBD GenAI use (\%) is the total generative artificial-intelligence
use rate from \hyperlink{Bick2026}{Bick, Blandin, and Deming (2026)} Real-Time Population Survey
December 2024; n/a denotes major groups not surveyed. BLS
workforce (\%) is the May 2024 Occupational Employment and Wage
Statistics employment share. Rows are ordered by descending Claude.ai
Wave 5 share. Abbreviations: SOC = Standard Occupational Classification;
AEI = Anthropic Economic Index; CAI = Claude.ai consumer; API =
first-party Application Programming Interface enterprise; Composite = AI
exposure score constructed following \hyperlink{Massenkoff2026}{Massenkoff and McCrory (2026)}; BBD
= Bick-Blandin-Deming Survey of Working Arrangements and Attitudes; BLS
= Bureau of Labor Statistics; OES = Occupational Employment and Wage
Statistics; CKK = Chen, Kane, Kozlowski et al. (2025); ACS = American
Community Survey; CPS-MORG = Current Population Survey-Outgoing Rotation
Group.}

\endgroup
\end{landscape}
\clearpage
\begingroup
\scriptsize
\setlength{\tabcolsep}{2pt}
\renewcommand{\arraystretch}{0.78}
\renewcommand{\baselinestretch}{0.85}\selectfont
\textbf{Table B.3. ACS to SOC 2018 Crosswalk Coverage}

{\def\LTcaptype{none} 
\begin{longtable}[]{@{}
  >{\raggedright\arraybackslash}p{0.4500\linewidth}
  >{\centering\arraybackslash}p{0.1833\linewidth}
  >{\centering\arraybackslash}p{0.1833\linewidth}
  >{\centering\arraybackslash}p{0.1833\linewidth}@{}}
\toprule\noalign{}
\begin{minipage}[b]{\linewidth}\raggedright
\textbf{Major group (SOC-2)}
\end{minipage} & \begin{minipage}[b]{\linewidth}\centering
\textbf{Distinct SOC-6}
\end{minipage} & \begin{minipage}[b]{\linewidth}\centering
\textbf{ACS records}
\end{minipage} & \begin{minipage}[b]{\linewidth}\centering
\textbf{Share (\%)}
\end{minipage} \\
\midrule\noalign{}
\endhead
\bottomrule\noalign{}
\endlastfoot
Management (11) & 31 & 998599 & 7.6205 \\
Business and Financial (13) & 30 & 795077 & 6.0674 \\
Computer and Mathematical (15) & 24 & 452665 & 3.4544 \\
Architecture and Engineering (17) & 19 & 214489 & 1.6368 \\
Life, Physical, and Social Science (19) & 19 & 103348 & 0.7887 \\
Community and Social Service (21) & 17 & 254569 & 1.9427 \\
Legal (23) & 5 & 60588 & 0.4624 \\
Educational Instruction and Library (25) & 13 & 989536 & 7.5514 \\
Arts, Design, Entertainment, Sports, Media (27) & 34 & 227508 &
1.7362 \\
Healthcare Practitioners and Technical (29) & 45 & 871276 & 6.6489 \\
Healthcare Support (31) & 13 & 415051 & 3.1673 \\
Protective Service (33) & 18 & 311014 & 2.3734 \\
Food Preparation and Serving (35) & 14 & 934168 & 7.1288 \\
Building and Grounds Maintenance (37) & 7 & 239148 & 1.825 \\
Personal Care and Service (39) & 24 & 356659 & 2.7217 \\
Sales (41) & 19 & 1396284 & 10.6553 \\
Office and Administrative Support (43) & 54 & 1710984 & 13.0569 \\
Farming, Fishing, and Forestry (45) & 7 & 21709 & 0.1657 \\
Construction and Extraction (47) & 33 & 606107 & 4.6253 \\
Installation, Maintenance, and Repair (49) & 29 & 391864 & 2.9904 \\
Production (51) & 62 & 615100 & 4.694 \\
Transportation and Material Moving (53) & 31 & 1070850 & 8.1719 \\
n/a (55) & 4 & 67479 & 0.5149 \\
n/a (99) & 1 & 0 & 0.0 \\
\end{longtable}
}

\needspace{6\baselineskip}
\textbf{Notes:} \emph{This table documents the crosswalk coverage
between American Community Survey occupation codes (occsoc) and 2018
Standard Occupational Classification 6-digit codes used throughout the
analysis. Each row reports a 2-digit Standard Occupational
Classification major group (with the 2-digit code in parentheses), the
number of distinct 6-digit Standard Occupational Classification 2018
codes observed in the American Community Survey 2015 to 2024 canonical
analytical sample, the unweighted record count (person-years), and the
share of the analytical sample. The Military (55) row reflects the small
set of military occupations retained in the wage-and-salary worker
analytical sample. The \textquotesingle n/a (99)\textquotesingle{} row
is a residual American Community Survey occupation category that does
not map to a specific Standard Occupational Classification major group;
it is excluded from the regression analysis. n/a denotes not applicable.
Abbreviations: SOC = Standard Occupational Classification; AEI =
Anthropic Economic Index; CAI = Claude.ai consumer; API = first-party
Application Programming Interface enterprise; Composite = AI exposure
score constructed following \hyperlink{Massenkoff2026}{Massenkoff and McCrory (2026)}; BBD =
Bick-Blandin-Deming Survey of Working Arrangements and Attitudes; BLS =
Bureau of Labor Statistics; OES = Occupational Employment and Wage
Statistics; CKK = Chen, Kane, Kozlowski et al. (2025); ACS = American
Community Survey; CPS-MORG = Current Population Survey-Outgoing Rotation
Group.}

\endgroup

\clearpage

\begingroup
\scriptsize
\setlength{\tabcolsep}{2pt}
\renewcommand{\arraystretch}{0.78}
\renewcommand{\baselinestretch}{0.85}\selectfont
\textbf{Table B.4. Wave-Pair Exposure Quartile Transitions}

{\def\LTcaptype{none} 
\begin{longtable}[]{@{}
  >{\raggedright\arraybackslash}p{0.4400\linewidth}
  >{\centering\arraybackslash}p{0.1400\linewidth}
  >{\centering\arraybackslash}p{0.1400\linewidth}
  >{\centering\arraybackslash}p{0.1400\linewidth}
  >{\centering\arraybackslash}p{0.1400\linewidth}@{}}
\toprule\noalign{}
\begin{minipage}[b]{\linewidth}\raggedright
\textbf{Wave pair}
\end{minipage} & \begin{minipage}[b]{\linewidth}\centering
\textbf{N occupations}
\end{minipage} & \begin{minipage}[b]{\linewidth}\centering
\textbf{Same quartile}
\end{minipage} & \begin{minipage}[b]{\linewidth}\centering
\textbf{One-quartile move}
\end{minipage} & \begin{minipage}[b]{\linewidth}\centering
\textbf{Two-plus-quartile move}
\end{minipage} \\
\midrule\noalign{}
\endhead
\bottomrule\noalign{}
\endlastfoot
Waves 1 to 2 & 715 & 68.8\% & 29.9\% & 1.3\% \\
Waves 1 to 3 & 670 & 58.5\% & 37.5\% & 4.0\% \\
Waves 1 to 4 & 685 & 60.3\% & 35.3\% & 4.4\% \\
Waves 1 to 5 & 689 & 58.2\% & 37.4\% & 4.4\% \\
Waves 2 to 3 & 670 & 62.7\% & 34.3\% & 3.0\% \\
Waves 2 to 4 & 681 & 63.4\% & 33.6\% & 2.9\% \\
Waves 2 to 5 & 684 & 59.9\% & 37.0\% & 3.1\% \\
Waves 3 to 4 & 669 & 72.5\% & 26.3\% & 1.2\% \\
Waves 3 to 5 & 671 & 69.2\% & 29.1\% & 1.8\% \\
Waves 4 to 5 & 702 & 85.5\% & 14.5\% & 0.0\% \\
\end{longtable}
}

\needspace{6\baselineskip}
\textbf{Notes:} \emph{This table assesses the temporal stability of
platform-visible occupational exposure rankings by reporting
wave-to-wave quartile transition rates across the five Anthropic
Economic Index waves. Each row reports the share of 6-digit Standard
Occupational Classification occupations remaining in the same
employment-weighted exposure quartile, moving by one quartile, or moving
by two or more quartiles between two Anthropic Economic Index waves.
Quartiles are constructed within wave using the employment-weighted
distribution of the Anthropic Economic Index conversation share, with
weights from Bureau of Labor Statistics May 2024 Occupational Employment
and Wage Statistics. N is the number of 6-digit Standard Occupational
Classification occupations with valid wave-pair exposure data. High
wave-pair movement (low same-quartile shares, high one- and
two-plus-quartile move shares) indicates substantial month-to-month
instability in occupation-level exposure rankings, motivating concern
about using a single wave\textquotesingle s platform-visible
distribution as a stationary measure of occupational exposure.
Abbreviations: SOC = Standard Occupational Classification; AEI =
Anthropic Economic Index; CAI = Claude.ai consumer; API = first-party
Application Programming Interface enterprise; Composite = AI exposure
score constructed following \hyperlink{Massenkoff2026}{Massenkoff and McCrory (2026)}; BBD =
Bick-Blandin-Deming Survey of Working Arrangements and Attitudes; BLS =
Bureau of Labor Statistics; OES = Occupational Employment and Wage
Statistics; CKK = Chen, Kane, Kozlowski et al. (2025); ACS = American
Community Survey; CPS-MORG = Current Population Survey-Outgoing Rotation
Group.}

\endgroup
\clearpage
\begingroup
\scriptsize
\setlength{\tabcolsep}{2pt}
\renewcommand{\arraystretch}{0.78}
\renewcommand{\baselinestretch}{0.85}\selectfont
\textbf{Table B.5. Robustness Compendium of Significance Counts by
Dimension}

{\def\LTcaptype{none} 
\begin{longtable}[]{@{}
  >{\raggedright\arraybackslash}p{0.4500\linewidth}
  >{\centering\arraybackslash}p{0.2750\linewidth}
  >{\centering\arraybackslash}p{0.2750\linewidth}@{}}
\toprule\noalign{}
\begin{minipage}[b]{\linewidth}\raggedright
\textbf{Source group}
\end{minipage} & \begin{minipage}[b]{\linewidth}\centering
\textbf{Sig at p\textless0.05 (full / excluding 2020-2021)}
\end{minipage} & \begin{minipage}[b]{\linewidth}\centering
\textbf{Implication}
\end{minipage} \\
\midrule\noalign{}
\endhead
\bottomrule\noalign{}
\endlastfoot
Cross-platform raw (10 canonical variants) & 12 / 2 & Cross-platform
dispersion in main DiD coefficient \\
Within-vendor channel substitution (9 cells) & 2 / 1 & composite recipe
with vendor-channel substitution at Waves 3 to 5 \\
Heterogeneity (5 channels x 21 subgroups) & 59 / 46 & Subgroup
heterogeneity in main DiD coefficient \\
Sample restriction (3 subsets x 7 variants; full sample) & 10 / not run
& Robustness to sample restrictions (prime-age, non-Hispanic white,
female) \\
Fixed-effects and specification sensitivity (5 configs x 7 variants) &
29 / not run & Robustness to fixed-effects and control specifications \\
Wage outcome (log of annual wage income; 7 variants) & 4 / 3 &
Intensive-margin wage robustness under canonical specification \\
Pre-trend joint Wald F-test (7 variants) & 6 reject / not run & Only
Composite Reweighted passes parallel-trends test; all other variants
reject at p less than 0.05 \\
Wild-cluster bootstrap (9 variants; SOC-major-22 cluster; legacy) & 0 /
not run & Conservative under SOC-major cluster count; analytic
state-clustered standard errors are the canonical inference \\
\end{longtable}
}

\needspace{6\baselineskip}
\textbf{Notes:} \emph{This table compiles the counts of statistically
significant DiD coefficients at p \textless{} 0.05
across canonical variants within each robustness dimension. The table is
a summary index for the full robustness compendium documented in Tables
B.7 to B.9 and the manuscript appendix. Counts are reported separately
for the full sample (American Community Survey 2015 to 2024) and the
sample excluding 2020 and 2021 (a robustness check that drops the two
COVID-19 disruption years). Canonical variants comprise: Baseline
Composite (Anthropic), Composite Reweighted, Composite with Wave 5 AEI
weights with Wave 5 Anthropic Economic Index weights, first-party
Application Programming Interface enterprise Wave 5, Claude.ai consumer
Wave 5, Microsoft Copilot, and Pooled Claude.ai consumer plus
first-party Application Programming Interface enterprise Wave 5.
Robustness dimensions span outcome variation (employment vs labor-force
participation vs log wage), fixed-effects and specification sensitivity,
sample restrictions (prime-age, non-Hispanic white, female), alternative
wage outcomes, pre-trend joint Wald F-tests for parallel-trends
violation, and wild-cluster bootstrap inference under
small-cluster-count regimes. The Implication column gives a one-line
interpretation of each robustness dimension. Abbreviations: SOC =
Standard Occupational Classification; AEI = Anthropic Economic Index;
CAI = Claude.ai consumer; API = first-party Application Programming
Interface enterprise; Composite = AI exposure score constructed
following \hyperlink{Massenkoff2026}{Massenkoff and McCrory (2026)}; BBD = Bick-Blandin-Deming
Survey of Working Arrangements and Attitudes; BLS = Bureau of Labor
Statistics; OES = Occupational Employment and Wage Statistics; CKK =
Chen, Kane, Kozlowski et al. (2025); ACS = American Community Survey;
CPS-MORG = Current Population Survey-Outgoing Rotation Group.}

\endgroup
\clearpage

\begingroup
\scriptsize
\setlength{\tabcolsep}{2pt}
\renewcommand{\arraystretch}{0.7}
\renewcommand{\baselinestretch}{0.85}\selectfont
\textbf{Table B.6. Master Difference-in-Differences Coefficient Grid
Across Source, Recipe, and Channel}

{\def\LTcaptype{none} 
\begin{longtable}[]{@{}
  >{\raggedright\arraybackslash}p{0.4500\linewidth}
  >{\raggedright\arraybackslash}p{0.1833\linewidth}
  >{\raggedright\arraybackslash}p{0.1833\linewidth}
  >{\raggedright\arraybackslash}p{0.1833\linewidth}@{}}
\toprule\noalign{}
\begin{minipage}[b]{\linewidth}\raggedright
\textbf{Variant}
\end{minipage} & \begin{minipage}[b]{\linewidth}\centering
\textbf{Full sample}
\end{minipage} & \begin{minipage}[b]{\linewidth}\centering
\textbf{Excluding 2020-2021}
\end{minipage} & \begin{minipage}[b]{\linewidth}\centering
\textbf{N (full / excluding 2020-2021)}
\end{minipage} \\
\midrule\noalign{}
\endhead
\bottomrule\noalign{}
\endlastfoot
\multicolumn{4}{@{}>{\raggedright\arraybackslash}p{1.0000\linewidth + 6\tabcolsep}@{}}{%
\emph{\textbf{Panel A. Raw Anthropic Economic Index conversation share
(Claude.ai consumer / first-party Application Programming Interface
enterprise / pooled)}}} \\
Claude.ai consumer (Wave 1) & \begin{minipage}[t]{\linewidth}\centering
-0.116***\\
(0.035)\strut
\end{minipage} & \begin{minipage}[t]{\linewidth}\centering
-0.017\\
(0.035)\strut
\end{minipage} & 10.95M / 8.92M \\
Claude.ai consumer (Wave 3) & \begin{minipage}[t]{\linewidth}\centering
-0.163***\\
(0.033)\strut
\end{minipage} & \begin{minipage}[t]{\linewidth}\centering
-0.072*\\
(0.037)\strut
\end{minipage} & 11.00M / 8.95M \\
Claude.ai consumer (Wave 5) & \begin{minipage}[t]{\linewidth}\centering
-0.222***\\
(0.042)\strut
\end{minipage} & \begin{minipage}[t]{\linewidth}\centering
-0.162***\\
(0.047)\strut
\end{minipage} & 11.18M / 9.10M \\
1P API enterprise (Wave 3) & \begin{minipage}[t]{\linewidth}\centering
-0.144***\\
(0.035)\strut
\end{minipage} & \begin{minipage}[t]{\linewidth}\centering
-0.062*\\
(0.035)\strut
\end{minipage} & 10.08M / 8.22M \\
1P API enterprise (Wave 5) & \begin{minipage}[t]{\linewidth}\centering
-0.152***\\
(0.033)\strut
\end{minipage} & \begin{minipage}[t]{\linewidth}\centering
-0.061*\\
(0.036)\strut
\end{minipage} & 10.16M / 8.28M \\
\multicolumn{4}{@{}>{\raggedright\arraybackslash}p{1.0000\linewidth + 6\tabcolsep}@{}}{%
\emph{\textbf{Panel B. Baseline Composite (Anthropic) under cross-wave
Anthropic Economic Index weighting}}} \\
Baseline Composite (Anthropic) &
\begin{minipage}[t]{\linewidth}\centering
-0.139***\\
(0.035)\strut
\end{minipage} & \begin{minipage}[t]{\linewidth}\centering
-0.003\\
(0.038)\strut
\end{minipage} & 8.63M / 6.95M \\
Composite Reweighted & \begin{minipage}[t]{\linewidth}\centering
-0.010\\
(0.045)\strut
\end{minipage} & \begin{minipage}[t]{\linewidth}\centering
-0.020\\
(0.042)\strut
\end{minipage} & 8.63M / 6.95M \\
Composite with Wave 5 AEI weights &
\begin{minipage}[t]{\linewidth}\centering
-0.155***\\
(0.031)\strut
\end{minipage} & \begin{minipage}[t]{\linewidth}\centering
-0.001\\
(0.034)\strut
\end{minipage} & 8.63M / 6.95M \\
\multicolumn{4}{@{}>{\raggedright\arraybackslash}p{1.0000\linewidth + 6\tabcolsep}@{}}{%
\emph{\textbf{Panel C. Microsoft Copilot artificial-intelligence
applicability score (raw and reweighted)}}} \\
Microsoft Copilot (raw) & \begin{minipage}[t]{\linewidth}\centering
-0.191***\\
(0.042)\strut
\end{minipage} & \begin{minipage}[t]{\linewidth}\centering
-0.161***\\
(0.040)\strut
\end{minipage} & 12.28M / 9.99M \\
Microsoft Copilot (reweighted) &
\begin{minipage}[t]{\linewidth}\centering
-0.110***\\
(0.039)\strut
\end{minipage} & \begin{minipage}[t]{\linewidth}\centering
-0.036\\
(0.041)\strut
\end{minipage} & 12.28M / 9.99M \\
\multicolumn{4}{@{}>{\raggedright\arraybackslash}p{1.0000\linewidth + 6\tabcolsep}@{}}{%
\emph{\textbf{Panel D. Composite, channel-restricted (within-vendor
channel substitution at Anthropic Economic Index Waves 3 to 5)}}} \\
Claude.ai consumer (Wave 3) & \begin{minipage}[t]{\linewidth}\centering
-0.087*\\
(0.052)\strut
\end{minipage} & \begin{minipage}[t]{\linewidth}\centering
+0.006\\
(0.050)\strut
\end{minipage} & 4.73M / 3.80M \\
Claude.ai consumer (Wave 4) & \begin{minipage}[t]{\linewidth}\centering
-0.098**\\
(0.047)\strut
\end{minipage} & \begin{minipage}[t]{\linewidth}\centering
+0.048\\
(0.048)\strut
\end{minipage} & 5.36M / 4.31M \\
Claude.ai consumer (Wave 5) & \begin{minipage}[t]{\linewidth}\centering
-0.076*\\
(0.043)\strut
\end{minipage} & \begin{minipage}[t]{\linewidth}\centering
+0.069\\
(0.044)\strut
\end{minipage} & 5.77M / 4.64M \\
1P API enterprise (Wave 3) & \begin{minipage}[t]{\linewidth}\centering
+0.015\\
(0.058)\strut
\end{minipage} & \begin{minipage}[t]{\linewidth}\centering
-0.042\\
(0.052)\strut
\end{minipage} & 4.58M / 3.69M \\
1P API enterprise (Wave 4) & \begin{minipage}[t]{\linewidth}\centering
+0.041\\
(0.047)\strut
\end{minipage} & \begin{minipage}[t]{\linewidth}\centering
-0.047\\
(0.041)\strut
\end{minipage} & 4.60M / 3.70M \\
1P API enterprise (Wave 5) & \begin{minipage}[t]{\linewidth}\centering
+0.055\\
(0.044)\strut
\end{minipage} & \begin{minipage}[t]{\linewidth}\centering
+0.025\\
(0.043)\strut
\end{minipage} & 4.50M / 3.63M \\
Pooled CAI + API (Wave 3) & \begin{minipage}[t]{\linewidth}\centering
-0.064\\
(0.046)\strut
\end{minipage} & \begin{minipage}[t]{\linewidth}\centering
-0.006\\
(0.041)\strut
\end{minipage} & 5.46M / 4.39M \\
Pooled CAI + API (Wave 4) & \begin{minipage}[t]{\linewidth}\centering
-0.098**\\
(0.038)\strut
\end{minipage} & \begin{minipage}[t]{\linewidth}\centering
+0.020\\
(0.036)\strut
\end{minipage} & 6.17M / 4.98M \\
Pooled CAI + API (Wave 5) & \begin{minipage}[t]{\linewidth}\centering
-0.039\\
(0.038)\strut
\end{minipage} & \begin{minipage}[t]{\linewidth}\centering
+0.091**\\
(0.038)\strut
\end{minipage} & 6.17M / 4.97M \\
\end{longtable}
}

\needspace{6\baselineskip}
\textbf{Notes:} \emph{This table is the master grid of
DiD coefficients across all canonical variants
used in the analysis, organized into four panels by source family. Panel
A reports raw Anthropic Economic Index conversation share variants
(Claude.ai consumer at Waves 1, 3, 5; first-party Application
Programming Interface enterprise at Waves 3, 5). Panel B reports the
Baseline Composite (Anthropic) under three weighting schemes (baseline
as published, employment-reweighted by Bureau of Labor Statistics, and
Wave 5 Anthropic Economic Index weights substitution). Panel C reports
the Microsoft Copilot artificial-intelligence applicability score
(\hyperlink{Tomlinson2025}{\hyperlink{Tomlinson2025}{Tomlinson et al. 2025}}) in raw and reweighted forms. Panel D reports the
within-vendor channel-substitution variants under the composite
construction at Anthropic Economic Index Waves 3 to 5 (Claude.ai
consumer / first-party Application Programming Interface enterprise /
pooled). The regression specification is individual-level weighted least
squares on the canonical American Community Survey analytical sample
(wage-and-salary workers, ages 16 to 64, 2015 to 2024). The dependent
variable is a binary indicator equal to one if the respondent is
employed (American Community Survey empstat). The treatment variable is
the standardized exposure measure interacted with a post-treatment
indicator (post = 1 if year is 2023 or later); coefficients reported as
beta-times-100 represent the percentage-point change in employment per
one-standard-deviation increase in exposure. Two-way fixed effects:
year, state, and 6-digit Standard Occupational Classification 2018 (SOC)
occupation. Controls: sex (female indicator), marital status (married
indicator), three educational-attainment indicators (high-school
graduate, some college, Bachelor\textquotesingle s or higher), age, age
squared, and number of own children. Exposure is z-scored within the
analytical sample using employment-weighted means and standard
deviations. Regressions are person-weighted using the American Community
Survey person weight. Standard errors are cluster-robust at the state
level (51 states including the District of Columbia) and reported in
parentheses. Each cell reports the coefficient (beta times 100) with the
cluster-robust standard error in parentheses. N reports unweighted
person-year observation counts in millions for the full and
excluding-2020-2021 specifications. Significance: *** p \textless{}
0.01; ** p \textless{} 0.05; * p \textless{} 0.10. Abbreviations: SOC =
Standard Occupational Classification; AEI = Anthropic Economic Index;
CAI = Claude.ai consumer; API = first-party Application Programming
Interface enterprise; Composite = AI exposure score constructed
following \hyperlink{Massenkoff2026}{Massenkoff and McCrory (2026)}; BBD = Bick-Blandin-Deming
Survey of Working Arrangements and Attitudes; BLS = Bureau of Labor
Statistics; OES = Occupational Employment and Wage Statistics; CKK =
Chen, Kane, Kozlowski et al. (2025); ACS = American Community Survey;
CPS-MORG = Current Population Survey-Outgoing Rotation Group.}

\endgroup
\clearpage

\begin{landscape}
\begingroup
\scriptsize
\setlength{\tabcolsep}{2pt}
\renewcommand{\arraystretch}{0.78}
\renewcommand{\baselinestretch}{0.85}\selectfont
\textbf{Table B.7. Heterogeneity in Cross-Source
Difference-in-Differences Coefficients on Employment by Subgroup and
Channel}

\emph{\textbf{Panel A. Demographics}}

{\def\LTcaptype{none} 
\begin{longtable}[]{@{}
  >{\raggedright\arraybackslash}p{0.1692\linewidth}
  >{\centering\arraybackslash}p{0.0692\linewidth}
  >{\centering\arraybackslash}p{0.0692\linewidth}
  >{\centering\arraybackslash}p{0.0692\linewidth}
  >{\centering\arraybackslash}p{0.0692\linewidth}
  >{\centering\arraybackslash}p{0.0692\linewidth}
  >{\centering\arraybackslash}p{0.0692\linewidth}
  >{\centering\arraybackslash}p{0.0692\linewidth}
  >{\centering\arraybackslash}p{0.0692\linewidth}
  >{\centering\arraybackslash}p{0.0692\linewidth}
  >{\centering\arraybackslash}p{0.0692\linewidth}
  >{\centering\arraybackslash}p{0.0692\linewidth}
  >{\centering\arraybackslash}p{0.0692\linewidth}@{}}
\toprule\noalign{}
\begin{minipage}[b]{\linewidth}\raggedright
\textbf{Subgroup}
\end{minipage} & \begin{minipage}[b]{\linewidth}\centering
\textbf{Massenkoff and McCrory baseline (Full)}
\end{minipage} & \begin{minipage}[b]{\linewidth}\centering
\textbf{Massenkoff and McCrory baseline (Excluding 2020-2021)}
\end{minipage} & \begin{minipage}[b]{\linewidth}\centering
\textbf{Massenkoff and McCrory reweighted (Full)}
\end{minipage} & \begin{minipage}[b]{\linewidth}\centering
\textbf{Massenkoff and McCrory reweighted (Excluding 2020-2021)}
\end{minipage} & \begin{minipage}[b]{\linewidth}\centering
\textbf{Claude.ai consumer (Wave 5) (Full)}
\end{minipage} & \begin{minipage}[b]{\linewidth}\centering
\textbf{Claude.ai consumer (Wave 5) (Excluding 2020-2021)}
\end{minipage} & \begin{minipage}[b]{\linewidth}\centering
\textbf{1P API enterprise (Wave 5) (Full)}
\end{minipage} & \begin{minipage}[b]{\linewidth}\centering
\textbf{1P API enterprise (Wave 5) (Excluding 2020-2021)}
\end{minipage} & \begin{minipage}[b]{\linewidth}\centering
\textbf{Microsoft Copilot (2024) (Full)}
\end{minipage} & \begin{minipage}[b]{\linewidth}\centering
\textbf{Microsoft Copilot (2024) (Excluding 2020-2021)}
\end{minipage} & \begin{minipage}[b]{\linewidth}\centering
\textbf{Pooled Claude.ai + 1P API (Wave 5) (Full)}
\end{minipage} & \begin{minipage}[b]{\linewidth}\centering
\textbf{Pooled Claude.ai + 1P API (Wave 5) (Excluding 2020-2021)}
\end{minipage} \\
\midrule\noalign{}
\endhead
\bottomrule\noalign{}
\endlastfoot
Disability & & & & & & & & & & & & \\
Has disability & +0.148 (0.162) & +0.098 (0.158) & -0.101 (0.183) &
-0.236 (0.172) & -0.521*** (0.160) & -0.559*** (0.188) & -0.174 (0.144)
& -0.182 (0.155) & -0.119 (0.145) & -0.263* (0.133) & -0.344** (0.158) &
-0.383** (0.180) \\
No disability & -0.138*** (0.035) & +0.016 (0.038) & +0.019 (0.043) &
+0.023 (0.040) & -0.191*** (0.042) & -0.121*** (0.044) & -0.152***
(0.032) & -0.055 (0.033) & -0.157*** (0.044) & -0.107** (0.040) &
-0.172*** (0.035) & -0.093** (0.036) \\
Education & & & & & & & & & & & & \\
Bachelor\textquotesingle s degree or higher & -0.004 (0.065) & +0.108
(0.069) & +0.069 (0.066) & +0.013 (0.066) & -0.089* (0.050) & -0.044
(0.050) & -0.161*** (0.046) & -0.087* (0.045) & +0.056 (0.055) & +0.098*
(0.052) & -0.134*** (0.046) & -0.075 (0.045) \\
Less than bachelor\textquotesingle s degree & -0.155*** (0.053) & -0.056
(0.053) & -0.058 (0.062) & -0.039 (0.059) & -0.319*** (0.060) &
-0.277*** (0.062) & -0.128** (0.058) & -0.061 (0.051) & -0.292***
(0.066) & -0.334*** (0.058) & -0.234*** (0.060) & -0.190*** (0.056) \\
Sex & & & & & & & & & & & & \\
Female & +0.074 (0.051) & +0.228*** (0.053) & +0.008 (0.049) & +0.015
(0.054) & -0.129** (0.059) & -0.082 (0.062) & -0.028 (0.050) & +0.063
(0.051) & +0.054 (0.051) & +0.103* (0.052) & -0.072 (0.055) & -0.015
(0.057) \\
Male & -0.388*** (0.046) & -0.263*** (0.057) & -0.035 (0.062) & -0.072
(0.068) & -0.307*** (0.049) & -0.237*** (0.051) & -0.235*** (0.045) &
-0.140*** (0.037) & -0.454*** (0.052) & -0.443*** (0.060) & -0.268***
(0.044) & -0.187*** (0.041) \\
Race and ethnicity & & & & & & & & & & & & \\
Non-Hispanic white & -0.053 (0.050) & +0.039 (0.055) & +0.026 (0.051) &
+0.004 (0.052) & -0.177*** (0.042) & -0.147*** (0.044) & -0.162***
(0.048) & -0.107** (0.048) & -0.066* (0.037) & -0.069* (0.039) &
-0.175*** (0.046) & -0.138*** (0.047) \\
Non-Hispanic Black & -0.062 (0.175) & +0.065 (0.161) & -0.082 (0.149) &
-0.049 (0.143) & -0.535*** (0.136) & -0.485*** (0.141) & -0.244* (0.132)
& -0.135 (0.119) & -0.396** (0.152) & -0.353** (0.141) & -0.395***
(0.134) & -0.317** (0.128) \\
Hispanic & -0.074 (0.089) & +0.089 (0.093) & -0.080 (0.111) & -0.075
(0.113) & -0.233*** (0.071) & -0.115 (0.071) & -0.167*** (0.044) &
+0.006 (0.057) & -0.094 (0.074) & -0.084 (0.066) & -0.208*** (0.048) &
-0.065 (0.058) \\
\end{longtable}
}

\emph{\textbf{Panel B. Wage, Occupation, Age, and Marital Status}}

{\def\LTcaptype{none} 
\begin{longtable}[]{@{}
  >{\raggedright\arraybackslash}p{0.1692\linewidth}
  >{\centering\arraybackslash}p{0.0692\linewidth}
  >{\centering\arraybackslash}p{0.0692\linewidth}
  >{\centering\arraybackslash}p{0.0692\linewidth}
  >{\centering\arraybackslash}p{0.0692\linewidth}
  >{\centering\arraybackslash}p{0.0692\linewidth}
  >{\centering\arraybackslash}p{0.0692\linewidth}
  >{\centering\arraybackslash}p{0.0692\linewidth}
  >{\centering\arraybackslash}p{0.0692\linewidth}
  >{\centering\arraybackslash}p{0.0692\linewidth}
  >{\centering\arraybackslash}p{0.0692\linewidth}
  >{\centering\arraybackslash}p{0.0692\linewidth}
  >{\centering\arraybackslash}p{0.0692\linewidth}@{}}
\toprule\noalign{}
\begin{minipage}[b]{\linewidth}\raggedright
\textbf{Subgroup}
\end{minipage} & \begin{minipage}[b]{\linewidth}\centering
\textbf{Massenkoff and McCrory baseline (Full)}
\end{minipage} & \begin{minipage}[b]{\linewidth}\centering
\textbf{Massenkoff and McCrory baseline (Excluding 2020-2021)}
\end{minipage} & \begin{minipage}[b]{\linewidth}\centering
\textbf{Massenkoff and McCrory reweighted (Full)}
\end{minipage} & \begin{minipage}[b]{\linewidth}\centering
\textbf{Massenkoff and McCrory reweighted (Excluding 2020-2021)}
\end{minipage} & \begin{minipage}[b]{\linewidth}\centering
\textbf{Claude.ai consumer (Wave 5) (Full)}
\end{minipage} & \begin{minipage}[b]{\linewidth}\centering
\textbf{Claude.ai consumer (Wave 5) (Excluding 2020-2021)}
\end{minipage} & \begin{minipage}[b]{\linewidth}\centering
\textbf{1P API enterprise (Wave 5) (Full)}
\end{minipage} & \begin{minipage}[b]{\linewidth}\centering
\textbf{1P API enterprise (Wave 5) (Excluding 2020-2021)}
\end{minipage} & \begin{minipage}[b]{\linewidth}\centering
\textbf{Microsoft Copilot (2024) (Full)}
\end{minipage} & \begin{minipage}[b]{\linewidth}\centering
\textbf{Microsoft Copilot (2024) (Excluding 2020-2021)}
\end{minipage} & \begin{minipage}[b]{\linewidth}\centering
\textbf{Pooled Claude.ai + 1P API (Wave 5) (Full)}
\end{minipage} & \begin{minipage}[b]{\linewidth}\centering
\textbf{Pooled Claude.ai + 1P API (Wave 5) (Excluding 2020-2021)}
\end{minipage} \\
\midrule\noalign{}
\endhead
\bottomrule\noalign{}
\endlastfoot
Wage quartile & & & & & & & & & & & & \\
Wage quartile 1 (lowest) & -0.166 (0.103) & -0.180* (0.099) & -0.268**
(0.112) & -0.285** (0.109) & -0.386*** (0.089) & -0.323*** (0.092) &
-0.217** (0.085) & -0.269*** (0.090) & -0.591*** (0.111) & -0.664***
(0.110) & -0.379*** (0.089) & -0.386*** (0.093) \\
Wage quartile 2 & -0.110* (0.056) & -0.044 (0.057) & -0.063 (0.047) &
-0.035 (0.047) & -0.077** (0.037) & -0.075* (0.042) & -0.043 (0.045) &
-0.017 (0.050) & -0.100* (0.056) & -0.137** (0.055) & -0.064 (0.042) &
-0.057 (0.047) \\
Wage quartile 3 & -0.127*** (0.032) & -0.059 (0.036) & -0.040 (0.045) &
-0.046 (0.048) & -0.048 (0.030) & -0.029 (0.030) & -0.023 (0.036) &
-0.005 (0.037) & -0.072** (0.031) & -0.052 (0.035) & -0.031 (0.035) &
-0.014 (0.035) \\
Wage quartile 4 (highest) & -0.166*** (0.025) & -0.109*** (0.025) &
-0.054 (0.042) & -0.052 (0.039) & -0.119*** (0.023) & -0.088*** (0.019)
& -0.077*** (0.022) & -0.051*** (0.018) & -0.152*** (0.021) & -0.101***
(0.020) & -0.092*** (0.022) & -0.065*** (0.019) \\
Occupation major group & & & & & & & & & & & & \\
Management and Professional (SOC 11-29) & +0.005 (0.062) & +0.132**
(0.064) & +0.166** (0.071) & +0.044 (0.066) & -0.102* (0.053) & -0.076
(0.052) & -0.186*** (0.050) & -0.127*** (0.047) & +0.102* (0.051) &
+0.196*** (0.052) & -0.162*** (0.051) & -0.115** (0.049) \\
Service (SOC 31-39) & -0.172 (0.131) & -0.259** (0.125) & -0.186 (0.114)
& -0.040 (0.112) & -0.000 (0.145) & -0.091 (0.163) & -0.502*** (0.116) &
-0.226* (0.115) & +0.071 (0.113) & -0.315*** (0.112) & -0.226 (0.139) &
-0.163 (0.151) \\
Sales and Office (SOC 41-43) & +0.119 (0.072) & +0.006 (0.073) & -0.071
(0.071) & -0.097 (0.069) & -0.336*** (0.086) & -0.424*** (0.090) &
+0.156** (0.068) & +0.182** (0.073) & +0.036 (0.064) & -0.046 (0.063) &
-0.152* (0.076) & -0.200** (0.082) \\
Blue-collar (SOC 45-53) & +0.051 (0.085) & -0.004 (0.086) & +0.189*
(0.106) & +0.142 (0.103) & -0.162 (0.114) & -0.065 (0.124) & -0.290**
(0.135) & -0.124 (0.145) & -0.161** (0.068) & -0.234*** (0.070) & -0.211
(0.133) & -0.060 (0.145) \\
Age strata & & & & & & & & & & & & \\
Younger than 35 & -0.148** (0.072) & -0.050 (0.069) & +0.036 (0.077) &
+0.001 (0.080) & -0.294*** (0.081) & -0.255*** (0.088) & -0.194***
(0.046) & -0.095* (0.048) & -0.358*** (0.075) & -0.376*** (0.076) &
-0.253*** (0.062) & -0.196*** (0.068) \\
Age 35 to 50 & -0.175*** (0.056) & +0.025 (0.061) & -0.115* (0.068) &
-0.103 (0.078) & -0.161*** (0.055) & -0.070 (0.056) & -0.181*** (0.047)
& -0.092* (0.048) & -0.083 (0.060) & +0.033 (0.060) & -0.175*** (0.051)
& -0.089* (0.052) \\
Age 50 or older & +0.000 (0.060) & +0.109** (0.051) & +0.083 (0.067) &
+0.107 (0.072) & -0.153*** (0.056) & -0.095* (0.054) & -0.041 (0.065) &
+0.032 (0.059) & +0.002 (0.054) & +0.018 (0.053) & -0.083 (0.061) &
-0.018 (0.056) \\
Marital status & & & & & & & & & & & & \\
Married & +0.013 (0.044) & +0.146*** (0.044) & +0.031 (0.062) & +0.047
(0.067) & -0.083** (0.038) & -0.014 (0.044) & -0.068* (0.040) & +0.009
(0.041) & +0.017 (0.037) & +0.102** (0.041) & -0.071* (0.041) & -0.000
(0.042) \\
\end{longtable}
}

\needspace{6\baselineskip}
\textbf{Notes:} \emph{Each cell reports the coefficient on the
interaction of standardized exposure with the post-treatment indicator
(post = 1 if year is 2023 or later) from individual-level weighted least
squares subset regressions on the canonical American Community Survey
analytical sample. Six channels: (1) Baseline Composite (Anthropic); (2)
Massenkoff and McCrory reweighted (Bureau of Labor Statistics population
weights); (3) Claude.ai consumer Wave 5; (4) first-party Application
Programming Interface enterprise Wave 5; (5) Microsoft Copilot 2024; (6)
Pooled Claude.ai consumer plus first-party Application Programming
Interface enterprise Wave 5. Standard errors clustered on state in
parentheses. Significance: *** p\textless0.01, ** p\textless0.05, *
p\textless0.10.}

\endgroup
\end{landscape}
\clearpage
\begingroup
\scriptsize
\setlength{\tabcolsep}{2pt}
\renewcommand{\arraystretch}{0.78}
\renewcommand{\baselinestretch}{0.85}\selectfont
\textbf{Table B.8. Robustness Dispersion in Outcome and Specification
Sensitivity}

{\def\LTcaptype{none} 
\begin{longtable}[]{@{}
  >{\raggedright\arraybackslash}p{0.4400\linewidth}
  >{\raggedright\arraybackslash}p{0.1400\linewidth}
  >{\raggedright\arraybackslash}p{0.1400\linewidth}
  >{\raggedright\arraybackslash}p{0.1400\linewidth}
  >{\raggedright\arraybackslash}p{0.1400\linewidth}@{}}
\toprule\noalign{}
\begin{minipage}[b]{\linewidth}\raggedright
\textbf{Perturbation}
\end{minipage} & \begin{minipage}[b]{\linewidth}\centering
\textbf{beta x 100 range}
\end{minipage} & \begin{minipage}[b]{\linewidth}\centering
\textbf{max/min ratio}
\end{minipage} & \begin{minipage}[b]{\linewidth}\centering
\textbf{Sign split (neg / pos)}
\end{minipage} & \begin{minipage}[b]{\linewidth}\centering
\textbf{Interpretation}
\end{minipage} \\
\midrule\noalign{}
\endhead
\bottomrule\noalign{}
\endlastfoot
\multicolumn{5}{@{}>{\raggedright\arraybackslash}p{1.0000\linewidth + 8\tabcolsep}@{}}{%
\emph{\textbf{Panel A. Outcome invariance across three labor-market
outcomes (employment, labor-force participation, log wage)}}} \\
Employment, full sample & 0.010 to 0.222 & 22.4x & 10 / 0 & All
negative; magnitudes diverge \\
Employment, excluding 2020-2021 & 0.001 to 0.162 & \textgreater100x & 10
/ 0 & All negative; magnitudes diverge \\
Labor-force participation, full sample & 0.005 to 0.126 & 25.8x & 5 / 5
& Sign mixing across variants \\
Labor-force participation, excluding 2020-2021 & 0.006 to 0.153 & 27.2x
& 2 / 8 & Cross-source dispersion preserved \\
Log wage, full sample & 0.009 to 0.752 & 84.4x & 5 / 2 & Cross-source
dispersion preserved \\
Log wage, excluding 2020-2021 & 0.042 to 0.616 & 14.6x & 5 / 2 &
Cross-source dispersion preserved \\
\multicolumn{5}{@{}>{\raggedright\arraybackslash}p{1.0000\linewidth + 8\tabcolsep}@{}}{%
\emph{\textbf{Panel B. Pre-treatment parallel-trends joint Wald
F-test}}} \\
Joint Wald F-test on 7 pre-period coefficients (full sample) & F = 1.54
to 23.72 & n/a & 6 reject / 1 pass at p \textless{} 0.05 & Only
Composite Reweighted variant passes parallel-trends test \\
Joint Wald F-test on 5 pre-period coefficients (excluding 2020-2021) & F
= 1.35 to 11.27 & n/a & 6 reject / 1 pass at p \textless{} 0.05 & Same
conclusion as full sample; pre-trend test robust to COVID-period
exclusion \\
\multicolumn{5}{@{}>{\raggedright\arraybackslash}p{1.0000\linewidth + 8\tabcolsep}@{}}{%
\emph{\textbf{Panel C. Sample-restriction robustness}}} \\
Full sample (16-64) & 0.010 to 0.222 & 22.4x & 10 / 0 & Baseline \\
Prime-age (25-54) & 0.017 to 0.182 & 10.6x & 7 / 0 & All negative;
magnitudes diverge \\
Non-Hispanic white only & 0.026 to 0.177 & 6.7x & 6 / 1 & Cross-source
dispersion preserved \\
Female only & 0.008 to 0.129 & 16.5x & 2 / 5 & Cross-source dispersion
preserved \\
\multicolumn{5}{@{}>{\raggedright\arraybackslash}p{1.0000\linewidth + 8\tabcolsep}@{}}{%
\emph{\textbf{Panel D. Specification-sensitivity robustness (fixed
effects and controls)}}} \\
Baseline (year + state + occupation fixed effects; full controls) &
0.010 to 0.222 & 22.4x & 7 / 0 & All negative; magnitudes diverge \\
Drop demographic controls & 0.024 to 0.221 & 9.1x & 6 / 1 & Robust to
dropping controls \\
State-by-year saturated fixed effects & 0.009 to 0.222 & 25.4x & 7 / 0 &
Robust to state-by-year fixed effects \\
Drop year fixed effects & 0.028 to 0.202 & 7.2x & 7 / 0 & Robust; all
negative \\
Drop occupation fixed effects & 0.069 to 0.613 & 8.8x & 3 / 4 & Sign
flips without occupation fixed effects \\
\end{longtable}
}

\needspace{6\baselineskip}
\textbf{Notes:} \emph{This table summarizes the dispersion of the
cross-source DiD coefficient across ten canonical
variants under four classes of robustness perturbation, organized into
four panels. Panel A reports outcome invariance (employment, labor-force
participation, log wage, each in full and excluding-2020-2021 samples).
Panel B reports the pre-treatment parallel-trends joint Wald F-test (the
F-statistic and pass/reject count at p \textless{} 0.05 for each
variant). Panel C reports sample-restriction robustness (prime-age 25 to
54, non-Hispanic white only, female only). Panel D reports
specification-sensitivity robustness (alternative fixed-effects
configurations and dropping demographic controls). Each row reports: the
range of beta-times-100 coefficients across variants; the
maximum-to-minimum absolute-magnitude ratio (a dispersion index where 1
means perfect cross-source agreement and \textgreater10 means an
order-of-magnitude or larger spread); the sign split (negative count /
positive count among variants); and a one-line interpretation. The
\textquotesingle Full\textquotesingle{} label indicates the full 2015 to
2024 sample; \textquotesingle excluding 2020-2021\textquotesingle{}
drops 2020 and 2021. The robustness dispersion grid shows that
cross-source dispersion is preserved under nearly all perturbations,
motivating the manuscript\textquotesingle s central conclusion that
platform-visibility distortion is the dominant source of variation in
estimated treatment effects rather than specification choices.
Abbreviations: SOC = Standard Occupational Classification; AEI =
Anthropic Economic Index; CAI = Claude.ai consumer; API = first-party
Application Programming Interface enterprise; Composite = AI exposure
score constructed following \hyperlink{Massenkoff2026}{Massenkoff and McCrory (2026)}; BBD =
Bick-Blandin-Deming Survey of Working Arrangements and Attitudes; BLS =
Bureau of Labor Statistics; OES = Occupational Employment and Wage
Statistics; CKK = Chen, Kane, Kozlowski et al. (2025); ACS = American
Community Survey; CPS-MORG = Current Population Survey-Outgoing Rotation
Group.}

\endgroup
\clearpage
\begingroup
\scriptsize
\setlength{\tabcolsep}{2pt}
\renewcommand{\arraystretch}{0.78}
\renewcommand{\baselinestretch}{0.85}\selectfont
\textbf{Table B.9. Massenkoff-McCrory Cross-Occupation Regression
Coefficients by Platform Family}

{\def\LTcaptype{none} 
\begin{longtable}[]{@{}
  >{\raggedright\arraybackslash}p{0.4500\linewidth}
  >{\centering\arraybackslash}p{0.2750\linewidth}
  >{\centering\arraybackslash}p{0.2750\linewidth}@{}}
\toprule\noalign{}
\begin{minipage}[b]{\linewidth}\raggedright
\textbf{Exposure measure}
\end{minipage} & \begin{minipage}[b]{\linewidth}\centering
\textbf{N}
\end{minipage} & \begin{minipage}[b]{\linewidth}\centering
\textbf{Coef per SD (x100)}
\end{minipage} \\
\midrule\noalign{}
\endhead
\bottomrule\noalign{}
\endlastfoot
Panel A. M-M baseline composite: published anchor and BLS-reweighted &
& \\
M-M baseline composite (published anchor) & 756 & -70.09*** \\
& & (17.90) \\
M-M baseline composite, BLS-reweighted (MM\_RW) & 766 & -38.28** \\
& & (17.35) \\
Panel B. Anthropic Claude.ai consumer (raw conversation share) & & \\
Anthropic Claude.ai consumer Wave 1 (Dec 2024) & 582 & +43.30 \\
& & (33.64) \\
Anthropic Claude.ai consumer Wave 2 (Feb 2025) & 579 & -8.16 \\
& & (34.23) \\
Anthropic Claude.ai consumer Wave 3 (Aug 2025) & 538 & -12.10 \\
& & (35.08) \\
Anthropic Claude.ai consumer Wave 4 (Nov 2025) & 559 & -28.12 \\
& & (33.46) \\
Anthropic Claude.ai consumer Wave 5 (Feb 2026) & 561 & -68.01** \\
& & (28.80) \\
Panel C. Anthropic 1P API enterprise (raw conversation share) & & \\
Anthropic 1P API enterprise Wave 3 (Aug 2025) & 480 & +27.61 \\
& & (37.16) \\
Anthropic 1P API enterprise Wave 4 (Nov 2025) & 485 & +23.88 \\
& & (36.38) \\
Anthropic 1P API enterprise Wave 5 (Feb 2026) & 480 & +34.00 \\
& & (36.03) \\
Panel D. Microsoft Bing Copilot 2024 (raw conversation share) & & \\
Microsoft Bing Copilot 2024 (\hyperlink{Tomlinson2025}{\hyperlink{Tomlinson2025}{Tomlinson et al. 2025}}) & 785 &
-150.04*** \\
& & (24.00) \\
Panel E. External cross-platform sources (SOC-major aggregated, expanded
to NEM) & & \\
OpenAI ChatGPT 2024 (\hyperlink{Chatterji2025}{\hyperlink{Chatterji2025}{Chatterji et al. 2025}}, SOC-major-expanded) & 1112 &
+174.00*** \\
& & (13.61) \\
\hyperlink{Bick2026}{Bick, Blandin, and Deming (2026)}, RPS SOC-major-expanded & 1091 &
+102.26*** \\
& & (13.28) \\
& & \\
\hyperlink{Bick2026}{Bick, Blandin, and Deming (2026)}, RPS six-digit SOC micro & 270 &
+166.81*** \\
& & (45.80) \\
& & \\
Manski partial-identification bounds on \textbar Coef per SD\textbar{}
(x100) & & \\
Lower bound: M-M composite, BLS-reweighted & & 38.28 \\
Upper bound: OpenAI ChatGPT 2024 (\hyperlink{Chatterji2025}{\hyperlink{Chatterji2025}{Chatterji et al. 2025}}, SOC-major-expanded) &
& 174.00 \\
Identified-set width & & 135.72 \\
\end{longtable}
}

\needspace{6\baselineskip}
\textbf{Notes:} \emph{Each row reports the WLS coefficient from a
regression of 2024-2034 BLS occupational employment-projection growth on
the platform\textquotesingle s AI exposure measure, weighted by 2024
occupational employment. Coefficients are scaled per standard deviation
of the exposure measure times 100, with employment-weighted standard
errors in parentheses. The new RPS
micro row at the bottom of Panel E reports the coefficient on the
six-digit SOC workforce AI use rate from the pooled August and November
2024 micro release. The RPS is a national online survey conducted on
Qualtrics with awareness-adjusted demographic weights and is not a
probability sample of the U.S. workforce. The sample for the RPS micro
row is restricted to the 270 SOC cells with at least five RPS
respondents. *** p\textless0.01, ** p\textless0.05, * p\textless0.10.}

\endgroup
\clearpage
\begingroup
\scriptsize
\setlength{\tabcolsep}{2pt}
\renewcommand{\arraystretch}{0.78}
\renewcommand{\baselinestretch}{0.85}\selectfont
\textbf{Table B.10. Cross-Platform Substitution across Five Published
Target Methods}

{\def\LTcaptype{none} 
\begin{longtable}[]{@{}
  >{\raggedright\arraybackslash}p{0.4500\linewidth}
  >{\centering\arraybackslash}p{0.1833\linewidth}
  >{\centering\arraybackslash}p{0.1833\linewidth}
  >{\centering\arraybackslash}p{0.1833\linewidth}@{}}
\toprule\noalign{}
\begin{minipage}[b]{\linewidth}\raggedright
\textbf{Paper / target / substitute}
\end{minipage} & \begin{minipage}[b]{\linewidth}\centering
\textbf{Value}
\end{minipage} & \begin{minipage}[b]{\linewidth}\centering
\textbf{Delta from published}
\end{minipage} & \begin{minipage}[b]{\linewidth}\centering
\textbf{N}
\end{minipage} \\
\midrule\noalign{}
\endhead
\bottomrule\noalign{}
\endlastfoot
Panel A. Yale Budget Lab (2025): top-quintile share &
Top-quintile occupational employment share (\%) & & 535 \\
Original published & 16.40\% & & \\
Our replication (Yale PCA composite) & 23.41\% & +7.01pp & \\
Cross-platform substitutes (raw exposure): & & & \\
Original composite score & 30.46\% & +14.06pp & \\
Claude.ai consumer Wave 1 & 30.34\% & +13.94pp & \\
Claude.ai consumer Wave 5 & 36.42\% & +20.02pp & \\
1P API enterprise Wave 5 & 37.65\% & +21.25pp & \\
Microsoft Bing Copilot 2024 & 22.65\% & +6.25pp & \\
& & & \\
Panel B. \hyperlink{Chen2025}{Chen-Kane-Kozlowski et al. (2025)}: SDiD log earnings &
Synthetic DiD above-median ATT in Jan 2010 dollars/week & & 1,391,824 \\
Original published & +\$89.00/wk & & 353 \\
Cross-platform substitutes (raw exposure): & & & \\
Original composite score & +\$57.79 (w) / +\$49.05 (s) & -\$31.21 (w)
& \\
Composite with Wave 5 AEI weights & +\$58.63 (w) / +\$48.00 (s) &
-\$30.37 (w) & \\
Claude.ai consumer Wave 1 & +\$50.94 (w) / +\$67.68 (s) & -\$38.06 (w)
& \\
Claude.ai consumer Wave 5 & +\$43.76 (w) / +\$51.17 (s) & -\$45.24 (w)
& \\
1P API enterprise Wave 5 & +\$44.69 (w) / +\$62.72 (s) & -\$44.31 (w)
& \\
Microsoft Bing Copilot 2024 & +\$16.67 (w) / +\$10.44 (s) & -\$72.33 (w)
& \\
& & & \\
Panel C. \hyperlink{Brynjolfsson2025}{Brynjolfsson-Chandar-Chen (2025)}: Q5 employment growth &
Percentage-point change in employment, ages 22-25 & & 196 \\
Original published & -10.00\% {[}-8.00, -12.00{]} & & \\
Cross-platform substitutes (raw exposure): & & & \\
1P API enterprise Wave 5 & -7.62\% {[}-6.78, -8.46{]} & +2.38pp & \\
& & & \\
Panel D. \hyperlink{Massenkoff2026}{Massenkoff and McCrory (2026)}, CPS unemployment
DiD & Set overlap (Jaccard) with original
top-quartile treatment & & 245 \\
Original published (own-method) & 1.00 & & \\
Cross-platform substitutes (raw exposure): & & & \\
Claude.ai consumer Wave 1 & 0.19 & -0.81 & \\
Claude.ai consumer Wave 5 & 0.19 & -0.81 & \\
1P API enterprise Wave 5 & 0.21 & -0.79 & \\
Microsoft Bing Copilot 2024 & 0.28 & -0.72 & \\
& & & \\
Panel E. \hyperlink{Tomlinson2025}{Tomlinson, Jaffe, Wang, Counts, Suri (2025)}: AI applicability &
Set overlap (Jaccard) with original top-quintile occupations & & 192 \\
Original published (own-method) & 1.00 & & \\
Cross-platform substitutes (raw exposure): & & & \\
Claude.ai consumer Wave 5 & 0.20 & -0.80 & \\
1P API enterprise Wave 5 & 0.16 & -0.84 & \\
\end{longtable}
}

\needspace{6\baselineskip}
\textbf{Notes:} \emph{Each panel reports a separate published target method and
the result of substituting platform-derived AI-exposure inputs into that
method, holding the rest of the published specification fixed. Companion
to main-text Table 5, which reports the analogous substitution exercise
for the BLS Employment Projections specification. Per the analytical
correction documented in Section I, reweighted versions of the
alternative-rubric (Eloundou tau-alpha) exposures show unstable
sign-flip behavior under occupation-level reweighting because the bias
enters at the task level rather than the occupation level. Only raw
(baseline) cross-platform substitutes are reported here. Main-text Table
5 reports the $\tau_b$ reweighting (reweighted composite score) which is
methodologically appropriate for the BLS Projections specification
alone. Yale Budget Lab (Panel A): published exposure is a principal
components composite across multiple AI inputs; we replicate at 6-digit
Standard Occupational Classification level (n=535) and substitute
platform-derived exposure inputs. CKK (Panel B): synthetic
DiD on monthly Current Population Survey Outgoing
Rotation Group data; the +\$89/wk anchor is not reproduced within the
+/-5 percent tolerance because of methodology differences
(per-treated-unit balanced donor pool retains four of thirty-three
post-period months; exposure-metric difference between Anthropic
Economic Index conversation share and Handa task-coverage aggregation).
BCC (Panel C): only one substitute reported because the rest of the
platform inputs were not aggregated to the BCC age-restricted occupation
sample. Massenkoff and McCrory CPS (Panel D) and Tomlinson Jaccard
(Panel E): set-overlap (Jaccard) statistics rather than coefficient
substitution; report the fraction of top-quartile (Massenkoff and
McCrory CPS) or top-quintile (Tomlinson) occupations preserved across
the substitution. Significance stars and standard errors are not
reported because the underlying methods use heterogeneous estimators
(PCA composite for Yale; synthdid jackknife for CKK;
DiD with bootstrapped intervals for BCC;
set-overlap for Massenkoff and McCrory CPS and Tomlinson).}

\endgroup
\clearpage

\begin{landscape}
\begingroup
\scriptsize
\setlength{\tabcolsep}{2pt}
\renewcommand{\arraystretch}{0.78}
\renewcommand{\baselinestretch}{0.85}\selectfont
\textbf{Table B.11. Diagnostic Evidence on the Composition ($\psi$) versus
Behavior ($\theta$) Channels in Cross-Wave Instability}

\emph{\textbf{Panel A. Within-Occupation Task-Share Stability (Spearman
$\rho$)}}

{\def\LTcaptype{none} 
\begin{longtable}[]{@{}
  >{\raggedright\arraybackslash}p{0.4400\linewidth}
  >{\centering\arraybackslash}p{0.1400\linewidth}
  >{\centering\arraybackslash}p{0.1400\linewidth}
  >{\centering\arraybackslash}p{0.1400\linewidth}
  >{\centering\arraybackslash}p{0.1400\linewidth}@{}}
\toprule\noalign{}
\begin{minipage}[b]{\linewidth}\raggedright
\textbf{Wave pair}
\end{minipage} & \begin{minipage}[b]{\linewidth}\centering
\textbf{N occupations}
\end{minipage} & \begin{minipage}[b]{\linewidth}\centering
\textbf{Mean $\rho$}
\end{minipage} & \begin{minipage}[b]{\linewidth}\centering
\textbf{Median $\rho$}
\end{minipage} & \begin{minipage}[b]{\linewidth}\centering
\textbf{Share with $\rho$ \textgreater{} 0.7}
\end{minipage} \\
\midrule\noalign{}
\endhead
\bottomrule\noalign{}
\endlastfoot
Waves 3 to 4 (Projects + Sonnet 4 to 4.5) & 404 & 0.683 & 0.800 &
65.3\% \\
Waves 4 to 5 (Opus 4.6 minor refresh) & 411 & 0.863 & 0.943 & 87.8\% \\
Waves 3 to 5 (across product cycle) & 403 & 0.700 & 0.808 & 67.0\% \\
Computer and Mathematical (SOC 15) subset: & & & & \\
Waves 3 to 4 & 18 & 0.866 & 0.898 & 94.4\% \\
Waves 4 to 5 & 18 & 0.912 & 0.967 & 94.4\% \\
Waves 3 to 5 & 18 & 0.901 & 0.896 & 100.0\% \\
\end{longtable}
}

\emph{\textbf{Panel B. Conversation-Share Compositional Shifts at Wave
Transitions}}

{\def\LTcaptype{none} 
\begin{longtable}[]{@{}
  >{\raggedright\arraybackslash}p{0.4400\linewidth}
  >{\raggedright\arraybackslash}p{0.1400\linewidth}
  >{\centering\arraybackslash}p{0.1400\linewidth}
  >{\centering\arraybackslash}p{0.1400\linewidth}
  >{\centering\arraybackslash}p{0.1400\linewidth}@{}}
\toprule\noalign{}
\begin{minipage}[b]{\linewidth}\raggedright
\textbf{Wave pair}
\end{minipage} & \begin{minipage}[b]{\linewidth}\raggedright
\textbf{Product event between waves}
\end{minipage} & \begin{minipage}[b]{\linewidth}\centering
\textbf{Six-digit SOC L1}
\end{minipage} & \begin{minipage}[b]{\linewidth}\centering
\textbf{SOC-major L1}
\end{minipage} & \begin{minipage}[b]{\linewidth}\centering
\textbf{SOC 15 share change}
\end{minipage} \\
\midrule\noalign{}
\endhead
\bottomrule\noalign{}
\endlastfoot
Waves 1 to 2 & Sonnet 3.5 to 3.7 minor refresh & 33.73 pp & 10.58 pp &
+0.26 pp \\
Waves 2 to 3 & Claude Code launch & 45.92 pp & 11.29 pp & +4.27 pp \\
Waves 3 to 4 & Projects expansion + Sonnet 4 to 4.5 & 27.91 pp & 12.15
pp & -2.33 pp \\
Waves 4 to 5 & Opus 4.6 minor refresh & 20.51 pp & 14.45 pp & -4.48
pp \\
\end{longtable}
}

\emph{\textbf{Panel C. Sub-Occupation Heterogeneity in Computer and
Mathematical (SOC 15) Major-Group Share Movement}}

{\def\LTcaptype{none} 
\begin{longtable}[]{@{}
  >{\raggedright\arraybackslash}p{0.4400\linewidth}
  >{\centering\arraybackslash}p{0.1400\linewidth}
  >{\centering\arraybackslash}p{0.1400\linewidth}
  >{\centering\arraybackslash}p{0.1400\linewidth}
  >{\centering\arraybackslash}p{0.1400\linewidth}@{}}
\toprule\noalign{}
\begin{minipage}[b]{\linewidth}\raggedright
\textbf{Wave pair}
\end{minipage} & \begin{minipage}[b]{\linewidth}\centering
\textbf{N sub-occupations}
\end{minipage} & \begin{minipage}[b]{\linewidth}\centering
\textbf{Mean growth ratio}
\end{minipage} & \begin{minipage}[b]{\linewidth}\centering
\textbf{Cross-sub-occ CV}
\end{minipage} & \begin{minipage}[b]{\linewidth}\centering
\textbf{SOC 15 share change}
\end{minipage} \\
\midrule\noalign{}
\endhead
\bottomrule\noalign{}
\endlastfoot
Waves 1 to 2 & 21 & 1.019 & 0.396 & +0.26 pp \\
Waves 2 to 3 & 21 & 1.415 & 0.848 & +4.27 pp \\
Waves 3 to 4 & 21 & 0.985 & 0.315 & -2.33 pp \\
Waves 4 to 5 & 21 & 1.046 & 0.200 & -4.48 pp \\
Waves 1 to 3 (across Claude Code launch) & 21 & 1.274 & 0.988 & +4.53
pp \\
\end{longtable}
}

\needspace{6\baselineskip}
\textbf{Notes:} \emph{Panel A reports the Spearman rank correlation
between within-occupation task-share distributions in two Anthropic
Economic Index waves, computed for each six-digit Standard Occupational
Classification (SOC) occupation with at least three tasks.
Within-occupation task share is the share of each \textit{O*NET} task in total
Claude.ai consumer conversations attributed to that occupation (global
geographic aggregation). The Computer and Mathematical subset is
reported separately. Waves 1 and 2 use a different raw task-level
aggregation and are not directly comparable to the W3 to W5 window.
Panel B reports cross-wave compositional shifts at each consecutive
wave-pair transition. L1 distance is the sum of absolute share changes
across all occupation cells (six-digit SOC, or 22 SOC major groups)
between the two waves. The SOC 15 share change is the percentage-point
movement in the Computer and Mathematical major-group share of total
Claude.ai consumer conversations. Panel C decomposes the cross-wave
movement of the SOC 15 share into its six-digit sub-occupation
components. For each wave pair, the growth ratio is the wave-(k+1)
sub-occupation share divided by the wave-k share, computed within each
SOC 15 sub-occupation with positive share in both waves; the
cross-sub-occupation coefficient of variation is the standard deviation
of growth ratios divided by their mean. A coefficient of variation near
zero would indicate proportional growth across sub-occupations
(consistent with within-occupation behavior-channel $\theta$ effects acting
uniformly); a coefficient of variation well above zero indicates
heterogeneous growth and is consistent with cross-occupation
composition-channel $\psi$ effects driven by new-user entry into specific
sub-occupations. The three sub-occupations contributing the largest
positive share change between Waves 1 and 3 are Database Architects
(2.2×); Software Developers (1.8×); Software Quality Assurance Analysts
and Testers (1.1×). Cluster-robust standard errors for the corresponding
wave-by-wave reweighted DiD are reported in Table 4.}

\endgroup
\end{landscape}
\clearpage
\begingroup
\scriptsize
\setlength{\tabcolsep}{2pt}
\renewcommand{\arraystretch}{0.78}
\renewcommand{\baselinestretch}{0.85}\selectfont
\textbf{Table B.12. \hyperlink{Bick2026}{Bick, Blandin, and Deming (2026)} Real-Time
Population Survey Micro Release}

\emph{\textbf{Panel A. RPS AI use rates at work, by SOC major group}}

{\def\LTcaptype{none} 
\begin{longtable}[]{@{}
  >{\raggedright\arraybackslash}p{0.4400\linewidth}
  >{\centering\arraybackslash}p{0.1400\linewidth}
  >{\centering\arraybackslash}p{0.1400\linewidth}
  >{\centering\arraybackslash}p{0.1400\linewidth}
  >{\centering\arraybackslash}p{0.1400\linewidth}@{}}
\toprule\noalign{}
\begin{minipage}[b]{\linewidth}\raggedright
\textbf{SOC Major Group}
\end{minipage} & \begin{minipage}[b]{\linewidth}\centering
\textbf{N}
\end{minipage} & \begin{minipage}[b]{\linewidth}\centering
\textbf{Use at work (ever)}
\end{minipage} & \begin{minipage}[b]{\linewidth}\centering
\textbf{Use at work (last wk)}
\end{minipage} & \begin{minipage}[b]{\linewidth}\centering
\textbf{Use at work (daily)}
\end{minipage} \\
\midrule\noalign{}
\endhead
\bottomrule\noalign{}
\endlastfoot
Management & 1363 & 0.524 & 0.468 & 0.158 \\
Computer and Mathematical & 544 & 0.547 & 0.463 & 0.146 \\
Business and Financial & 475 & 0.477 & 0.399 & 0.095 \\
Farming, Fishing, Forestry & 57 & 0.388 & 0.345 & 0.048 \\
Education, Training, Library & 304 & 0.399 & 0.340 & 0.065 \\
Healthcare Practitioners & 300 & 0.372 & 0.292 & 0.134 \\
Construction and Extraction & 284 & 0.353 & 0.280 & 0.084 \\
Arts, Entertainment, Sports, Media & 265 & 0.340 & 0.278 & 0.079 \\
Architecture and Engineering & 150 & 0.313 & 0.274 & 0.094 \\
Community and Social Service & 157 & 0.295 & 0.249 & 0.026 \\
Life, Physical, Social Science & 95 & 0.319 & 0.211 & 0.040 \\
Sales & 536 & 0.256 & 0.201 & 0.043 \\
Installation, Maintenance, Repair & 171 & 0.315 & 0.195 & 0.060 \\
Office and Administrative Support & 656 & 0.227 & 0.195 & 0.046 \\
Production & 354 & 0.262 & 0.193 & 0.033 \\
Healthcare Support & 287 & 0.181 & 0.150 & 0.016 \\
Building and Grounds Cleaning & 142 & 0.230 & 0.140 & 0.038 \\
Personal Care & 180 & 0.158 & 0.137 & 0.000 \\
Protective Service & 68 & 0.133 & 0.133 & 0.010 \\
Transportation and Material Moving & 303 & 0.167 & 0.126 & 0.044 \\
Legal & 68 & 0.198 & 0.124 & 0.006 \\
Food Preparation & 263 & 0.096 & 0.075 & 0.008 \\
\end{longtable}
}

\emph{\textbf{Panel B. AI product shares among at-work AI users, n =
1,785}}

{\def\LTcaptype{none} 
\begin{longtable}[]{@{}
  >{\raggedright\arraybackslash}p{0.5000\linewidth}
  >{\centering\arraybackslash}p{0.5000\linewidth}@{}}
\toprule\noalign{}
\begin{minipage}[b]{\linewidth}\raggedright
\textbf{AI Product}
\end{minipage} & \begin{minipage}[b]{\linewidth}\centering
\textbf{National share}
\end{minipage} \\
\midrule\noalign{}
\endhead
\bottomrule\noalign{}
\endlastfoot
ChatGPT & 0.731 \\
Gemini & 0.460 \\
Embedded & 0.403 \\
GitHub Copilot & 0.173 \\
Claude & 0.123 \\
Midjourney & 0.120 \\
DALL-E & 0.072 \\
Perplexity & 0.063 \\
Scribe & 0.052 \\
Other & 0.049 \\
Synthesia & 0.037 \\
\end{longtable}
}

\emph{\textbf{Panel C. Between-occupation alignment of platform user
base with RPS use rate}}

{\def\LTcaptype{none} 
\begin{longtable}[]{@{}
  >{\raggedright\arraybackslash}p{0.4500\linewidth}
  >{\centering\arraybackslash}p{0.1833\linewidth}
  >{\centering\arraybackslash}p{0.1833\linewidth}
  >{\centering\arraybackslash}p{0.1833\linewidth}@{}}
\toprule\noalign{}
\begin{minipage}[b]{\linewidth}\raggedright
\textbf{Platform}
\end{minipage} & \begin{minipage}[b]{\linewidth}\centering
\textbf{Spearman rho}
\end{minipage} & \begin{minipage}[b]{\linewidth}\centering
\textbf{p-value}
\end{minipage} & \begin{minipage}[b]{\linewidth}\centering
\textbf{N occupations}
\end{minipage} \\
\midrule\noalign{}
\endhead
\bottomrule\noalign{}
\endlastfoot
Claude.ai consumer (Wave 5) & +0.389 & \textless0.001 & 239 \\
1P API enterprise (Wave 5) & +0.279 & \textless0.001 & 216 \\
Microsoft Copilot & +0.252 & \textless0.001 & 270 \\
\end{longtable}
}

\needspace{6\baselineskip}
\textbf{Notes:} \emph{The RPS is a
national online survey of working-age adults conducted on Qualtrics with
awareness-adjusted demographic weights. It is not a probability sample
of the U.S. workforce. Panel A reports weighted AI use rates at work by
SOC major group from the pooled August and November 2024 micro release.
Panel B reports the share of at-work AI users employing each AI product,
n = 1,785, with multiple products per respondent permitted. Panel C
reports the between-occupation Spearman correlation between the
platform-to-workforce density ratio psi (per-occupation platform share
divided by OEWS workforce share) and the RPS at-work AI use rate at the
six-digit SOC level. Sample for Panel C restricted to SOC cells with at
least five RPS respondents.}

\endgroup
\clearpage
\begin{figure}[!p]
\centering
{\normalsize\bfseries Figure B.1. Holm-Bonferroni and Benjamini-Hochberg forest plot for the fifteen-outcome cross-domain family.}\par\medskip
\includegraphics[width=0.97\linewidth,height=0.78\textheight,keepaspectratio]{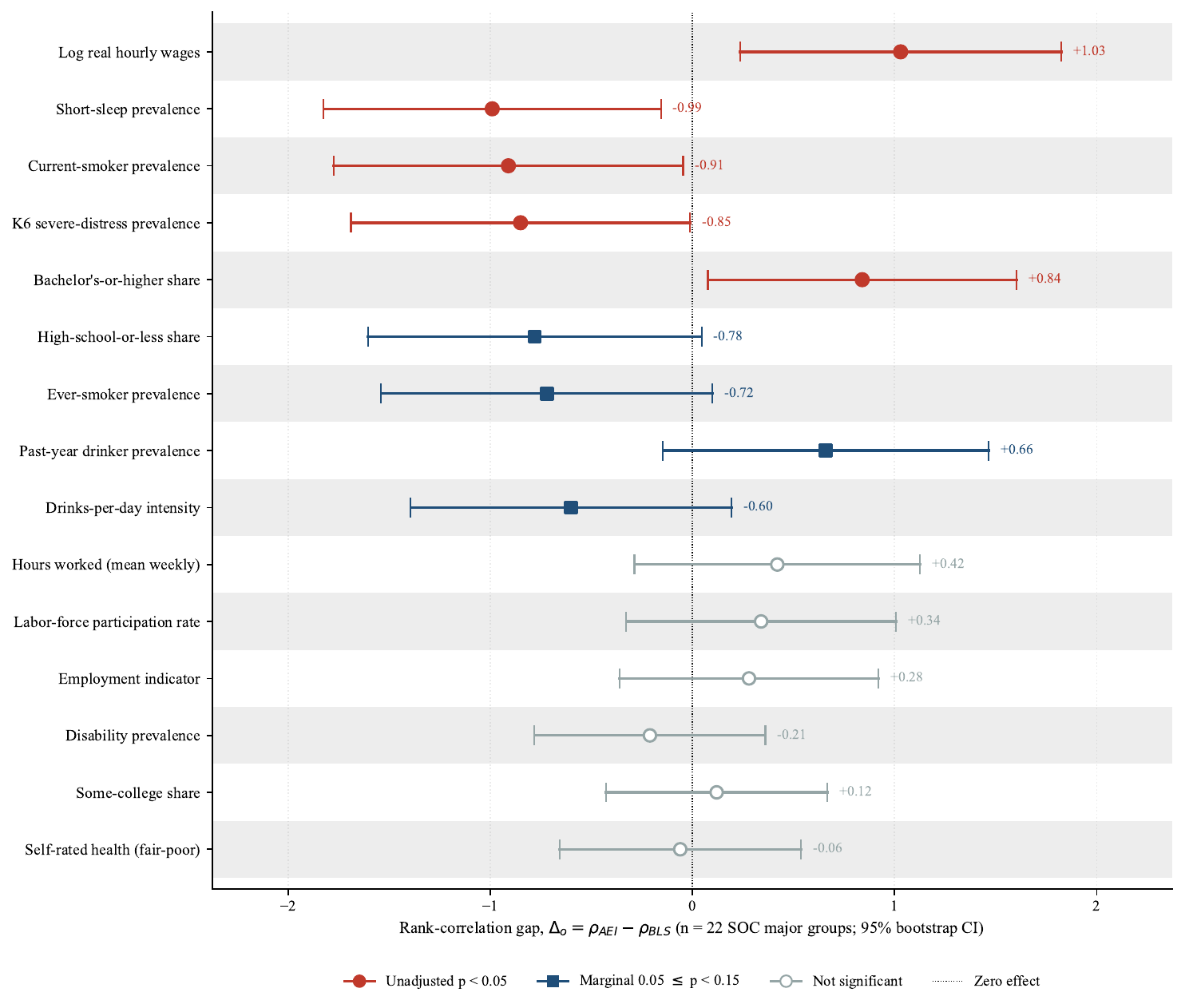}
\par\medskip
\begin{minipage}{0.95\linewidth}
\color[HTML]{333333}\footnotesize
{\bfseries Notes:} \textit{Each row is one of fifteen occupation-level outcomes drawn from the ACS 2011-2022 and NHIS Sample Adult 2017-2018, ordered by absolute Spearman rank-correlation gap $\Delta_o = \rho_{AEI} - \rho_{BLS}$ at the 22 SOC major-group level (n=22). Horizontal bars are 95-percent cluster-bootstrap CIs from 5,000 replications. Red filled circles mark unadjusted-significant outcomes (p<0.05); navy filled squares mark marginal outcomes (0.05$\le$ p<0.15); light gray hollow circles mark not significant. Under Holm-Bonferroni FWER=0.05, zero outcomes survive; under BH-FDR q=0.10, zero survive; under BH-FDR q=0.20, seven outcomes survive.}
\end{minipage}
\end{figure}
\clearpage
\begin{figure}[!p]
\centering
{\normalsize\bfseries Figure B.2. SOC major-group rank trajectories across five Anthropic Economic Index waves.}\par\medskip
\includegraphics[width=0.97\linewidth,height=0.78\textheight,keepaspectratio]{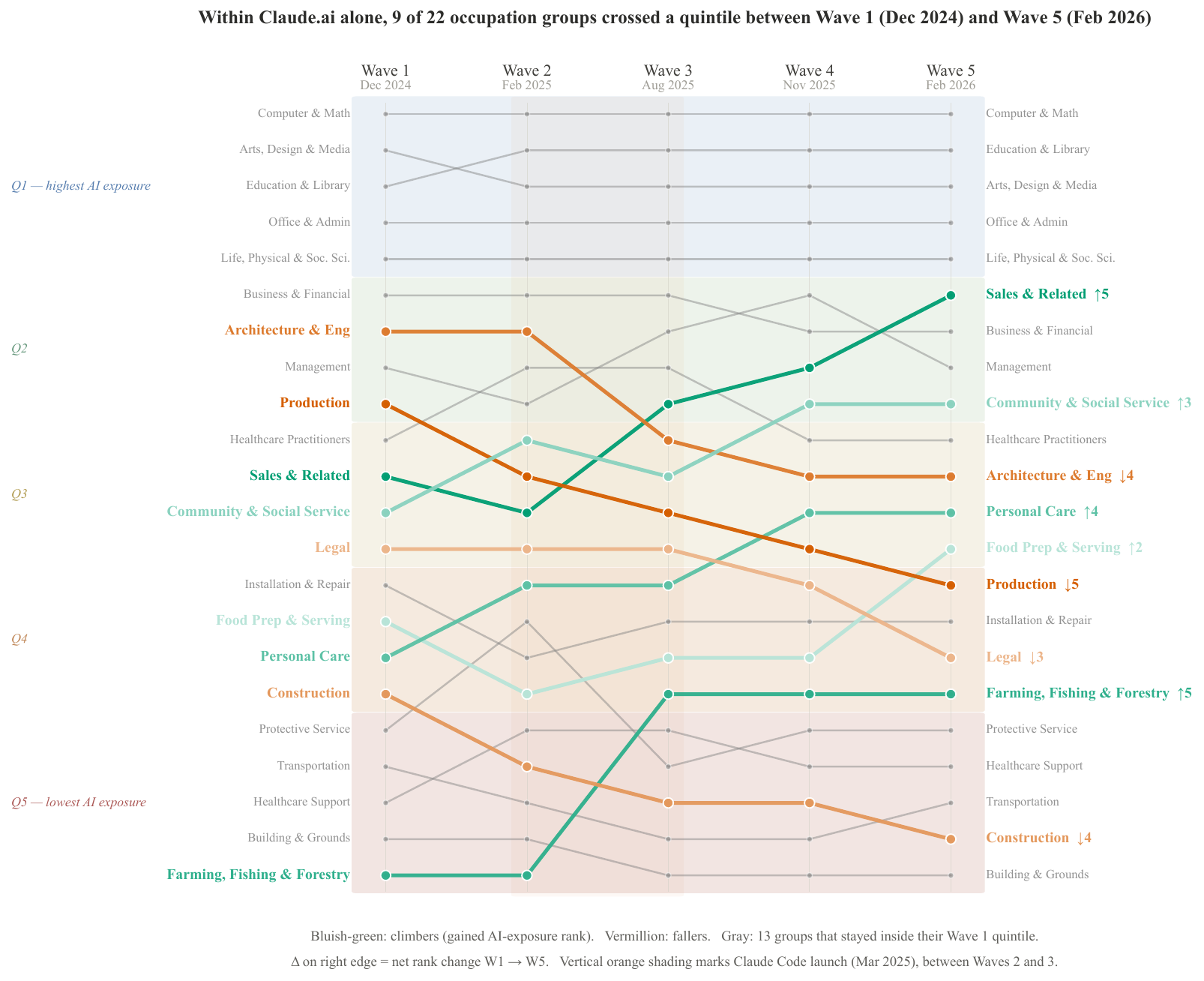}
\par\medskip
\begin{minipage}{0.95\linewidth}
\color[HTML]{333333}\footnotesize
{\bfseries Notes:} \textit{Each line tracks one SOC major group's Anthropic Claude.ai conversation-share rank across the five publicly released AEI waves (Wave 1 December 2024 through Wave 5 February 2026). Rank 1 = highest share; rank 22 = lowest. Vertical reference lines mark the Claude Code launch (between Waves 2 and 3) and the Claude Projects expansion (between Waves 3 and 4), the two product events that drove the largest cross-wave compositional shifts.}
\end{minipage}
\end{figure}
\clearpage
\begin{landscape}
\begin{figure}[!p]
\centering
{\normalsize\bfseries Figure B.3. Economic vulnerability versus platform visibility across 95 detailed occupations.}\par\medskip
\includegraphics[width=1.0\linewidth,height=0.85\textheight,keepaspectratio]{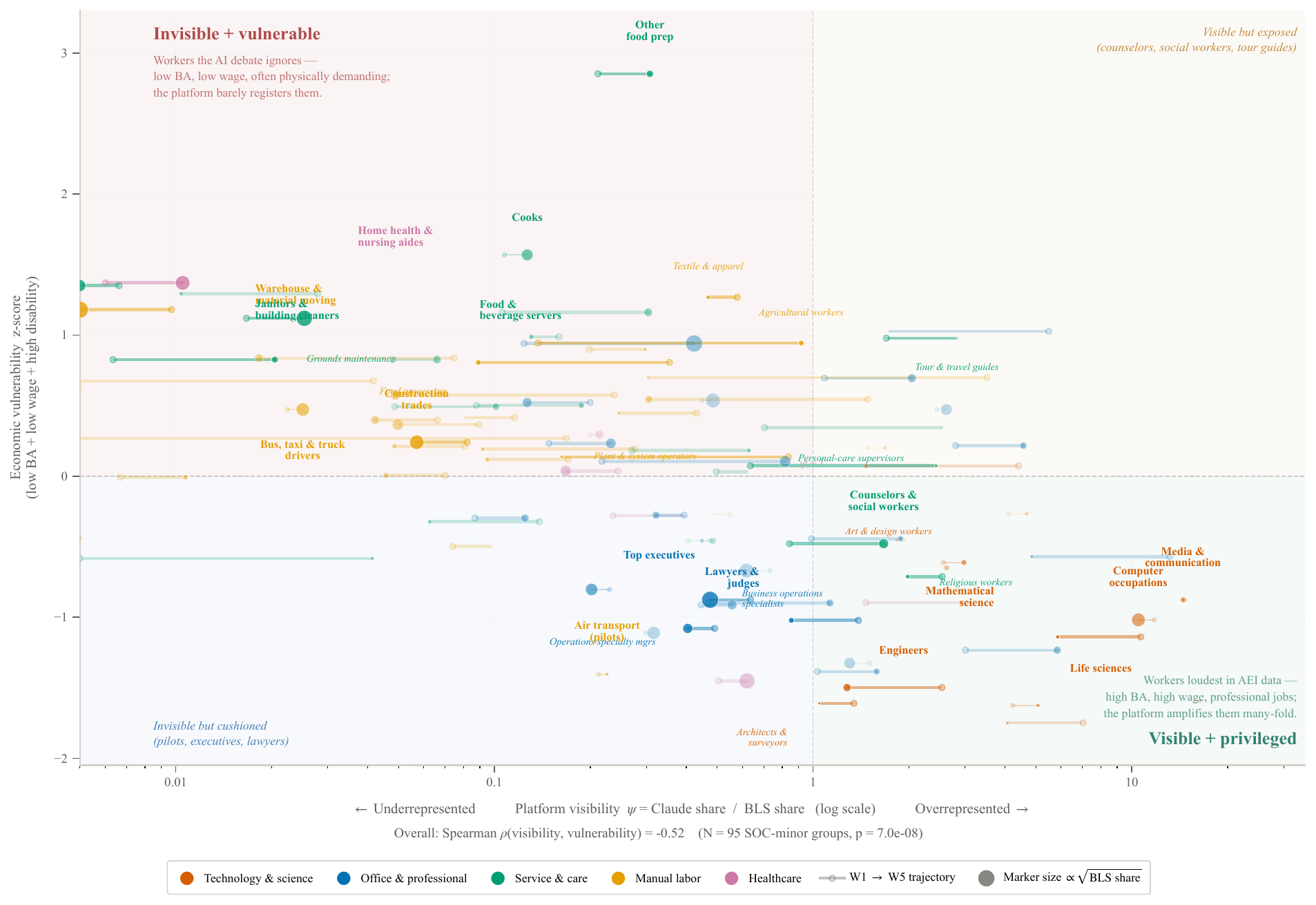}
\par\medskip
\begin{minipage}{0.95\linewidth}
\color[HTML]{333333}\footnotesize
{\bfseries Notes:} \textit{Horizontal axis: economic vulnerability index combining wage rank, displacement risk, and unemployment-insurance recipiency. Vertical axis: platform-visibility share (Anthropic Wave 5 Claude.ai conversation share divided by BLS OES employment share). Each point is one of the 95 most-populous six-digit SOC occupations. The lower-left quadrant (high vulnerability, low visibility) contains occupations whose DiD coefficient inherits the largest user-base bias.}
\end{minipage}
\end{figure}
\end{landscape}
\clearpage
\begin{figure}[!p]
\centering
{\normalsize\bfseries Figure B.4. Occupational composition and demographic profile across the 95 most-populous detailed SOC occupations.}\par\medskip
\includegraphics[width=1.0\linewidth,height=0.83\textheight,keepaspectratio]{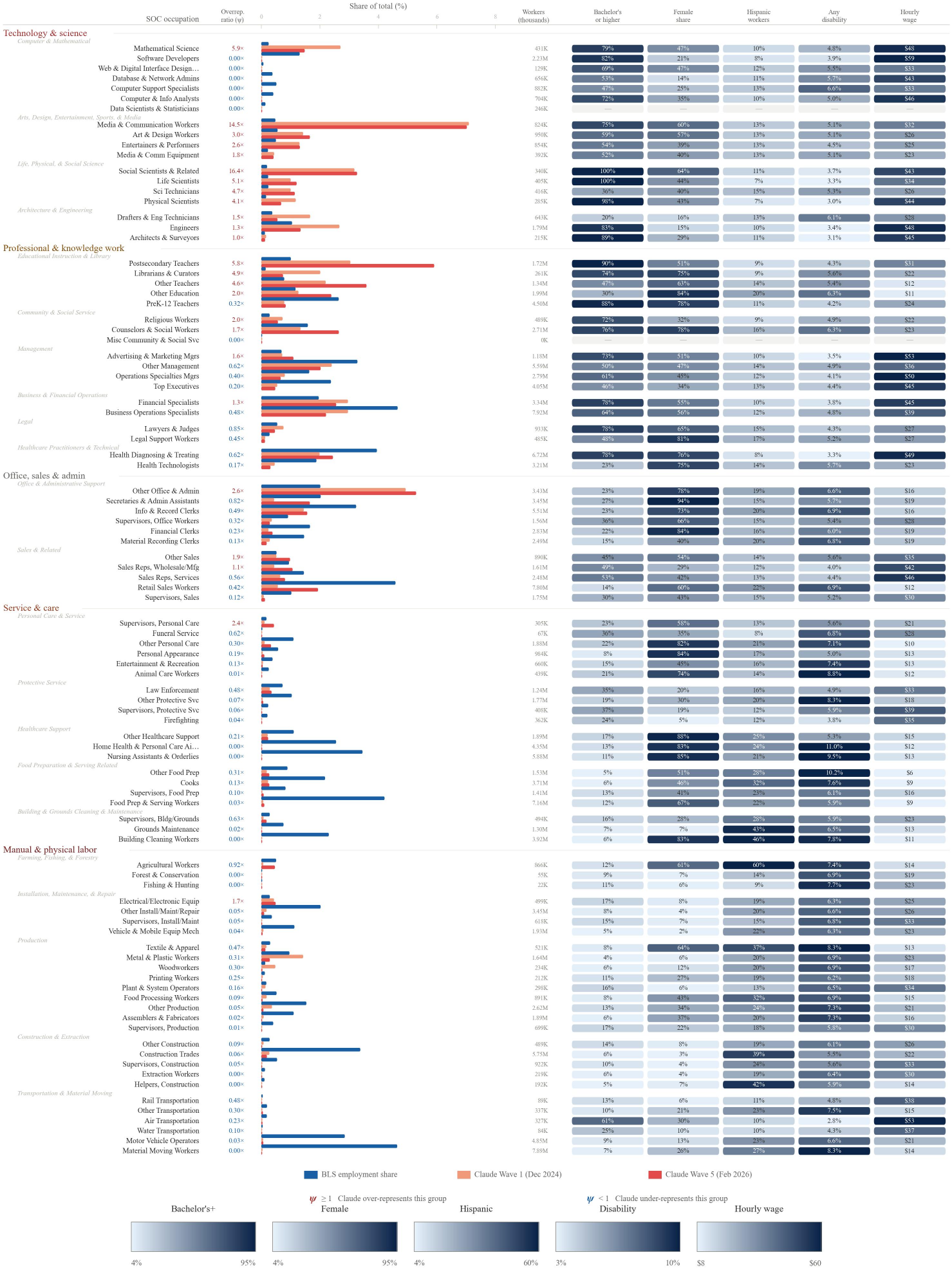}
\par\medskip
\begin{minipage}{0.95\linewidth}
\color[HTML]{333333}\footnotesize
{\bfseries Notes:} \textit{Each row is one of the 95 most-populous six-digit SOC occupations ranked by BLS workforce employment. Cells report the BLS workforce share, Anthropic Wave 5 Claude.ai conversation share, over-representation ratio $\psi$, and demographic composition from ACS 2024. The over-representation ratio $\psi$ exceeds ten for several Computer and Mathematical occupations and falls below 0.05 for Transportation, Food Preparation, and Construction occupations.}
\end{minipage}
\end{figure}
\clearpage
\begin{figure}[!p]
\centering
{\normalsize\bfseries Figure B.5. Occupation rank flow from platform-derived to workforce-reweighted exposure rankings.}\par\medskip
\includegraphics[width=1.0\linewidth,height=0.83\textheight,keepaspectratio]{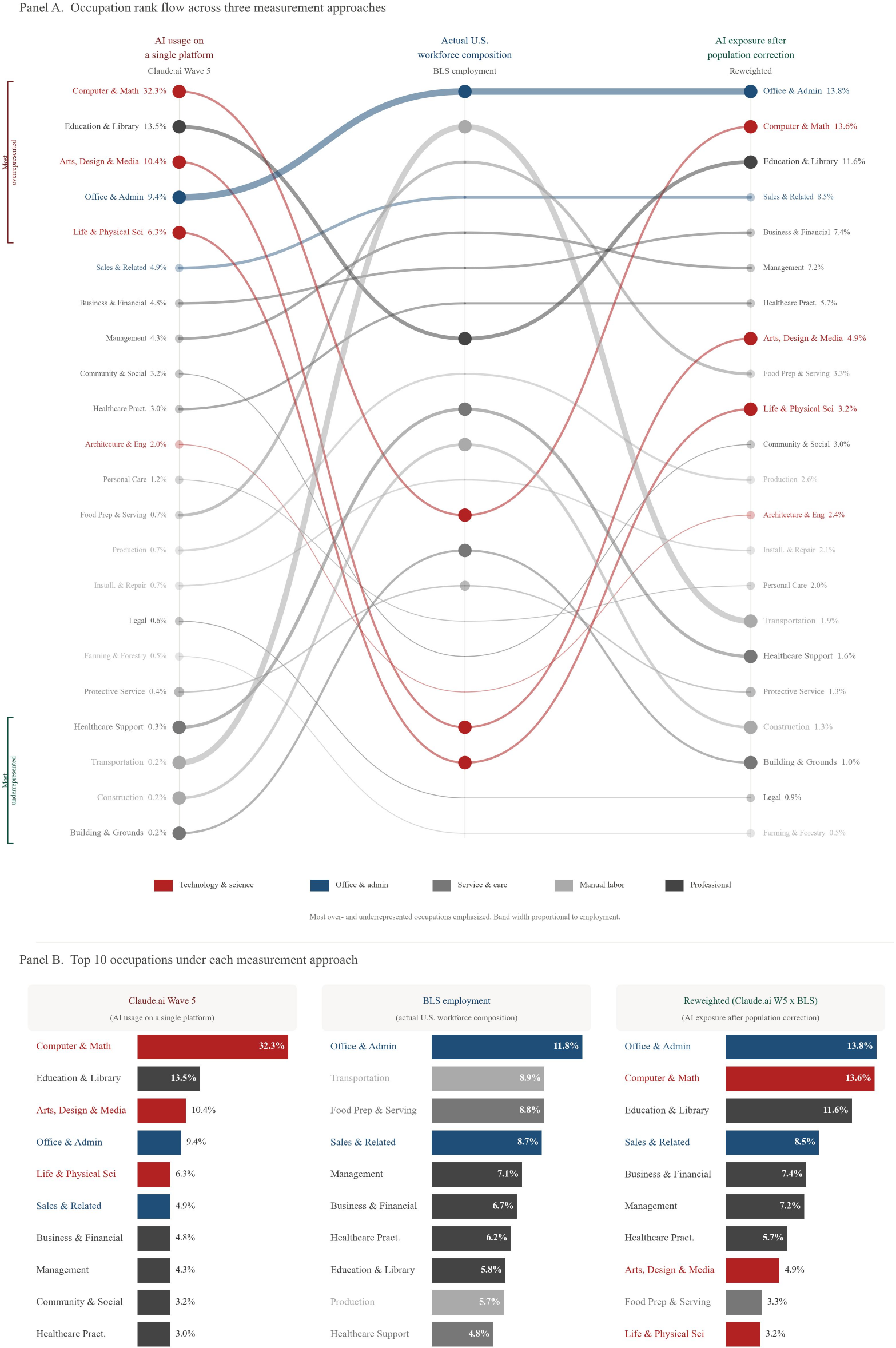}
\par\medskip
\begin{minipage}{0.95\linewidth}
\color[HTML]{333333}\footnotesize
{\bfseries Notes:} \textit{Alluvial flow diagram of rank movement at the six-digit SOC level. Left axis: Anthropic Wave 5 Claude.ai conversation-share ranking. Right axis: Composite Reweighted ranking. The top decile under the unreweighted ranking is dominated by Computer and Mathematical occupations; under reweighting, that decile shifts toward Office and Administrative Support, Sales, and Customer Service occupations.}
\end{minipage}
\end{figure}
\clearpage
\begin{landscape}
\begin{figure}[!p]
\centering
{\normalsize\bfseries Figure B.6. Within-platform heterogeneity in the post-2023 employment coefficient across fifteen demographic and occupation subgroups.}\par\medskip
\includegraphics[width=0.99\linewidth,height=0.80\textheight,keepaspectratio]{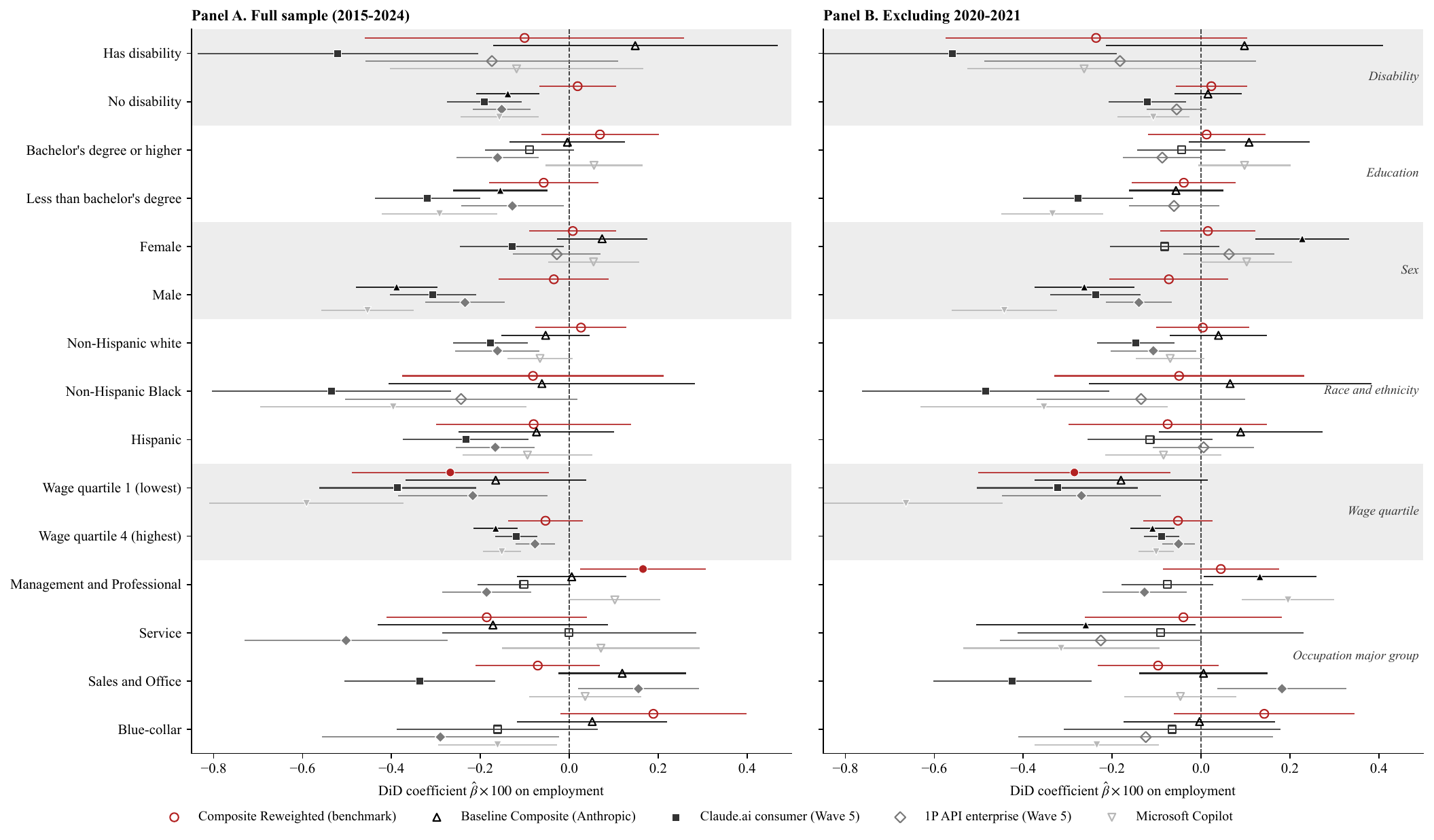}
\par\medskip
\begin{minipage}{0.95\linewidth}
\color[HTML]{333333}\footnotesize
{\bfseries Notes:} \textit{Forest plot of post-2023 DiD coefficient ($\beta \times 100$) on the employment indicator from equation (7). Panel A: full sample. Panel B: excluding 2020 and 2021. Subgroups are stratified by disability, education, sex, race-ethnicity, wage quartile, and occupation major group. Five exposure variants plotted per subgroup: Composite Reweighted (red open circle, benchmark), Baseline Composite (Anthropic) (open triangle), Claude.ai consumer Wave 5 (filled square), 1P API Wave 5 (open diamond), Microsoft Copilot raw (open inverted triangle). Bars are 95-percent cluster-robust CIs at the six-digit SOC.}
\end{minipage}
\end{figure}
\end{landscape}
\clearpage
\begin{figure}[!p]
\centering
{\normalsize\bfseries Figure B.7. Three measures per SOC major group from the RPS micro release.}\par\medskip
\includegraphics[width=0.97\linewidth,height=0.78\textheight,keepaspectratio]{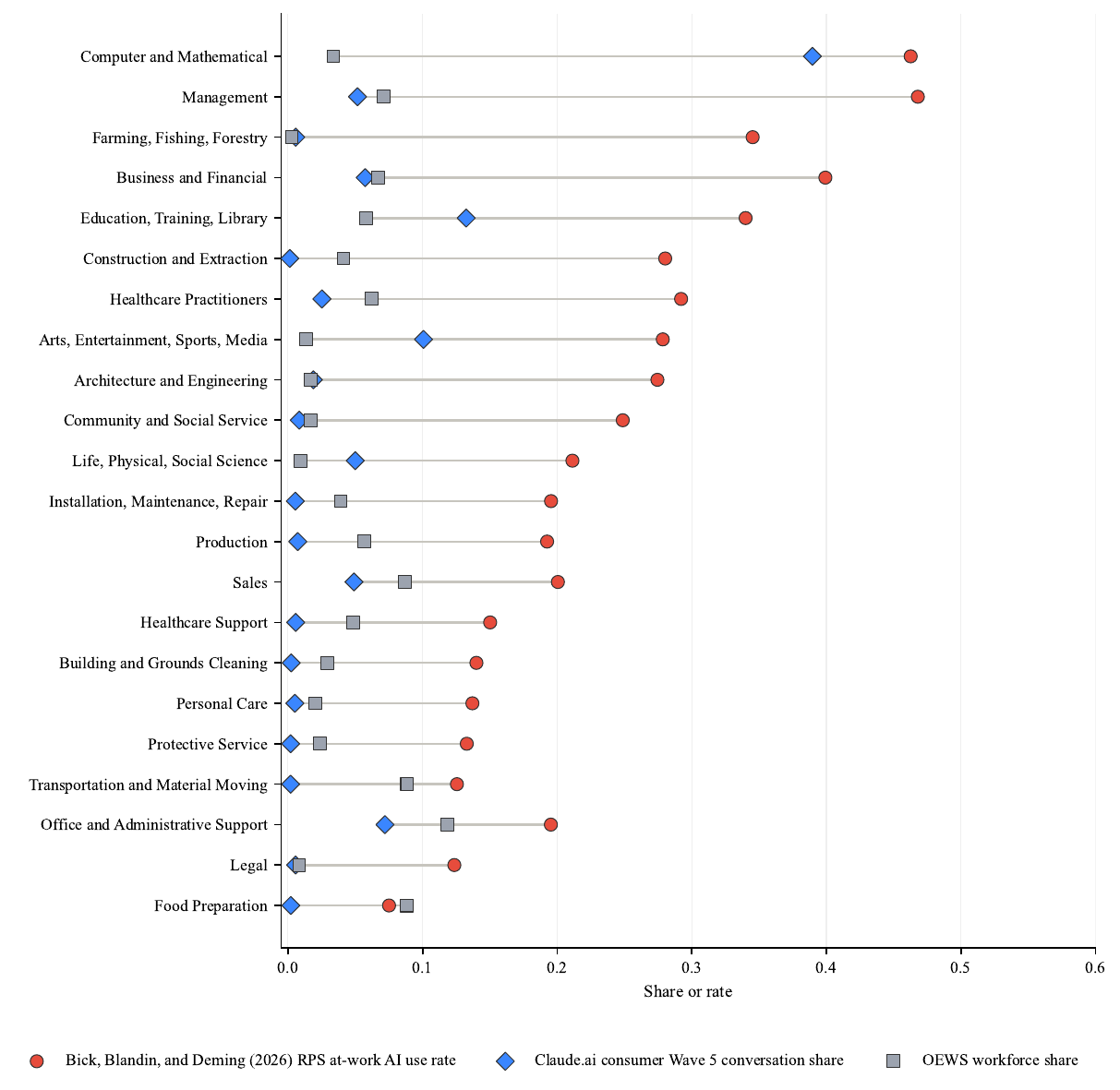}
\par\medskip
\begin{minipage}{0.95\linewidth}
\color[HTML]{333333}\footnotesize
{\bfseries Notes:} \textit{Each row reports three points by SOC major group, ordered by spread (max minus min) descending. The red circle is the RPS at-work AI use rate from the pooled August and November 2024 micro release. The blue diamond is the Anthropic Claude.ai consumer Wave 5 conversation share. The gray square is the OEWS workforce employment share. The thin gray line connects the minimum and maximum across the three measures, indicating the spread within each occupation.}
\end{minipage}
\end{figure}
\clearpage
\section*{Online Appendix C. Implied Incidence of a Retraining Program (Policy Simulation)}
\addcontentsline{toc}{section}{Online Appendix C}
To illustrate the policy implications of the platform-selection bias documented in the main text, we conduct a hypothetical \$10-billion AI retraining fund allocation exercise. Two rules are compared. Under the platform-weighted rule, the allocation follows Anthropic Wave 5 Claude.ai conversation shares: occupations with the highest platform exposure receive proportionally more retraining dollars. Under the workforce-reweighted rule, the allocation follows BLS May 2024 OEWS employment shares: dollars track where workers are, not where platform conversations are.

The platform-weighted rule redirects approximately \$3.87 billion (39 percent of the fund) to occupations with above-median wages and Bachelor's-degree-or-higher shares above 60 percent, primarily Computer and Mathematical, Business and Financial Operations, and Office and Administrative Support occupations. The workforce-reweighted rule directs the same dollars toward middle-wage occupations with Bachelor's-or-higher shares closer to the workforce mean, primarily Sales, Office Support, Transportation, Production, and Food Preparation.

Figure C.1 visualizes the incidence shift. The platform-weighted bars show a redistribution of dollars toward high-skill occupations; the workforce-reweighted bars match the underlying workforce composition. The \$3.87-billion shift is the implied compositional bias when a policy maker uses platform-derived exposure as an allocation key without workforce reweighting. The exercise treats the platform-weighted rule as a benchmark for what would happen if the published Anthropic conversation shares were taken as a direct measure of training need, holding total program dollars fixed.

The simulation is illustrative rather than prescriptive. The Bureau of Labor Statistics workforce distribution is the relevant denominator if the policy aim is workforce-wide coverage, but a policy specifically targeting workers most likely to be affected by AI productivity gains may have different weighting objectives. The \$3.87-billion gap quantifies the magnitude of the user-base channel in a concrete dollar amount and isolates the compositional component from the substantive question of how AI exposure maps to retraining need.

\begin{figure}[!p]
\centering
{\normalsize\bfseries Figure C.1. Implied incidence of a hypothetical \$10-billion AI retraining fund (policy simulation).}\par\medskip
\includegraphics[width=0.97\linewidth,height=0.78\textheight,keepaspectratio]{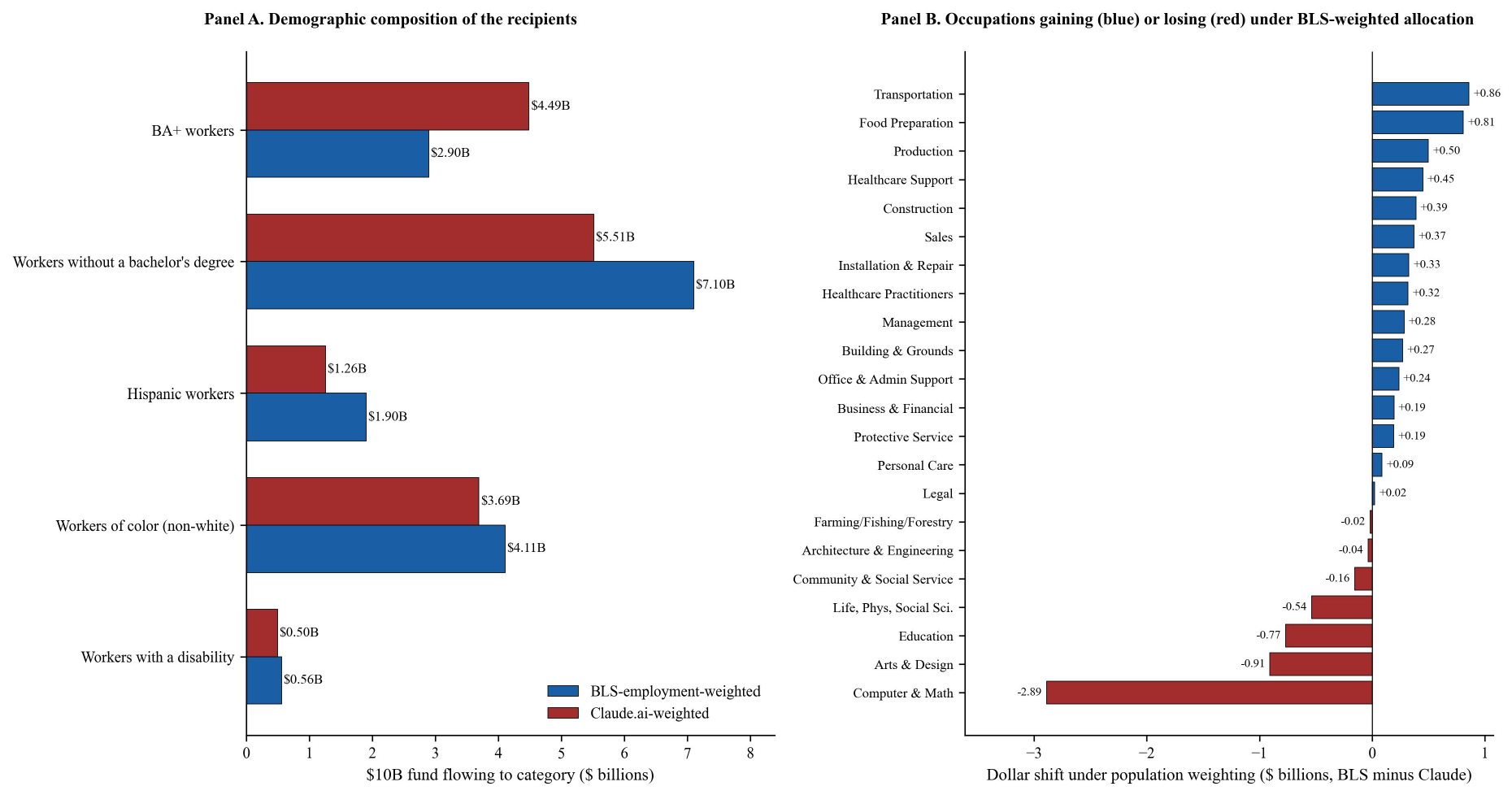}
\par\medskip
\begin{minipage}{0.92\linewidth}
\color[HTML]{333333}\footnotesize
{\bfseries Notes:} \textit{Each bar reports the share of the \$10-billion fund directed to a SOC major-group bin under two allocation rules. The platform-weighted rule allocates dollars in proportion to Anthropic Wave 5 Claude.ai conversation shares; the workforce-weighted rule allocates in proportion to BLS May 2024 OEWS employment shares. The platform-weighted rule directs approximately \$3.87 billion (39 percent) to high-wage, high-education occupations; the workforce-weighted rule directs the same dollars to middle-wage occupations with workforce-mean education shares.}
\end{minipage}
\end{figure}
\clearpage


\end{document}